\documentclass[10pt]{article}
\PassOptionsToPackage{authoryear,sort}{natbib}

\usepackage[preprint]{neurips_2026}
\makeatletter
\renewcommand{\@notice}{}
\makeatother

\usepackage[utf8]{inputenc}
\usepackage[T1]{fontenc}
\usepackage{microtype}

\usepackage{amsmath}
\usepackage{amsfonts}
\usepackage{amssymb}
\usepackage{algorithm}
\usepackage{algpseudocode}

\usepackage{graphicx}
\usepackage{listings}
\graphicspath{{fig/}}
\usepackage{subcaption}
\usepackage{booktabs}
\usepackage{longtable}
\usepackage{multirow}
\usepackage{makecell}
\usepackage{tabularx}
\usepackage{array}
\usepackage{adjustbox}
\usepackage{wrapfig}
\usepackage{float}
\usepackage{placeins}

\usepackage[table]{xcolor}
\definecolor{AblationAccent}{HTML}{245879}
\definecolor{AblationTint}{HTML}{EDF4F8}
\usepackage{enumitem}
\usepackage[most]{tcolorbox}
\usepackage{nicefrac}
\usepackage{titletoc}
\usepackage{url}
\usepackage[commandnameprefix=ifneeded]{changes}
\makeatletter
\newcommand{\addedblock}{\ifbool{Changes@optiondraft}{\color{blue}}{}}
\makeatother
\usepackage{hyperref}
\hypersetup{
    colorlinks=true,
    linkcolor=blue,
    citecolor=blue,
    urlcolor=blue
}

\usepackage{aux/macros}
\usepackage{amsmath,amsfonts,bm}

\def\eqref#1{equation~\ref{#1}}
\def\1{\bm{1}}

\DeclareMathAlphabet{\mathsfit}{\encodingdefault}{\sfdefault}{m}{sl}
\SetMathAlphabet{\mathsfit}{bold}{\encodingdefault}{\sfdefault}{bx}{n}

\usepackage{framed}

\usepackage{colortbl}
\definecolor{TableHeader}{HTML}{EDF2F7}
\definecolor{GainLight}{HTML}{EAF5EC}
\definecolor{GainMedium}{HTML}{D4EAD8}
\definecolor{GainStrong}{HTML}{BCE0C4}
\definecolor{LossLight}{HTML}{FBEDEC}
\definecolor{LossMedium}{HTML}{F5D8D5}
\definecolor{LossStrong}{HTML}{EDC0BC}

\definecolor{VarLight}{HTML}{F4F6F8}
\definecolor{VarMedium}{HTML}{E5EDF4}
\definecolor{VarStrong}{HTML}{CBDCEB}

\usepackage[most]{tcolorbox}

\definecolor{ExampleBg}{HTML}{F5F8FB}
\definecolor{ExampleBorder}{HTML}{C9D8E5}
\definecolor{ExampleTitleBg}{HTML}{E7EFF6}

\newtcolorbox{trajectoryexample}[1]{
    breakable,
    colback=ExampleBg,
    colframe=ExampleBorder,
    colbacktitle=ExampleTitleBg,
    coltitle=black,
    fonttitle=\bfseries\small,
    title={#1},
    boxrule=0.5pt,
    arc=1mm,
    left=8pt,
    right=8pt,
    top=7pt,
    bottom=7pt,
    before skip=8pt,
    after skip=8pt
}

\setlist{nosep,leftmargin=*}

\title{%
  $\vcenter{\hbox{\includegraphics[height=0.6in]{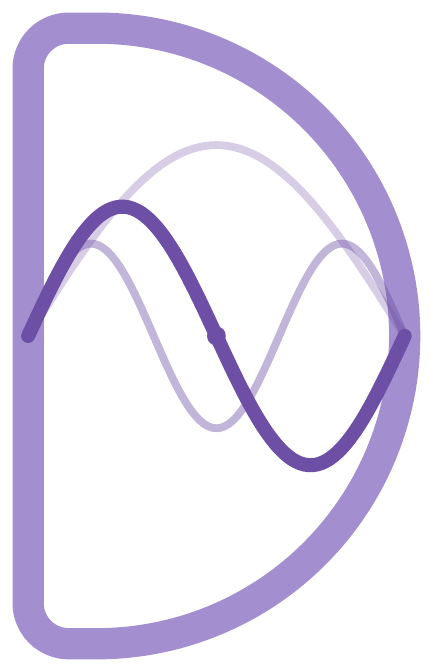}}}$%
  \hspace{0.12in}%
  \begin{tabular}{@{}c@{}}
  \benchname: Can AI Agents Perform \\
  Real-World Physics Engineering Design?
  \end{tabular}%
}
\author{%
  \textbf{Ali Momeni\textsuperscript{1,2*}}\quad
  \textbf{Babak Rahmani\textsuperscript{3*}}\quad
  \textbf{Jiahui Wang\textsuperscript{1}}\quad
  \textbf{Amir Jafargholi\textsuperscript{1}}\\[4pt]
  \textbf{David Norris\textsuperscript{2}}\enspace
  \textbf{Matthias Bethge\textsuperscript{3}}\enspace
  \textbf{Nader Engheta\textsuperscript{4}}\enspace
  \textbf{Romain Fleury\textsuperscript{1}}%
}
\date{}

\usepackage{todonotes}

\begin{document}
\raggedbottom

% Enlarge the opening page while retaining the normal body-page geometry.
\savegeometry{paperbody}
\newgeometry{
  left=0.65in,right=0.65in,
  top=0.55in,bottom=0.55in,
  headheight=12pt,headsep=25pt,footskip=30pt
}

\maketitle

% Keep affiliations and equal contribution in one compact first-page footnote.
\begingroup
  \renewcommand{\thefootnote}{}
  \setlength{\parindent}{0pt}
  \makeatletter
  \renewcommand{\@makefntext}[1]{\noindent#1}
  \makeatother
  \interfootnotelinepenalty=10000
  \footnotetext{%
    \textsuperscript{1}Laboratory of Wave Engineering, School of Electrical Engineering, Swiss Federal Institute of Technology in Lausanne (EPFL), Lausanne, Switzerland.\par
    \textsuperscript{2}Optical Materials Engineering Laboratory, Department of Mechanical and Process Engineering, ETH Zurich, 8092 Zurich, Switzerland.\par
    \textsuperscript{3}T\"ubingen AI Center.\par
    \textsuperscript{4}University of Pennsylvania, Department of Electrical and Systems Engineering, Philadelphia, Pennsylvania, U.S.A.\par
    \textsuperscript{*}Equal contribution.%
  }
\endgroup

\begin{abstract}
AI agents can increasingly write code, use engineering software, and complete complex workflows. But can they turn an engineering specification into a working physical design? Unlike conventional coding tasks, physics-based engineering design requires agents to navigate physical constraints, interpret simulation feedback, and iteratively improve solutions whose performance cannot be determined from code alone. We introduce \benchname{}, a benchmark of 26 end-to-end physics-based engineering tasks, comprising 16 open-ended Design tasks and 10 Reimplementation tasks based on published devices. Agents work with engineering software under limited time and simulation budgets, with their designs independently evaluated for physical validity and performance. Across 11 model--agent configurations, the highest overall normalized score is 0.474, while the highest engineering pass rate is only 19.2\% (5 of 26 tasks). On open-ended Design tasks, the number of valid final submissions ranges from 2 to all 16 across configurations. Even when agents produce valid reimplementations, they rarely match published performance. We release the benchmark, evaluation harness, and accompanying resources at \href{\benchurl}{\benchname}, with plans to expand the task collection to support continued evaluation of autonomous engineering agents.
\end{abstract}

{% tighter spacing so Figure 1 stays on page 1
\setlength{\intextsep}{3pt}
\begin{figure}[H]
    \centering
    \vspace{-0.5\baselineskip}
    \includegraphics[width=\linewidth]{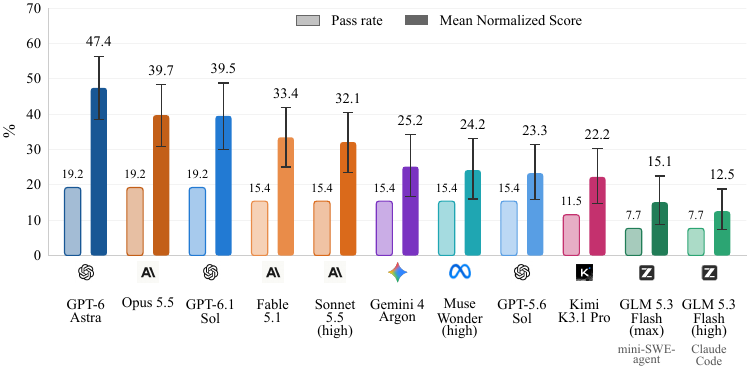}
    \setlength{\abovecaptionskip}{2pt}
    \caption{\textbf{Overall benchmark performance.}
    Mean normalized score ($\times100$) and engineering
    pass rate (\%) across all 26 tasks, comprising
    16 Design and 10 Reimplementation tasks.
    Invalid submissions contribute zero. GPT-5.6 Sol, GPT-6.1 Sol, and GPT-6 Astra use Codex CLI; Opus 5.5, Fable 5.1, Sonnet 5.5, Kimi K3.1 Pro, and GLM 5.3 Flash (high) use Claude Code; Gemini 4 Argon and GLM 5.3 Flash (max) use mini-SWE-agent; and Muse Wonder uses Muse Code. Error bars show 95\% confidence intervals.}
    \label{fig:overall-score}
\end{figure}
}

\loadgeometry{paperbody}

\section{Introduction}

AI agents are beginning to take on tasks that previously required substantial human expertise. In machine learning, they can conduct research experiments~\citep{lu2024ai,lupidi2026airs}, explore neural architectures for pretraining~\citep{pepe2026agentic}, and automate model post-training~\citep{rank2026posttrainbench}. Benchmarks such as GDPval, AutomationBench, and EnterpriseOps-Gym extend evaluation to professional and operational work~\citep{patwardhan2026gdpval,shepard2026automationbench,malay2026enterpriseops}. Beyond ML, AI systems have contributed to scientific discovery by solving longstanding mathematical problems and scientific hypothesis generation~\citep{openai2026mathematics,gottweis2026accelerating}. These advances raise the possibility of automating more of scientific and engineering research. For engineering design in physics, this means turning specifications into physical designs using the same specialized software and simulation tools as human engineers. Designs must satisfy fabrication and operating constraints while balancing competing objectives. A design can execute without error and still fail to meet its physical requirements. The question we are asking is \textit{can AI agents complete this process end to end, as a human engineer would}?

Recent benchmarks have begun evaluating AI agents on increasingly
complex scientific and engineering tasks, moving beyond reasoning
and data analysis towards experimentation and design.
Terminal-Bench-Science~\citep{terminalbenchscience2026} includes
physical simulation and device design, while
FrontierPhysics~\citep{you2026frontierphysics} evaluates open-ended
physics research.
EngDesign~\citep{guo2026toward} and
Frontier-Eng~\citep{chi2026frontier} extend agent evaluation
to engineering design and optimization, while recent systems
demonstrate autonomous photonic design
~\citep{sharma2025ai,kharel2026autonomous}.
However, evaluating general-purpose agents on complete physics-based
engineering workflows, from constructing devices and circuits using
domain-specific design software to validating and iteratively
improving them through physical simulation, remains challenging (Table~\ref{tab:related_work_comparison}).

Consider a wavelength-division multiplexing device that separates
light of different wavelengths into four output channels (Figure~\ref{fig:picbench-task}). The device uses optical rings whose response is tuned by electrical heaters.
The agent must design these structures while accounting for electrical power, thermal crosstalk, and fabrication constraints.
Improving the separation of optical channels may increase power consumption or disturb neighbouring channels. Even a design with good optical performance can fail layout or
connectivity checks. The agent must work across several physical models and revise
its design using their feedback.

\begin{figure}[!t]
    \centering
    \vspace{-0.25cm}
    \includegraphics[width=\textwidth]{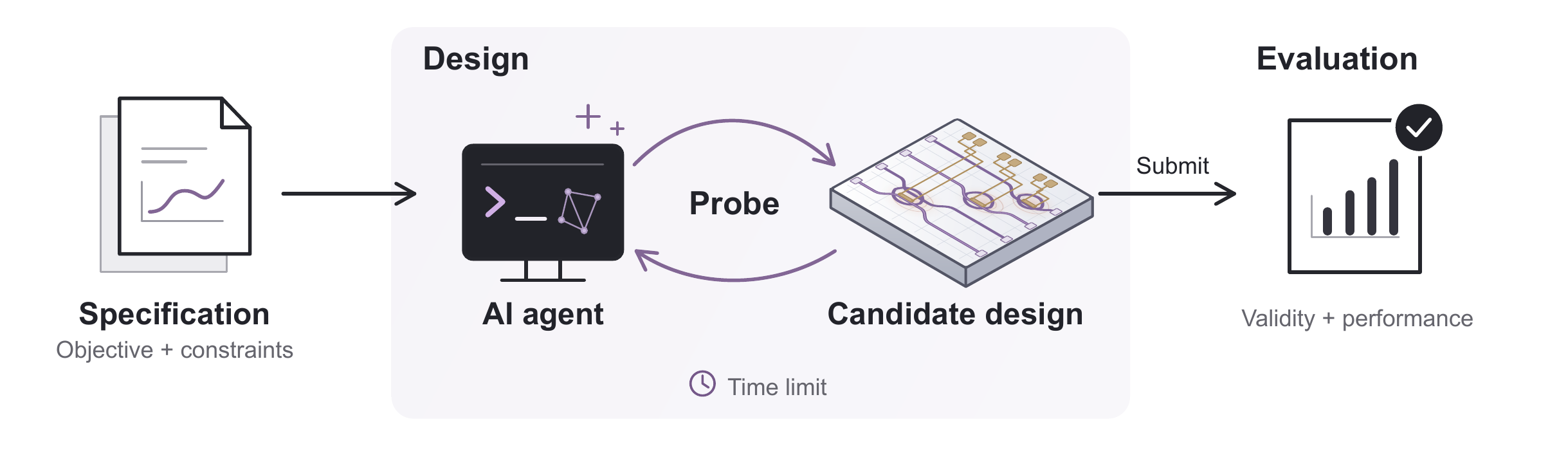}
 \caption{\textbf{\benchname{} evaluation workflow.} Given an engineering specification with performance objectives and physical constraints, an AI agent iteratively develops and tests candidate designs using feedback from a simulation-based evaluator. The agent works within fixed time and evaluator-probe budgets, and its final submission is independently assessed for validity and physical performance.}
    \label{fig:picbench-pipeline}
\end{figure}

We introduce \benchname{} (Figure~\ref{fig:picbench-pipeline}), a benchmark of 26 end-to-end physics-based engineering tasks developed by domain experts. It includes 16 open-ended Design tasks, where agents develop and optimize their own solutions, and 10 Reimplementation tasks based on devices reported in published literature. The tasks cover individual physics components, integrated circuits, and multiphysics systems. Agents work with engineering software under limited time and simulation budgets, and their final submissions are independently evaluated for validity and physical performance.

Across 11 model--agent configurations, the highest overall normalized score is 0.474, with an engineering pass rate of 19.2\% (Figure~\ref{fig:overall-score}). On open-ended Design tasks, the number of tasks with valid final submissions varies from two to all 16 across configurations. On Reimplementation tasks, all final submissions are valid, but agents rarely reproduce the performance reported in the literature. These results expose a gap between producing physically valid designs and achieving their performance targets.

\FloatBarrier

\section{\benchname}
\label{sec:benchmark}

\subsection{Task formulation}
\label{sec:task-formulation}

\begin{figure}[!t]
    \centering
    \includegraphics[width=\linewidth]{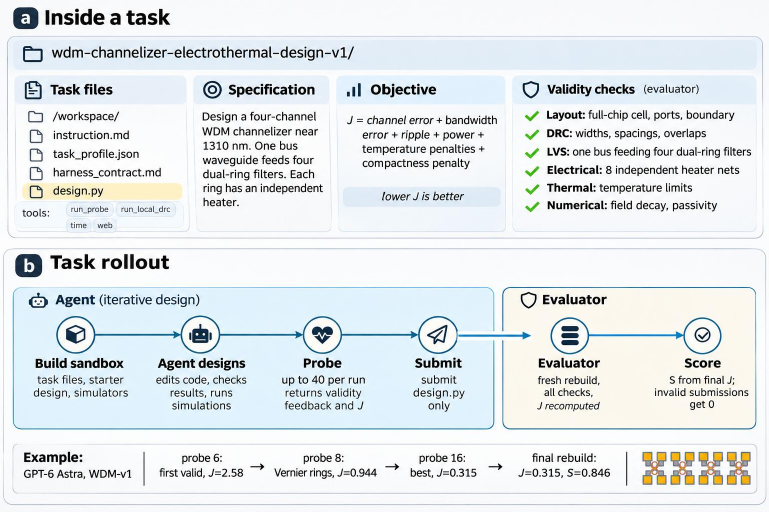}
\caption{\textbf{Inside a \benchname{} task.}
\textbf{(a)} The WDM-v1 workspace, showing the files and tools available to the agent, an example engineering specification, and the validity checks applied by the evaluator.
\textbf{(b)} An example GPT-6 Astra trajectory on WDM-v1, showing how the objective changes across evaluator probes before the final submission is independently rebuilt and evaluated (see Appendix~\ref{app:working-wdm_v1_astra} for a full working example).}
    \label{fig:picbench-task}
\end{figure}

Each task provides an engineering specification, an editable workspace, and access to physics-based simulation and evaluation tools (Figure~\ref{fig:picbench-task}). The specification defines the intended function, physical constraints, and performance objective. Agents develop their designs by modifying Python code and design files, running available engineering software, and testing candidates through a limited number of evaluator probes. Each probe returns feedback on execution, validity, and physical performance, allowing agents to revise their designs within a fixed working budget.

The final submission is independently rebuilt and evaluated. The evaluator checks task-specific requirements, including geometry, connectivity, fabrication constraints, and physical operating limits, before measuring performance. Section~\ref{sec:evaluation-metrics} describes the scoring procedure.

\subsection{Dataset construction}
\label{sec:dataset-construction}

The benchmark contains 26 tasks developed by domain experts. We distinguish two task families based on the information available to the agent and the nature of the design problem.

In \emph{Design} (16 tasks), agents receive a functional specification without a prescribed solution. They must choose an appropriate design, implement it, and improve its performance through experimentation.

In \emph{Reimplementation} (10 tasks), agents receive a published paper describing a device but not its original implementation. They must recover the design choices from the publication, construct an executable implementation, and attempt to reproduce or improve the reported performance.

% \begin{figure}[!t]
%     \centering
%     \includegraphics[width=\linewidth]{picbench-task-anatomy.pdf}
%     \caption{\textbf{Anatomy of a \benchname{} task.}
%     \textbf{(a)} Task construction, from specification and reference design to a verified evaluator.
%     \textbf{(b)} The WDM-v1 workspace, showing agent-accessible files and protected evaluation components.
%     \textbf{(c)} An agent rollout with intermediate evaluator probes and independent final grading. The example shows GPT-6 Astra on WDM-v1 (Appendix~\ref{app:working-wdm_v1_astra}).}
%     \label{fig:picbench-task}
% \end{figure}

Each task in \benchname{} is developed by domain experts, who prepare the specification, starter design, and simulation-based evaluator. Experts verify that each task is solvable using a reference implementation: for Design tasks, they construct a design that satisfies the validity requirements; for Reimplementation tasks, they verify the feasibility of reproducing the published device. They also test the evaluators against valid and deliberately invalid submissions to check that constraints are enforced and performance is computed correctly. Only tasks with a verified reference solution and a functioning evaluator are included in the benchmark. Further details are provided in Appendix~\ref{app:task-solvability}.

\begin{figure}[!t]
    \centering
    \includegraphics[width=\linewidth]{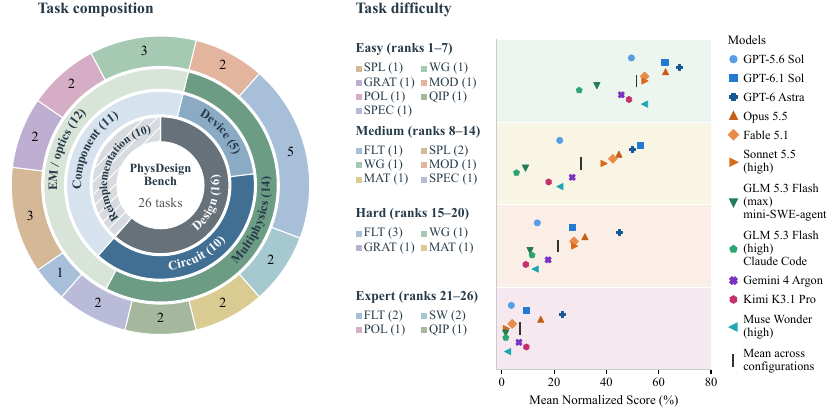}
\caption[Task composition and difficulty.]{\textbf{Task composition and difficulty.}
\textbf{Left:} Distribution of tasks by primary Optica topic~\citep{optica_ocis,optica_topics}, physics scope, design scale, and task family.
\textbf{Right:} Tasks ranked by mean normalized score across 11 model--agent configurations, with invalid submissions assigned a score of zero. Difficulty groups reflect relative agent performance.
\textbf{Topics (abbreviation, count):} \input{fig/benchmark_composition_difficulty_topic_key.tex}}
    \label{fig:topic-coverage}
    \label{fig:task-difficulty-ranges}
    \label{fig:composition-difficulty}
\end{figure}

\paragraph{Task coverage.}
Figure~\ref{fig:topic-coverage} summarizes the tasks by physical regime, design scale, and application. Reimplementation focuses on individual components evaluated through electromagnetic simulation. Design extends from individual components to complete circuits, with 14 of its 16 tasks involving coupled optical, electrical, and thermal processes.

Design spaces range from a few geometric parameters to 6{,}000 binary nanohole sites and full-chip layouts with nearly 500 elements. Some tasks require jointly optimizing 180 heater settings or designing control logic for 390{,}624 possible routing requests. Evaluation can span wavelength sweeps and up to 12 fabrication corners. The tasks cover six material platforms with the full taxonomy provided in Appendix~\ref{app:task-taxonomy}.

\subsection{Verification}
\label{sec:verification}

Following~\citep{merrill2026terminalbench}, we verify each task for physical validity, solvability, and evaluation integrity.

\paragraph{Physical validity.}

Evaluators check whether submitted designs satisfy the specified engineering requirements before assigning a performance score. All Design tasks include design-rule checking (DRC), with 15 using an open-source foundry rule deck~\citep{wu2026special}. Layout-versus-schematic (LVS) checks verify optical and electrical connectivity, while additional gates enforce operating limits and numerical simulation validity. Circuit-level tasks can involve approximately 50 checks, covering requirements such as geometry, electrical connections, thermal limits, and simulation convergence. Reimplementation tasks enforce constraints from the source publications alongside numerical validity checks. A submission that fails a required check is ineligible and does not proceed to simulation.

\paragraph{Solvability.}
Every task has at least one reference design verified against its final evaluator. Design tasks use expert reference implementations, while Reimplementation tasks are anchored to published devices and their reported performance. The verification procedure and supporting results are described in Appendix~\ref{app:task-solvability}.

\paragraph{Evaluation integrity.}
Agents cannot inspect or modify the evaluator. Their workspaces and the evaluator run in separate containers, and only the submitted design is transferred for grading. The evaluator rebuilds the artifact and independently recomputes performance. Intermediate probes provide feedback during development, but reported scores come exclusively from independent final grading.
Evaluation budgets and reporting metrics are described in Section~\ref{sec:experimental-protocol}.

% Keep the existing benchmark floats before the new experimental section.
% \FloatBarrier

\section{Experimental Setup}
\label{sec:experimental-protocol}

\paragraph{Models and agents.}
We evaluate 11 model--agent configurations: GPT-5.6 Sol, GPT-6.1 Sol, and GPT-6 Astra with Codex CLI; Opus 5.5, Claude Fable 5.1, Sonnet 5.5, Kimi K3.1 Pro, and GLM 5.3 Flash with Claude Code; Gemini 4 Argon and GLM 5.3 Flash with mini-SWE-agent; and Muse Wonder with Muse Code. Models use their default reasoning settings unless otherwise specified.

\paragraph{Evaluation metrics.}
\label{sec:evaluation-metrics}
We evaluate agents using three complementary metrics:
valid submission rate (VSR), engineering pass rate (EPR),
and mean normalized score (MNS). These measure an agent's
ability to produce valid designs, meet engineering
performance targets, and optimize task objectives,
respectively. All metrics are computed from independently
evaluated final submissions.

\textbf{Valid Submission Rate (VSR).}
A submission is valid if it executes successfully and
satisfies all task-specific validity constraints.
For an agent $a$ evaluated on $N$ tasks indexed by
$\mathcal T \in \{1,\ldots,N\}$, we define
\begin{equation}
    \mathrm{VSR}(a)
    = \frac{1}{N}\sum_{\mathcal T=1}^{N}
    \mathbb{I}_{\mathrm{valid}}(a,\mathcal T),
\end{equation}
where $\mathbb{I}_{\mathrm{valid}}$ is one if the
final submission is valid and zero otherwise.

\textbf{Engineering Pass Rate (EPR).}
A valid design does not necessarily achieve its intended
function. Each task therefore specifies a set of
\textit{performance targets} that capture its principal
functional requirements, such as transmission, insertion
loss, or bandwidth. We allow a 5\% tolerance relative
to these targets to account for small deviations from
the nominal engineering requirements. A task is passed
if the final submission is valid and achieves at least
95\% of each specified performance target. We define
\begin{equation}
    \mathrm{EPR}(a)
    = \frac{1}{N}\sum_{\mathcal T=1}^{N}
    \mathbb{I}_{\mathrm{pass}}(a,\mathcal T),
\end{equation}
where $\mathbb{I}_{\mathrm{pass}}$ is one if the
final submission satisfies all pass conditions and
zero otherwise. The task-specific performance targets
are listed in Appendix~\ref{app:task-taxonomy}.

\textbf{Mean Normalized Score (MNS).}
While EPR measures whether a design meets its principal
functional requirements, each task also defines a
continuous objective $J_{\mathcal T}(a)$ to be minimized.
This objective may include additional considerations,
such as chip footprint or resource consumption, that
are not part of the performance targets. Since objectives
differ across tasks, we normalize the objective of each
valid submission as
\begin{equation}
    S_{\mathcal T}(a)
    = \frac{1}{1+J_{\mathcal T}(a)/c_{\mathcal T}},
    \qquad
    c_{\mathcal T} =
    \begin{cases}
        J_{\mathrm{scale},\mathcal T}, & \text{Design},\\
        J_{\mathrm{baseline},\mathcal T}, &
        \text{Reimplementation}.
    \end{cases}
    \label{eq:reporting-score}
\end{equation}
For Design tasks, $c_{\mathcal T}$ is a fixed
engineering scale; for Reimplementation, it is
the published reference objective. A score of $0.5$
corresponds to the respective reference value, with
higher scores indicating better performance.
We aggregate scores across tasks and assign zero to invalid final submissions.
\begin{equation}
    \mathrm{MNS}(a)
    = \frac{1}{N}\sum_{\mathcal T=1}^{N}
    S_{\mathcal T}(a),
\end{equation}

\section{Results}
\label{sec:results-main}

\subsection{Overall performance}

Figure~\ref{fig:overall-score} summarizes performance across all 26 tasks, with full results reported in Table~\ref{tab:full-results}. GPT-6 Astra achieves the highest mean normalized score (0.474), followed by Opus 5.5 (0.397) and GPT-6.1 Sol (0.395). Performance varies more widely on Design tasks, where scores range from 0.061 to 0.510, than on Reimplementation (0.229 to 0.417). Rankings also change between the two families: Opus 5.5 outperforms GPT-6.1 Sol on Design (0.429 vs.\ 0.393), but the ordering reverses on Reimplementation (0.347 vs.\ 0.399).

Engineering pass rates tell a different story. Despite their differences in normalized score, GPT-6 Astra, Opus 5.5, and GPT-6.1 Sol each pass only five of the 26 tasks (19.2\%). On Design, pass rates range from 6.2\% to 12.5\% across configurations, compared with 10\% to 30\% on Reimplementation.

Figure~\ref{fig:task-outcomes-activities} (top) helps explain these low pass rates by distinguishing invalid submissions from valid designs that fall short of the performance targets. On Design tasks, the number of valid final submissions ranges from 2 to 16 across configurations. GPT-6 Astra produces valid designs on all 16 tasks, yet only two meet the engineering pass criteria. Similarly, Opus 5.5 and GPT-6.1 Sol produce 15 and 14 valid designs, respectively, but each passes only two tasks. On Reimplementation tasks, all final submissions are valid, while only one to three tasks are passed per configuration.

\begin{figure}[!htbp]
    \centering
    \includegraphics[width=\linewidth]{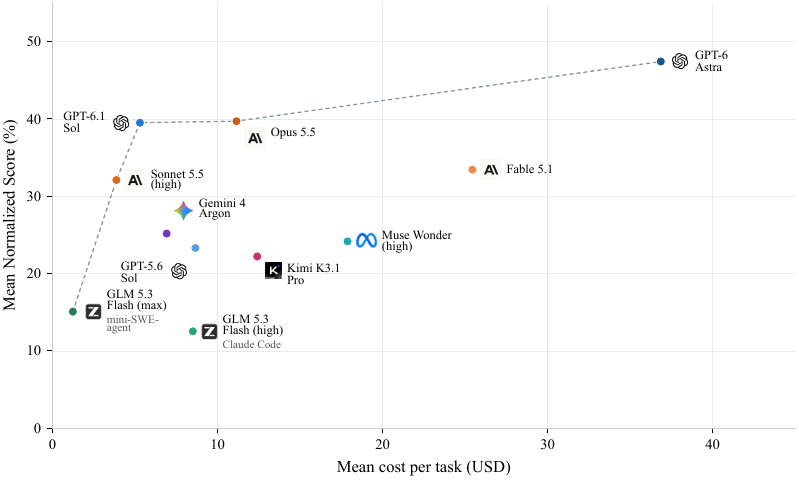}
    \par\vspace{0.5em}
    \includegraphics[width=\linewidth]{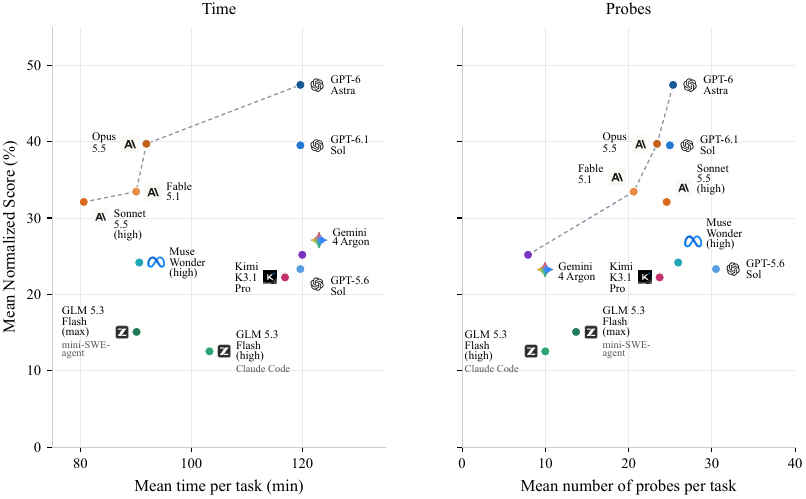}
    \caption{\textbf{Overall performance versus resource use.}
    Mean normalized score ($\times 100$) across all 26 tasks plotted against mean inference cost per task (top), agent execution time (bottom left), and mean evaluator probes per task (bottom right). Dashed lines show the Pareto frontiers of the evaluated configurations. Inference costs exclude simulation and infrastructure, while execution time excludes setup and independent final grading. Failed probes count toward resource use, but final grading does not consume a probe.}
    \label{fig:overall-resource-frontiers}
\end{figure}

The leading configurations also differ in resource use (Figure~\ref{fig:overall-resource-frontiers}). The Pareto frontiers show how the trade-offs change depending on whether inference cost, execution time, or evaluator probes are the limiting resource. GPT-6.1 Sol and Opus 5.5 achieve similar overall scores, but the former costs approximately \$5 per task in inference, compared with \$11 for the latter. Opus 5.5, however, completes tasks in 92 minutes on average, compared with nearly 120 minutes for GPT-6.1 Sol. GPT-6 Astra achieves the highest score at approximately \$37 per task and nearly the full time budget. Probe use introduces another consideration: Gemini 4 Argon averages fewer than eight probes per task despite consuming nearly the full time budget, compared with roughly 23 for Opus 5.5 and 25 for GPT-6 Astra. Additional comparisons by task family, including inference cost and token consumption, are provided in Figures~\ref{fig:score-vs-cost} and~\ref{fig:score-vs-tokens}. Figure~\ref{fig:budget-progress} shows how performance develops over time and evaluator probes.

\subsection{Performance across tasks}
\label{sec:task-performance}

\begin{figure}[t!]
    \centering
    \includegraphics[width=\linewidth]{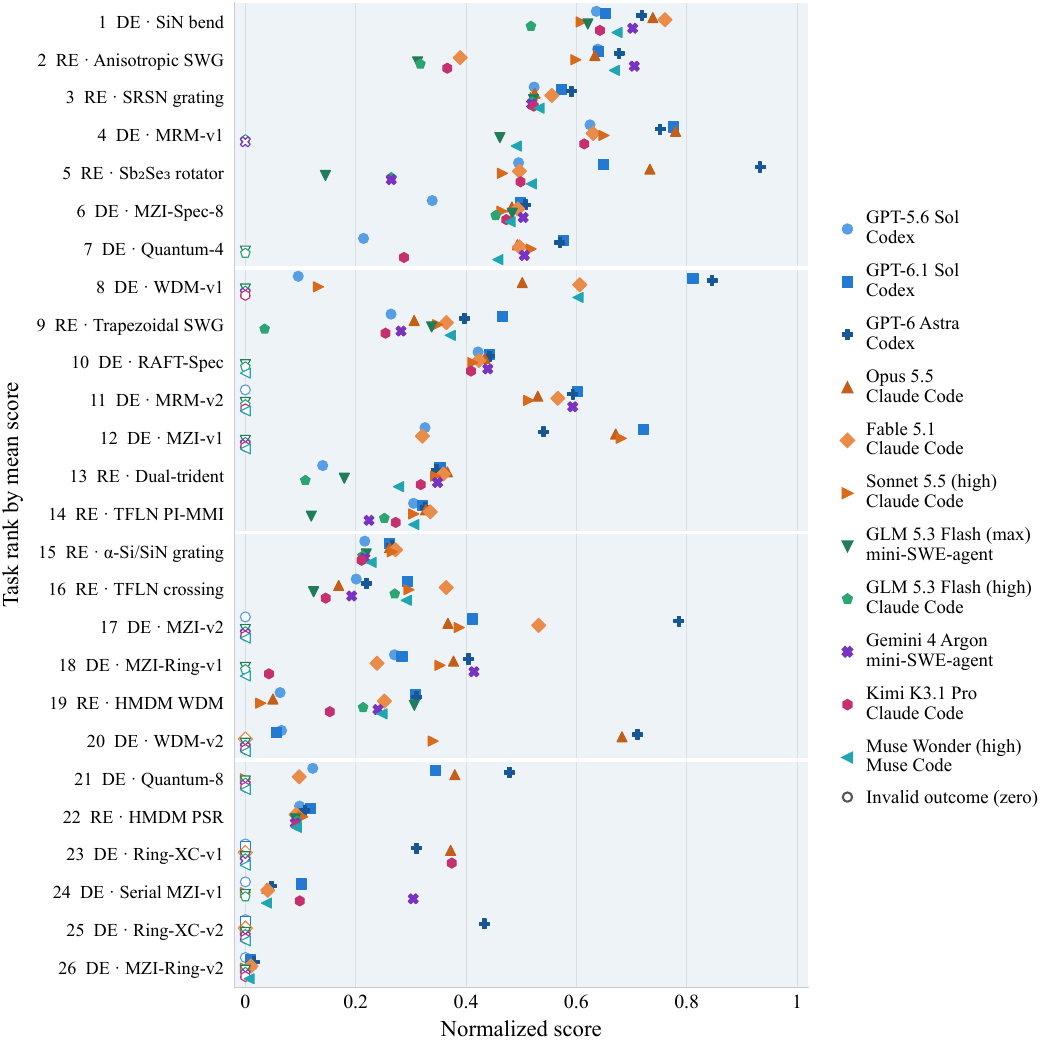}
    \caption{\textbf{Task-level performance.}
The 26 tasks are ranked by mean normalized score across 11 model--agent configurations. DE and RE denote Design and Reimplementation tasks. Each marker represents one configuration, with hollow markers indicating invalid final submissions (score zero). Colors and shapes distinguish configurations. Scores are based on independently evaluated final submissions. Markers are slightly offset vertically to reduce overlap.}
    \label{fig:task-performance}
\end{figure}

The task-level results show that strong aggregate performance does not extend uniformly across problems (Figure~\ref{fig:task-performance}). Even GPT-6 Astra, the leading configuration overall, achieves a Design score of at least 0.5 on nine of the 16 tasks, and at least 0.75 on only three. Rankings also change between closely related problems. For example, GPT-6.1 Sol substantially outperforms Opus 5.5 on the four-channel WDM channelizer, while the ordering reverses on the eight-channel version. Although based on individual runs, this illustrates how performance can depend on the specific design problem. Open-weight configurations generally achieve lower scores on Design, while the gap is smaller on Reimplementation. Full benchmark results are reported in Table~\ref{tab:full-results}, with score distributions in Appendix~\ref{app:score-distributions}. Section~\ref{sec:failure-analysis} examines agent trajectories to better understand the difficulties behind these outcomes..

% \FloatBarrier

\section{Analysis}
\label{sec:agent-analysis}

\subsection{Agent ablation}
\label{sec:harness-ablation}

To examine the effect of the agent scaffold, we evaluate GPT-6.1 Sol and Sonnet 5.5 with mini-SWE-agent on all 16 Design tasks, comparing them with their original Codex CLI and Claude Code configurations.

Both scaffolds achieve comparable average performance, but mini-SWE-agent is substantially more token-efficient (Table~\ref{tab:harness-ablation}). Sonnet's mean score increases from 0.316 with Claude Code to 0.341 with mini-SWE-agent, while GPT-6.1 Sol's decreases slightly from 0.393 to 0.379. Meanwhile, mini-SWE-agent reduces token consumption by 38\% for Sonnet and 63\% for GPT-6.1 Sol, with little change in execution time. Despite similar aggregate scores, performance varies across individual tasks, as detailed in Appendix~\ref{app:harness-ablation}.

\begin{table}[htbp]
    \centering
\caption{\textbf{Agent-scaffold ablation on Design tasks.}
Performance and resource use across 16 Design tasks. Scores are based on independently evaluated final submissions, with invalid or ineligible submissions assigned zero. The Valid column reports the number of tasks with eligible submissions from both scaffolds. Score margins show 95\% task-bootstrap intervals. Token counts include input, output, and cached input tokens. Bold indicates the higher score or lower token use within each model.}
    \label{tab:harness-ablation}
    \begingroup
    \definecolor{AblationAccent}{HTML}{245879}
    \definecolor{AblationTint}{HTML}{EDF4F8}
    \small
    \setlength{\tabcolsep}{4pt}
    \renewcommand{\arraystretch}{1.08}
    \begin{tabular}{@{}l*{5}{c}@{}}
\toprule
\textbf{Agent}
 & \makecell{\textbf{Both}\\\textbf{valid}}
 & \makecell{\textbf{Mean Normalized Score}\\(\%)}
 & \makecell{\textbf{Total time}\\(min)}
 & \textbf{\makecell{Total\\probes}}
 & \makecell{\textbf{Tokens}\\(M)} \\
\midrule
\rowcolor{AblationTint}
\multicolumn{6}{@{}l}{\strut\textbf{GPT-6.1 Sol}\enspace\textcolor{black!60}{(medium)}} \\
Codex & 12/16 & $\mathbf{39.3}\,\pm\,14.0$ & 119.7 & 22.1 & 21.22 \\
mini-SWE-agent & 12/16 & $37.9\,\pm\,12.6$ & 117.9 & 24.5 & \textbf{7.86} \\
\addlinespace[3pt]
\rowcolor{AblationTint}
\multicolumn{6}{@{}l}{\strut\textbf{Sonnet 5.5}\enspace\textcolor{black!60}{(high)}} \\
Claude Code & 10/16 & $31.6\,\pm\,12.3$ & 95.4 & 23.0 & 21.87 \\
mini-SWE-agent & 10/16 & $\mathbf{34.1}\,\pm\,13.7$ & 95.6 & 21.2 & \textbf{13.56} \\
\bottomrule
\end{tabular}
    \endgroup
\end{table}

\subsection{Extended budget}
\label{sec:extended-budget}

To assess whether additional resources help agents overcome difficulties in open-ended design, we evaluate GLM 5.3 Flash (max) with mini-SWE-agent on all 16 Design tasks under two budgets: 2 hours with 40 evaluator probes and 4 hours with 60 probes.

The larger budget increases the mean Design normalized score from 0.098 to 0.150 and doubles the number of eligible final submissions from three to six. Performance improves on four tasks and declines on two, while ten tasks still lack an eligible solution under the larger budget. Average probe use increases from 13.25 to 20.56 per task, and inference cost from \$1.82 to \$3.23 (see Appendix~\ref{app:extended-budget} for more details). Additional resources therefore help recover some unsuccessful designs, but most tasks remain unresolved for this configuration.

\subsection{Robustness across repeated runs}
\label{sec:run-robustness}

To assess repeatability, we evaluate Sonnet 5.5 (high) with Claude Code three times across all 26 tasks. The overall mean score across runs is $0.316 \pm 0.011$ (mean $\pm$ standard deviation), with greater variation on Design ($0.310 \pm 0.020$) than Reimplementation ($0.325 \pm 0.006$). The overall mean varies little between runs, while task-level scores show strong consistency in their relative ordering (ICC = 0.93). Nevertheless, individual Design tasks show considerable variation, particularly WDM-v2, WDM-v1, and MZI-Ring-v1, while others consistently fail across all three attempts. Detailed repeatability statistics and task-level results are provided in Appendix~\ref{app:run-robustness}.

\subsection{Agent workflows}
\label{sec:agent-workflows}

\begin{figure}[!t]
    \centering
    \includegraphics[width=\linewidth]{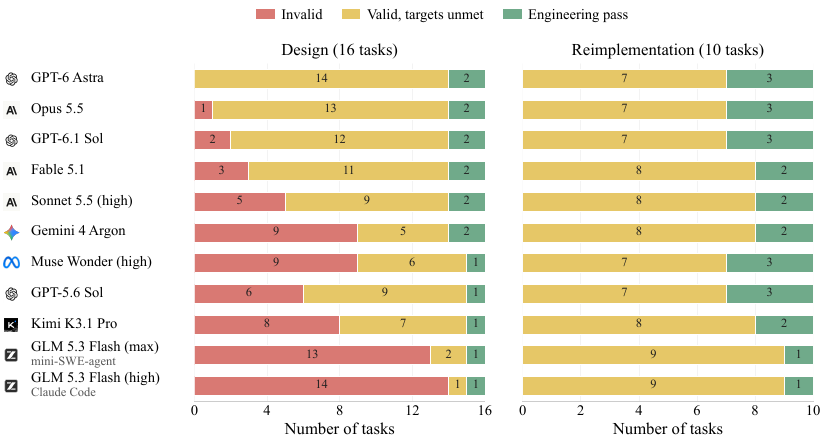}

    \vspace{0.5em}

    \includegraphics[width=\linewidth]{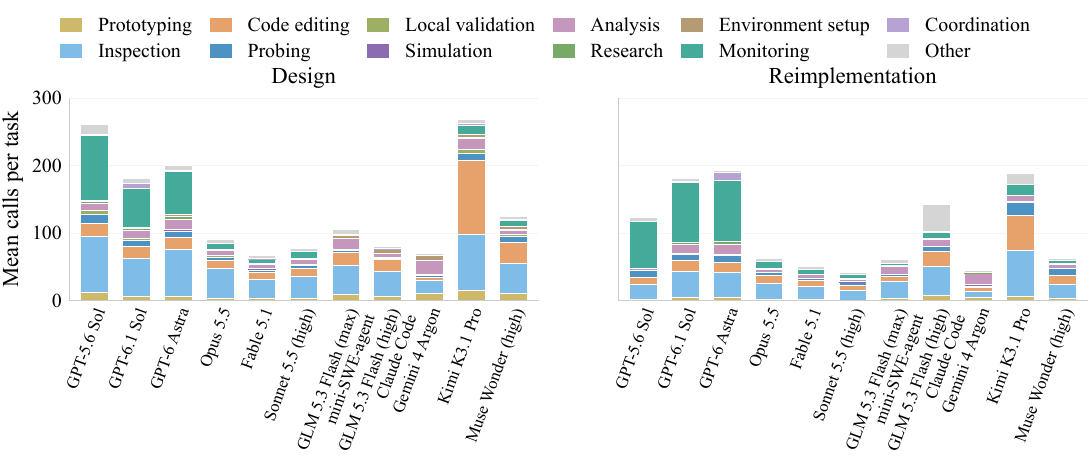}

\caption{\textbf{Task outcomes and agent activities.}
\textbf{Top:} Final-submission outcomes across 16 Design and
10 Reimplementation tasks. Red indicates invalid submissions,
yellow indicates valid submissions that do not meet all
engineering performance targets, and green indicates
engineering passes.
\textbf{Bottom:} Mean activity counts per task for Design
(left) and Reimplementation (right), grouped by activity type.}

    \label{fig:task-outcomes-activities}
\end{figure}

Beyond final performance, we examine how agents interact with the engineering environment. Figure~\ref{fig:task-outcomes-activities} shows final task outcomes (top) and agent activities across 285 recorded trajectories (bottom), with additional tool-use statistics provided in Appendix~\ref{app:tool-use}. The OpenAI configurations running with Codex frequently monitor ongoing processes, such as checking the status of running commands and simulations. Anthropic's Claude Code configurations, in contrast, spend more interactions inspecting the workspace, including reading source files, logs, and simulation outputs. Gemini 4 Argon shows a more balanced distribution of activities, while Muse Wonder primarily inspects and edits code. Among the open-weight models, Kimi K3.1 Pro stands out for its frequent code-editing interactions, particularly on Design tasks.

\subsection{Failure analysis}
\label{sec:failure-analysis}

We analyze failures at two levels: the engineering requirements
violated by final submissions and the decisions made by agents
during their design workflows. The diagnostic taxonomy for the
former is defined in Appendix~\ref{app:diagnostic-taxonomy}.
The latter analysis builds on the trajectory-based failure
analysis of Terminal-Bench~\citep{merrill2026terminalbench},
extending its taxonomy to capture scientific work.
The extended taxonomy and review protocol are described in
Appendix~\ref{app:behavioural-analysis}.

\begin{figure}[!h]
    \centering
    \includegraphics[width=0.9\linewidth]{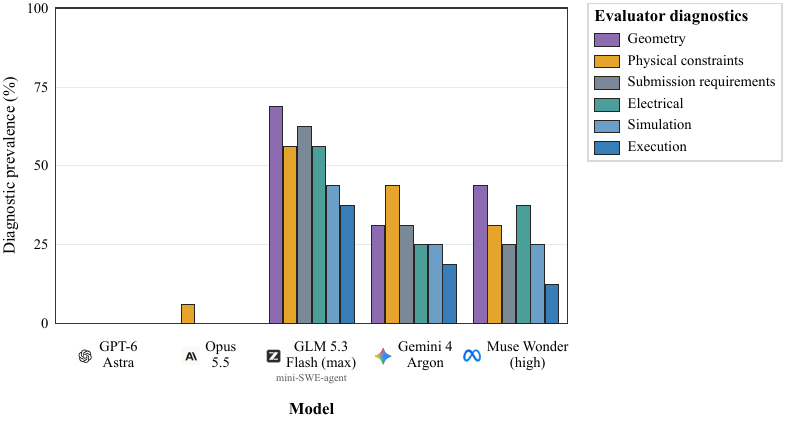}
    \caption{\textbf{Final-submission diagnostics on Design tasks.}
    Prevalence of evaluator diagnostics across five model--agent
    configurations (80 submissions; 16 tasks per configuration).
    Percentages are calculated over all submissions, with
    diagnostics recorded only for invalid submissions.
    Categories may overlap. The full diagnostic taxonomy
    and results for the remaining configurations are provided
    in Appendix~\ref{app:evaluator-failures}.}
    \label{fig:final-error-categories}
\end{figure}

Across all 11 configurations, 70 of 176 Design final submissions (39.8\%) are invalid or ineligible. Figure~\ref{fig:final-error-categories} shows the diagnostic breakdown for five configurations, with the remaining results in Appendix~\ref{app:evaluator-failures}. 

Failure patterns differ considerably across configurations. Among OpenAI models, GPT-6 Astra produces eligible submissions on all 16 tasks, while GPT-6.1 Sol has only two submissions with physical-constraint diagnostics. Anthropic's Opus 5.5 similarly has only one submission with a physical-constraint violation. Gemini 4 Argon encounters physical-constraint violations most frequently (43.8\%), whereas Muse Wonder shows more geometry (43.8\%) and electrical (37.5\%) failures. The open-weight configurations have broader diagnostic profiles: GLM 5.3 Flash (high) encounters geometry violations on 87.5\% of tasks, alongside frequent submission and electrical failures, while Kimi K3.1 Pro is particularly affected by geometry (50\%) and electrical (31.2\%) constraints.

\begin{figure}[!h]
    \centering
    \includegraphics[width=0.9\linewidth]{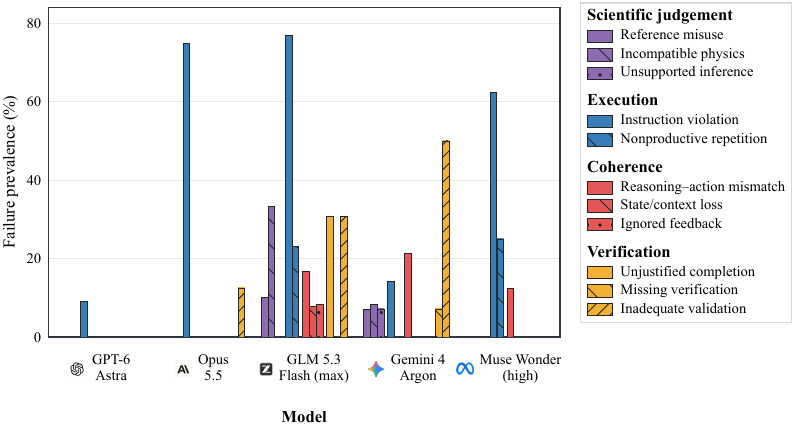}
    \caption{\textbf{Behavioural failures on Design tasks.}
    Prevalence of failure subtypes identified through LLM-based
    trajectory review across 54 Design trajectories:
    11 GPT-6 Astra, 8 Opus 5.5, 13 GLM 5.3 Flash,
    14 Gemini 4 Argon, and 8 Muse Wonder.
    Categories may overlap. The full taxonomy, counting rules,
    and Reimplementation results are provided in
    Appendix~\ref{app:behavioural-analysis}.}
    \label{fig:behavioural-failures-design}
\end{figure}

Not all errors encountered during experimentation persist to
the final submission. GPT-6 Astra, for example, receives
diagnostic feedback on 148 Design probes but produces eligible
final submissions on all 16 tasks
(Appendix~\ref{app:probe-diagnostics}).
To examine how agents respond to such feedback and make design
decisions, we use GPT-5.6 Sol (high) as the judge to review
89 trajectories (54 Design and 35 Reimplementation) against
fourteen behavioural failure subtypes.
Figure~\ref{fig:behavioural-failures-design} reports the
Design results, with the Reimplementation results provided
in Appendix~\ref{app:behavioural-analysis}.

Execution failures are the most frequently identified category
in the trajectory reviews.
Instruction or constraint violations occur in 75\% of reviewed
Opus 5.5 trajectories, 77\% for GLM 5.3 Flash, and 63\% for
Muse Wonder, compared with 9\% for GPT-6 Astra.

The engineering-specific categories reveal different problems.
Half of Gemini 4 Argon's reviewed trajectories contain
inadequate validation, illustrated by the following example:

\begin{leftbar}
\noindent\textbf{Gemini 4 Argon --- Eight-mode projector (V3).}
\textit{The agent describes an isolated thermal calibration as
exact, although the evaluator models all thirty heat-source
zones jointly, including crosstalk. Its official probes stop
at earlier hierarchy checks, so the trajectory does not
establish how this approximation affected final performance.}
\end{leftbar}

GLM 5.3 Flash shows a broader range of failures, including
incompatible physical assumptions (33\%), inadequate validation
(31\%), and nonproductive repetition (23\%).
One trajectory also illustrates a failure to act on an
identified correction:
\newline
\begin{leftbar}
\noindent\textbf{GLM 5.3 Flash --- Four-mode projector (C4).}
\textit{The agent identifies a phase-sign error and calculates
a corrected vector, but subsequently continues using the
original control-generation path. The correction is not
carried through to the implementation, although its
contribution to the final outcome cannot be determined
from the trajectory alone.}
\end{leftbar}

Further examples are provided in
Appendix~\ref{app:behavioural-examples}.

The picture is different on Reimplementation tasks.
Among 24 reviewed trajectories that fail to exceed the
published performance baseline, 21 have no identified
behavioural failure beyond instruction violations.
The full Reimplementation results and failure taxonomy
are provided in Appendix~\ref{app:behavioural-analysis}.

\section{Related work}
\label{sec:related-work}

\begin{table*}[t!]
\centering
\small
\renewcommand{\arraystretch}{1.22}
\setlength{\tabcolsep}{4pt}

\providecommand{\cmark}{\textcolor{green!55!black}{\checkmark}}
\providecommand{\pmark}{\textcolor{orange!85!black}{$\triangle$}}
\providecommand{\xmark}{\textcolor{red!70!black}{$\times$}}
\providecommand{\rwname}[2]{#1\newline{\scriptsize\citep{#2}}}

\definecolor{rwsection}{HTML}{F1F3F6}
\definecolor{rwhighlight}{HTML}{EAF1F8}
\definecolor{rwline}{HTML}{CDD5DE}

\caption{
\textbf{Comparison of scientific and engineering benchmarks.}
\cmark{} denotes a central capability, \pmark{} a capability
present in selected tasks, and \xmark{} a capability
not evaluated as a focus.
\emph{Agentic} denotes using an agentic harness.
\emph{Reimplementation / optimization} covers reproduction of
published results or improvement of quantitative objectives.
\emph{Open-ended design} allows agents to choose the form of
a solution.
\emph{End-to-end engineering design} requires construction of
a complete engineering artifact whose functionality is evaluated.
\emph{Simulation} denotes physical or engineering simulation,
and \emph{Multiphysics} denotes evaluation involving interacting
physical domains.
}
\label{tab:related_work_comparison}

\begin{tabularx}{\textwidth}{
@{}
>{\raggedright\arraybackslash}p{0.285\textwidth}
*{6}{>{\centering\arraybackslash}X}
@{}
}

\toprule[0.9pt]

\textbf{Benchmark} &
\makecell[c]{\scriptsize\textbf{Agentic}} &
\makecell[c]{\scriptsize\textbf{Reimpl. /}\\
             \scriptsize\textbf{optim.}} &
\makecell[c]{\scriptsize\textbf{Open-ended}\\
             \scriptsize\textbf{design}} &
\makecell[c]{\scriptsize\textbf{End-to-end}\\
             \scriptsize\textbf{engineering}\\
             \scriptsize\textbf{design}} &
\makecell[c]{\scriptsize\textbf{Simulation}} &
\makecell[c]{\scriptsize\textbf{Multi-}\\
             \scriptsize\textbf{physics}} \\

\midrule[0.6pt]

\rowcolor{rwsection}
\multicolumn{7}{@{}l}{%
\hspace{5pt}\scriptsize\bfseries
Scientific research and optimization} \\[2pt]

\rwname{ScienceAgentBench}{chen2025scienceagentbench}
& \pmark & \pmark & \pmark & \xmark & \pmark & \xmark \\

\rwname{PhySciBench}{jiang2026deep}
& \pmark & \xmark & \pmark & \xmark & \pmark & \xmark \\

\rwname{ResearchClawBench}{xu2026researchclawbench}
& \cmark & \pmark & \pmark & \xmark & \pmark & \pmark \\

\rwname{NatureBench}{wang2026naturebench}
& \cmark & \cmark & \cmark & \xmark & \pmark & \xmark \\

\rwname{PRBench}{qiu2026prbench}
& \cmark & \cmark & \xmark & \xmark & \pmark & \xmark \\

\rwname{RE-Bench}{wijk2025rebench}
& \cmark & \cmark & \cmark & \xmark & \xmark & \xmark \\

\addlinespace[4pt]

\rowcolor{rwsection}
\multicolumn{7}{@{}l}{%
\hspace{5pt}\scriptsize\bfseries
Physical science and engineering} \\[2pt]

\rwname{FEABench}{mudur2025feabench}
& \pmark & \xmark & \xmark & \cmark & \cmark & \pmark \\

\rwname{PICBench}{wu2025picbench}
& \pmark & \xmark & \xmark & \cmark & \cmark & \xmark \\

\rwname{EngDesign}{guo2026toward}
& \pmark & \pmark & \cmark & \cmark & \cmark & \pmark \\

\rwname{Terminal-Bench-Science}{terminalbenchscience2026}
& \cmark & \pmark & \pmark & \pmark & \pmark & \pmark \\

\rwname{FrontierPhysics}{you2026frontierphysics}
& \cmark & \pmark & \cmark & \xmark & \cmark & \pmark \\

\rwname{Frontier-Eng}{chi2026frontier}
& \cmark & \cmark & \cmark & \cmark & \cmark & \pmark \\

\addlinespace[3pt]
\specialrule{0.6pt}{1pt}{0pt}

\rowcolor{rwhighlight}
\textbf{\benchname{}}
& \cmark & \cmark & \cmark & \cmark & \cmark & \cmark \\

\bottomrule[0.9pt]

\end{tabularx}
\end{table*}

\paragraph{Scientific research and discovery.}
Scientific-agent benchmarks have expanded beyond question answering
to tasks that require computation, analysis, and research workflows.
ScienceAgentBench~\citep{chen2025scienceagentbench} evaluates scientific
data analysis, while Terminal-Bench-Science~\citep{terminalbenchscience2026}
covers computational research tasks across scientific domains, including
physical design and simulation.
PhySciBench~\citep{jiang2026deep} examines reasoning and
research tasks in physics and chemistry.
ResearchClawBench~\citep{xu2026researchclawbench},
NatureBench~\citep{wang2026naturebench}, and
FrontierPhysics~\citep{you2026frontierphysics} extend evaluation
to less prescribed research problems.
In particular, FrontierPhysics evaluates agents on open-ended
physics research involving theoretical and computational work,
experimentation, and scientific findings.
Other benchmarks focus on scientific discovery:
DiscoveryWorld~\citep{jansen2024discoveryworld} and
PhysGym~\citep{chen2026physgym} require agents to investigate
systems through interaction, while
LLM-SRBench~\citep{shojaee2025llm} evaluates equation
discovery from data.

\paragraph{Reimplementation and optimization.}
PaperBench~\citep{starace2025paperbench} and
PRBench~\citep{qiu2026prbench} evaluate agents on reproducing
published research, while RE-Bench~\citep{wijk2025rebench}
and ALE-Bench~\citep{imajuku2026ale} evaluate iterative
improvement against quantitative objectives.
\benchname{} combines these two settings: agents reconstruct
published device designs or develop designs from open-ended
specifications, then improve their performance through
simulation and iterative evaluation.

\paragraph{Engineering and physical design.}
EngDesign~\citep{guo2026toward} and
Frontier-Eng~\citep{chi2026frontier} evaluate engineering design
and optimization using domain-specific simulators and verifiers.
EngiAI~\citep{molinari2026engiai} studies tool-connected
engineering workflows, while FEABench~\citep{mudur2025feabench}
evaluates finite-element modelling from natural-language
specifications.
In photonics, PICBench~\citep{wu2025picbench} evaluates circuit
generation against expert designs, while
PhIDO~\citep{sharma2025ai} and
\citet{kharel2026autonomous} demonstrate agent-based photonic
design workflows.
These efforts complement established computational inverse-design
methods~\citep{molesky2018inverse}.
\benchname{} provides a standardized evaluation of autonomous
physics-based engineering design, spanning open-ended design
and published-device reimplementation, with independently
evaluated engineering artifacts, physical validity checks,
and quantitative performance objectives.
Table~\ref{tab:related_work_comparison} compares the
evaluation settings.

% \FloatBarrier
\section{Limitations}
\label{sec:limitations}

Evaluating complete engineering workflows is expensive. Each task requires substantial agent execution time, model inference, and physical simulation. These costs prevented us from repeating every task across all model--agent configurations. Our robustness analysis shows relatively stable aggregate performance across three complete runs of one configuration (Section~\ref{sec:run-robustness}; Appendix~\ref{app:run-robustness}), although individual task outcomes vary substantially. Most configurations are evaluated only once per task, so small differences in model rankings should be interpreted with caution.

The benchmark contains 26 tasks covering different design objectives, physical systems, and design scales. Although these tasks cover a range of engineering problems, they do not capture the full diversity of physics-based design, particularly manufacturing processes and experimental workflows. Performance on \benchname{} should therefore not be taken as a general measure of engineering capability across all physical domains.

Our evaluation relies on established physics simulators to assess designs under consistent conditions. However, simulation results depend on the underlying physical models and assumptions, and a design that passes the evaluator may still encounter fabrication difficulties or behave differently in physical experiments. Expert review and reference solutions also cannot guarantee that every specification, constraint, or evaluator is free of errors. We commit to correcting identified issues and documenting changes as the benchmark evolves.

Releasing tasks and evaluation code publicly introduces a risk of benchmark contamination. Models may encounter benchmark materials during training. To maintain meaningful evaluation as agent capabilities improve, we will continue expanding \benchname{} with new tasks, periodically revise the benchmark, and explore private evaluation sets to reduce contamination risks.
\section{Discussion}
\label{sec:discussion}

Our evaluation shows that AI agents can carry out many steps of physics-based engineering design, from constructing devices and running simulations to revising designs based on feedback. However, completing these steps does not necessarily lead to a successful engineering outcome. The strongest agents produce valid designs across a large fraction of the benchmark, yet the highest engineering pass rate is only 19.2\%. On Reimplementation tasks, all final submissions are valid, but agents rarely recover the performance reported in the literature.

The challenge is not simply to produce an executable design or run more simulations. Open-ended design requires agents to choose promising structures, identify which parameters matter, interpret physical feedback, and decide what to test next. Our extended-budget experiment shows that additional resources can improve performance without resolving many of these difficulties. The observed failures in verification and feedback use further suggest that progress will require more than stronger coding capabilities or larger execution budgets. As AI agents become more capable, it remains to be seen whether these difficulties can be overcome and how their abilities extend to more demanding engineering problems. We plan to expand \benchname{} accordingly, with larger and more complex workflows involving interacting physical systems and multiple stages of modelling, optimization, and verification.

\section*{Acknowledgements}

B.R. acknowledges funding from the European Union's Horizon Europe research and innovation programme under the Marie Skłodowska-Curie Actions (Grant Agreement No. 101151549). We thank Mohammadreza Farhadinia, and Radmehr Karimian for valuable feedback on the benchmark tasks, and Benjamin Savinson for insightful discussions and constructive feedback on the manuscript. We also thank Flexcompute\footnote{\url{https://flexcompute.com}} for providing part of the simulation credits.

%%%%%%%%%%%%%%%%%%5

% Finish all main-paper material before the references and appendix.
\clearpage
\bibliographystyle{abbrvnat}
\bibliography{references}

% Appendix entry point. 

\clearpage
\appendix

% Appendix contents styling
\startcontents[appendix]

% Title
\noindent
\colorbox{white}{%
    \parbox{\dimexpr\linewidth-2\fboxsep\relax}{%
        \vspace{0.65em}
        \hspace{0.8em}
        {\Large\bfseries\color{black}Table of Contents}
        \vspace{0.65em}
    }%
}

\vspace{1em}

\begingroup
\hypersetup{
    linkcolor=black,
    linktoc=all
}

% Section entries
\titlecontents{section}[2em]
    {\addvspace{0.85em}\bfseries\color{black}}
    {\contentslabel{2em}}
    {}
    {\normalfont\color{black}
     \titlerule*[0.65pc]{.}\contentspage}

% Subsection entries
\titlecontents{subsection}[4.5em]
    {\addvspace{0.2em}\small\color{black}}
    {\contentslabel{2.5em}}
    {}
    {\titlerule*[0.65pc]{.}\contentspage}

\printcontents[appendix]{}{1}{\setcounter{tocdepth}{2}}

\endgroup

\vspace{1.2em}
{\color{black}\noindent\rule{\linewidth}{0.5pt}}

\clearpage

% Full benchmark results

\section{Full benchmark results}
\label{app:full-results}

Table~\ref{tab:full-results} reports the results for all
11 model--agent configurations across Design,
Reimplementation, and the full benchmark.

\begin{table}[H]
    \centering
    \caption{\textbf{Full benchmark results.}
    Mean normalized scores (MNS, $\times100$) with 95\%
    bootstrap confidence intervals, engineering pass rates
    (EPR, \%), and mean resource use per task.
    Resource use includes evaluator probes, execution time,
    and input/output tokens. Bold indicates the highest
    value in each performance column, including ties.}
    \label{tab:full-results}
    \begingroup
    \definecolor{FullResultsHeader}{HTML}{EDF3F7}
    \definecolor{FullResultsStripe}{HTML}{F7F9FA}
    \definecolor{FullResultsAccent}{HTML}{38576D}
    \fontsize{7.2}{9.5}\selectfont
    \setlength{\tabcolsep}{2pt}
    \setlength{\extrarowheight}{2.5pt}
    \renewcommand{\arraystretch}{1.24}
    \newcommand{\fullresultgroup}[1]{{\fontsize{7.4}{9}\selectfont\textbf{#1}}}
    \newcommand{\fullresultmetric}[1]{{\fontsize{6.8}{8}\selectfont#1}}
    \newcommand{\fullresultconfig}[3]{%
        \begin{minipage}[c]{\linewidth}\raggedright\fontsize{7.8}{9.4}\selectfont%
            \raisebox{-0.2ex}{\includegraphics[width=1.05em,height=1.05em,keepaspectratio]{logos/#1.png}}\hspace{0.35em}#2\\[1pt]
            {\fontsize{6.4}{7.8}\selectfont\textcolor{black!60}{#3}}%
        \end{minipage}%
    }
    \arrayrulecolor{FullResultsAccent}
    
% Generated by plotting.export_full_results; do not edit the numbers manually.
\begin{tabularx}{\linewidth}{@{}Xrr@{\hspace{6pt}}rr@{\hspace{6pt}}rr@{\hspace{6pt}}>{\fontsize{6.8}{8.5}\selectfont}r>{\fontsize{6.8}{8.5}\selectfont}r>{\fontsize{6.8}{8.5}\selectfont}r>{\fontsize{6.8}{8.5}\selectfont}r@{}}
\toprule
\rowcolor{FullResultsHeader}
& \multicolumn{2}{c}{\fullresultgroup{Design}}
& \multicolumn{2}{c}{\fullresultgroup{Reimplementation}}
& \multicolumn{2}{c}{\fullresultgroup{Overall}}
& \multicolumn{4}{c}{\fullresultgroup{Mean use per task}} \\
\cmidrule(lr){2-3}
\cmidrule(lr){4-5}
\cmidrule(lr){6-7}
\cmidrule(lr){8-11}
\rowcolor{FullResultsHeader}
\fullresultmetric{\textbf{Model / agent}}
& \multicolumn{1}{c}{\fullresultmetric{\textbf{MNS}}}
& \multicolumn{1}{c}{\fullresultmetric{\textbf{EPR}}}
& \multicolumn{1}{c}{\fullresultmetric{\textbf{MNS}}}
& \multicolumn{1}{c}{\fullresultmetric{\textbf{EPR}}}
& \multicolumn{1}{c}{\fullresultmetric{\textbf{MNS}}}
& \multicolumn{1}{c}{\fullresultmetric{\textbf{EPR}}}
& \multicolumn{1}{c}{\fullresultmetric{\textbf{Probes}}}
& \multicolumn{1}{c}{\fullresultmetric{\makecell{\textbf{Time}\\\textbf{(min)}}}}
& \multicolumn{1}{c}{\fullresultmetric{\makecell{\textbf{Input}\\\textbf{(M)}}}}
& \multicolumn{1}{c}{\fullresultmetric{\makecell{\textbf{Output}\\\textbf{(M)}}}} \\
\midrule

\rowcolor{FullResultsStripe}
\fullresultconfig{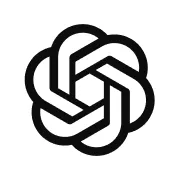}{GPT-6 Astra}{Codex} & $\mathbf{50.99}\,\pm\,11.94$ & $\mathbf{12.5}$ & $\mathbf{41.75}\,\pm\,15.39$ & $\mathbf{30.0}$ & $\mathbf{47.43}\,\pm\,9.03$ & $\mathbf{19.2}$ & 25.35 & 119.66 & 23.245 & 0.178 \\
\fullresultconfig{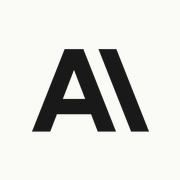}{Opus 5.5}{Claude Code} & $42.88\,\pm\,11.90$ & $\mathbf{12.5}$ & $34.65\,\pm\,13.52$ & $\mathbf{30.0}$ & $39.72\,\pm\,8.81$ & $\mathbf{19.2}$ & 23.42 & 91.91 & 10.491 & 0.115 \\
\rowcolor{FullResultsStripe}
\fullresultconfig{openai}{GPT-6.1 Sol}{Codex} & $39.30\,\pm\,13.95$ & $\mathbf{12.5}$ & $39.86\,\pm\,10.40$ & $\mathbf{30.0}$ & $39.52\,\pm\,9.44$ & $\mathbf{19.2}$ & 24.96 & 119.66 & 20.045 & 0.210 \\
\fullresultconfig{anthropic}{Fable 5.1}{Claude Code} & $32.60\,\pm\,12.85$ & $\mathbf{12.5}$ & $34.80\,\pm\,7.79$ & 20.0 & $33.45\,\pm\,8.45$ & 15.4 & 20.62 & 90.07 & 11.440 & 0.127 \\
\rowcolor{FullResultsStripe}
\fullresultconfig{anthropic}{Sonnet 5.5 (high)}{Claude Code} & $31.64\,\pm\,12.26$ & $\mathbf{12.5}$ & $32.84\,\pm\,10.43$ & 20.0 & $32.10\,\pm\,8.53$ & 15.4 & 24.58 & 80.62 & 14.723 & 0.142 \\
\fullresultconfig{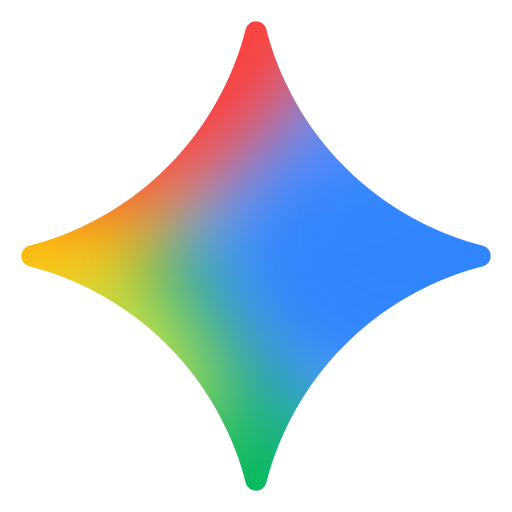}{Gemini 4 Argon}{mini-SWE-agent} & $21.64\,\pm\,13.03$ & $\mathbf{12.5}$ & $30.84\,\pm\,11.39$ & 20.0 & $25.18\,\pm\,9.04$ & 15.4 & 7.92 & 120.00 & 14.527 & 0.247 \\
\rowcolor{FullResultsStripe}
\fullresultconfig{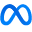}{Muse Wonder (high)}{Muse Code} & $17.19\,\pm\,13.10$ & 6.2 & $35.33\,\pm\,10.37$ & $\mathbf{30.0}$ & $24.17\,\pm\,8.91$ & 15.4 & 25.96 & 90.63 & $29.164$ & $0.210$ \\
\fullresultconfig{openai}{GPT-5.6 Sol}{Codex} & $19.47\,\pm\,10.96$ & 6.2 & $29.45\,\pm\,11.86$ & $\mathbf{30.0}$ & $23.31\,\pm\,8.08$ & 15.4 & 30.50 & 119.66 & 24.084 & 0.092 \\
\rowcolor{FullResultsStripe}
\fullresultconfig{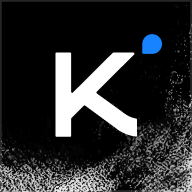}{Kimi K3.1 Pro}{Claude Code} & $18.39\,\pm\,11.98$ & 6.2 & $28.35\,\pm\,8.81$ & 20.0 & $22.22\,\pm\,8.07$ & 11.5 & 23.73 & 116.89 & 24.460 & 0.129 \\
\fullresultconfig{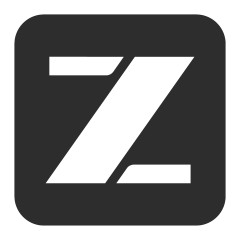}{GLM 5.3 Flash (max)}{mini-SWE-agent} & $9.79\,\pm\,10.79$ & 6.2 & $23.56\,\pm\,8.41$ & 10.0 & $15.08\,\pm\,7.40$ & 7.7 & 13.69 & 90.13 & 31.946 & 0.352 \\
\rowcolor{FullResultsStripe}
\fullresultconfig{zai}{GLM 5.3 Flash (high)}{Claude Code} & $6.07\,\pm\,8.91$ & 6.2 & $22.88\,\pm\,8.38$ & 10.0 & $12.54\,\pm\,6.25$ & 7.7 & 10.00 & 103.27 & 10.443 & 0.231 \\
\bottomrule
\end{tabularx}

    \arrayrulecolor{black}
    \endgroup
\end{table}

\clearpage

% Additional performance and resource analyses

\section{Additional Results}
\label{app:extended-results}

\subsection{Score distributions}
\label{app:score-distributions}

\begin{figure}[H]
    \centering
    \includegraphics[width=\linewidth]{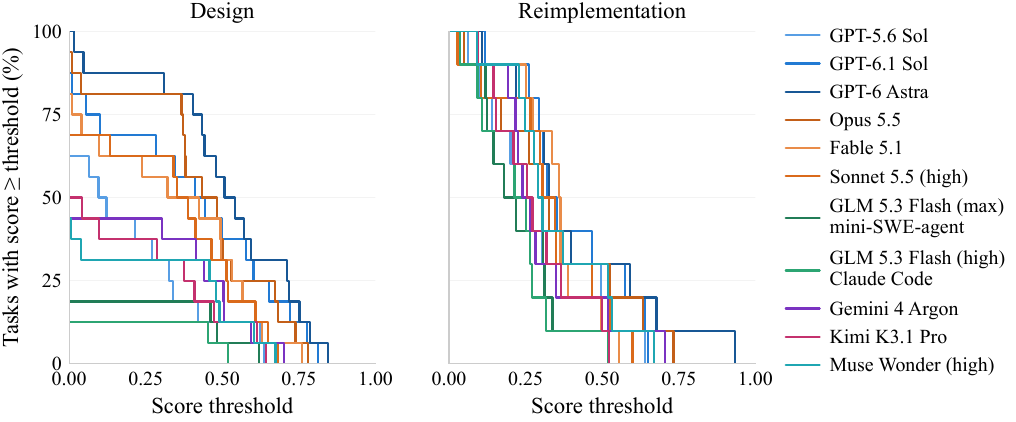}
    \caption{\textbf{Score distributions across tasks.}
    Fraction of Design (left) and Reimplementation (right) tasks reaching each normalized-score threshold. Scores use independently evaluated final submissions, with zero assigned to invalid or ineligible outcomes. For Reimplementation, $S=0.5$ corresponds to the published baseline.}
    \label{fig:score-distributions}
\end{figure}

Figure~\ref{fig:score-distributions} shows the fraction of tasks reaching each normalized-score threshold. High scores remain concentrated in relatively few Design tasks, even for the strongest configurations. GPT-6 Astra reaches $S\geq 0.5$ on nine of 16 Design tasks and $S\geq 0.75$ on only three. Reimplementation scores are more closely clustered, with most submissions falling below the published baseline ($S=0.5$).

\FloatBarrier

\subsection{Performance over time and probes}
\label{app:time-probe-trajectories}

\begin{figure}[H]
    \centering
    \includegraphics[width=0.95\linewidth]{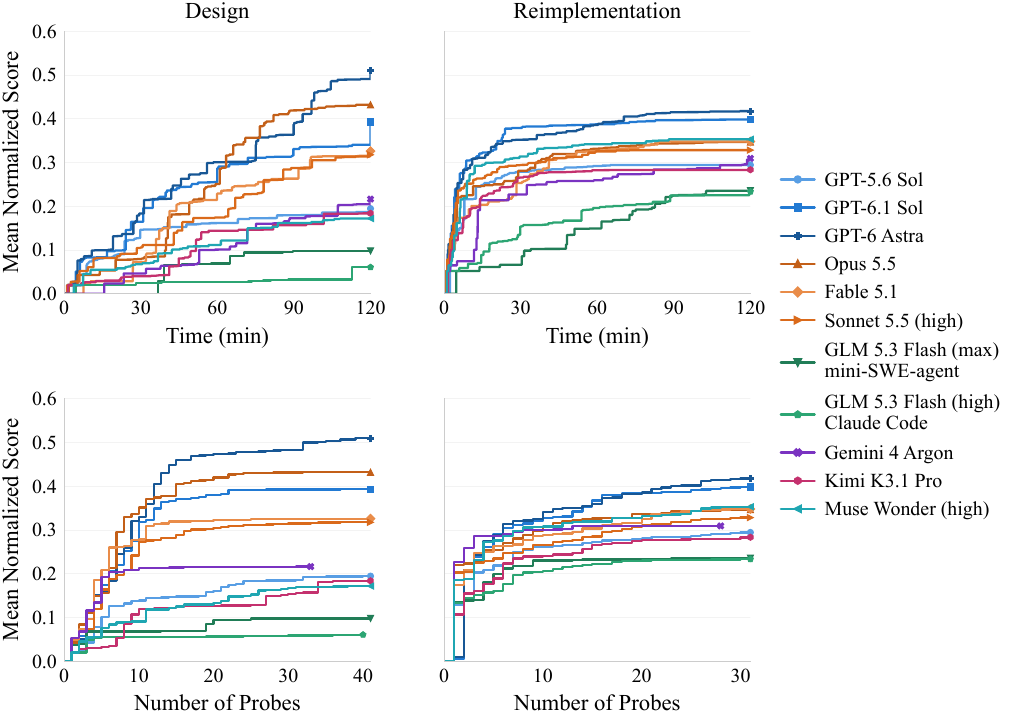}
    \caption{\textbf{Performance over time and probes.}
    Mean best-so-far normalized score against elapsed time (top) and evaluator probes (bottom), separately for Design (left) and Reimplementation (right). Scores are averaged over a fixed task cohort, with zero before the first eligible result. Curves retain their best score after a run ends. Final grading appears at termination without consuming an evaluator probe.}
    \label{fig:budget-progress}
\end{figure}

Figure~\ref{fig:budget-progress} tracks the best evaluated performance reached during each run, including the independent final evaluation at termination. Reimplementation configurations tend to achieve much of their performance early, while Design improvements continue over a larger portion of the available budget. For example, GPT-6.1 Sol reaches a mean Reimplementation score of 0.379 within 30 minutes, compared with 0.399 at termination. On Design, GPT-6 Astra reaches 0.473 after 20 evaluator probes and 0.510 at termination. These trajectories describe the progress of evaluated candidates; reported benchmark scores are based exclusively on final submissions.

\FloatBarrier

\clearpage
\subsection{Tool use}
\label{app:tool-use}

\begin{figure}[htbp]
    \centering
    \includegraphics[width=\linewidth]{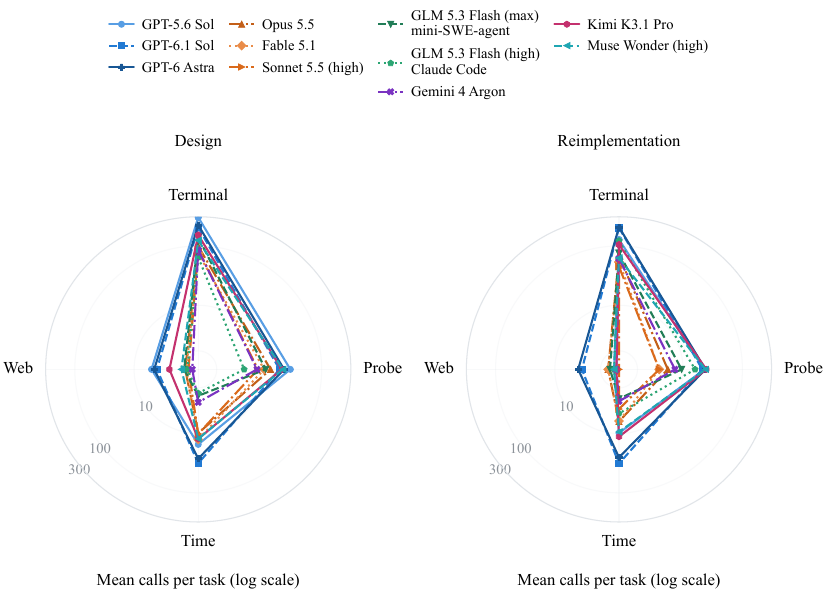}
    \caption{\textbf{Tool-interface use.}
    Mean calls per observed task for Web (search and fetch), Probe (evaluator invocation), Time (\texttt{benchmark\_status}), and Terminal (shell and process interactions). Design is shown on the left and Reimplementation on the right. Both panels use the same $\log(1+x)$ radial scale, with ticks labelled in raw mean calls. Compound calls may contribute to multiple interfaces, and one invocation may launch multiple evaluator probes. Sonnet has published Design trajectories only; GPT-5.6 Sol Design averages use 15 observed trajectories.}
    \label{fig:tool-usage-radar}
\end{figure}

Figure~\ref{fig:tool-usage-radar} compares four tool interfaces: Web, Probe, Time, and Terminal. Terminal calls dominate across configurations, while Web use averages fewer than five calls per task. Tool-use patterns nevertheless vary substantially. On Design, GPT-5.6 Sol averages 293.2 Terminal calls, compared with 75.6 for Fable. On Reimplementation, Kimi averages 25.2 Probe calls, compared with 5.2 for Opus. These counts measure interactions with the environment, not the computational effort or effectiveness of the work performed. The corresponding activity-level analysis is presented in Section~\ref{sec:agent-workflows}.

\FloatBarrier

\subsection{Inference cost}
\label{app:api-cost}

\begin{figure}[htbp]
    \centering
    \includegraphics[width=\linewidth]{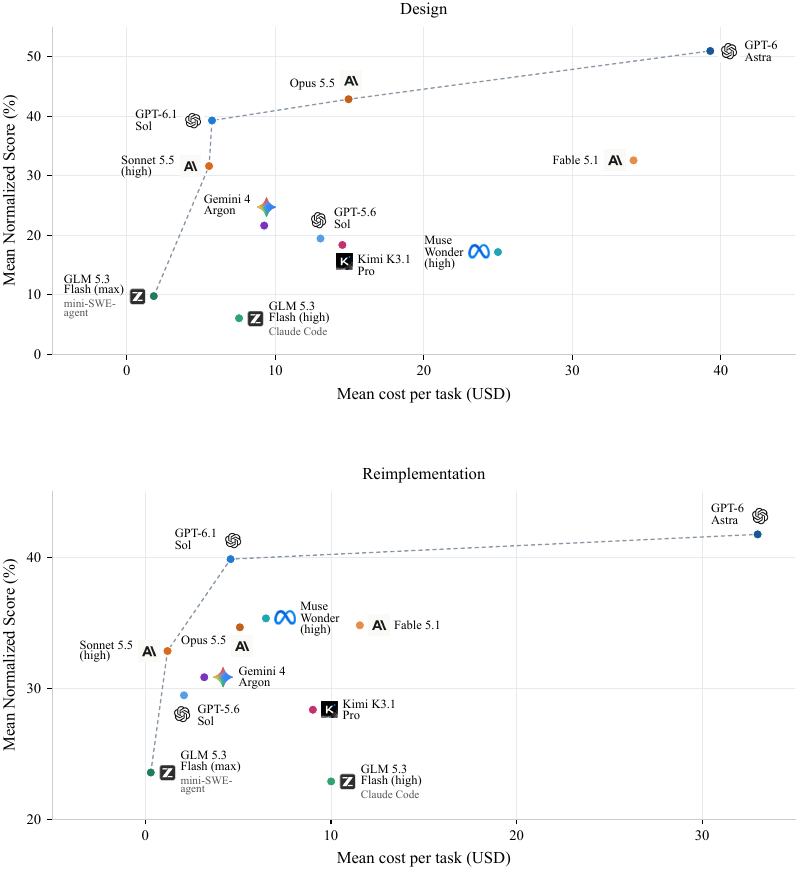}
    \caption{\textbf{Performance versus inference cost.}
    Mean normalized score against mean complete-run inference cost per task for Design (top) and Reimplementation (bottom). Costs include unsuccessful attempts but exclude simulation and infrastructure. Dashed lines indicate Pareto frontiers among evaluated configurations.}
    \label{fig:score-vs-cost}
\end{figure}

Figure~\ref{fig:score-vs-cost} separates the performance--cost trade-offs by task family. GPT-6 Astra leads both families but incurs substantially higher inference costs. On Reimplementation, for example, Astra scores 0.417 at \$32.98 per task, compared with 0.399 at \$4.59 for GPT-6.1 Sol. The Pareto frontiers identify the configurations offering the best observed trade-offs between performance and inference cost. These comparisons exclude simulation and infrastructure expenses.

\FloatBarrier

\clearpage
\subsection{Token use}
\label{app:token-use}

\begin{figure}[H]
    \centering
    \includegraphics[width=\linewidth]{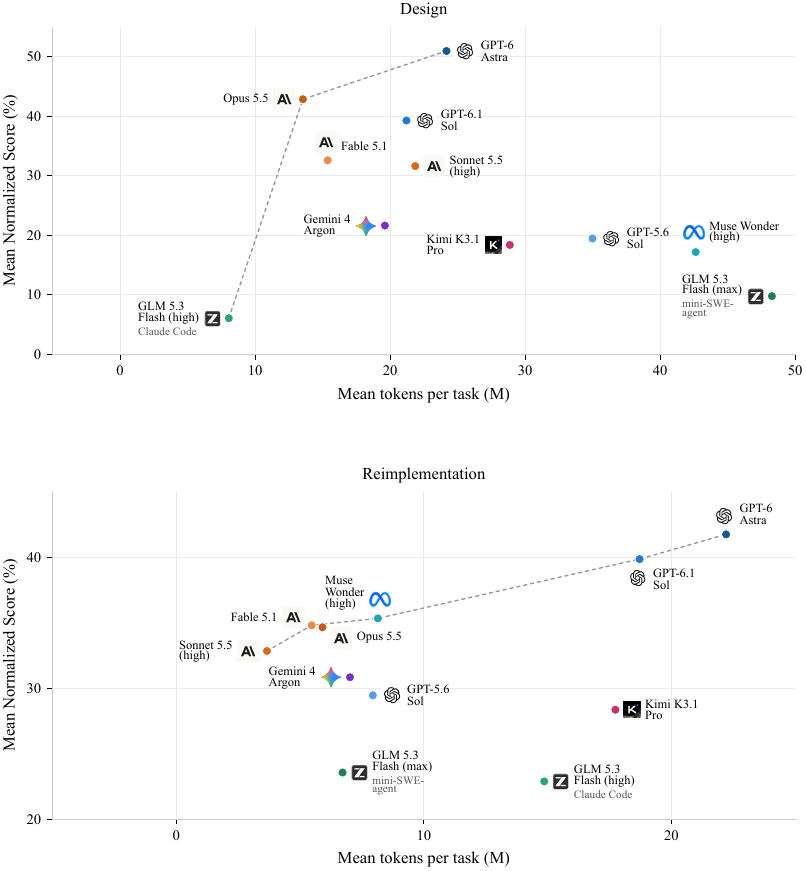}
    \caption{\textbf{Performance versus token use.}
    Mean normalized score ($\times100$) against mean total tokens per task (millions) for Design (top) and Reimplementation (bottom). Token counts include input and output, with cached tokens included in input. Dashed lines indicate Pareto frontiers among evaluated configurations.}
    \label{fig:score-vs-tokens}
\end{figure}

Figure~\ref{fig:score-vs-tokens} compares performance against total token consumption for each task family. Unlike inference cost, token use is independent of provider pricing and captures differences in the volume of model input and output. Counts include unsuccessful attempts and cached input tokens.

\FloatBarrier

\clearpage

% Task construction, coverage, and verification
\section{Benchmark tasks and verification}
\label{app:benchmark-tasks}

\subsection{Task coverage and classification}
\label{app:topic-coverage}

Figure~\ref{fig:topic-coverage} summarizes the benchmark
by task family, design scale, physics regime, and
application topic. We assign each task a primary topic
based on its engineering function, following Optica's
classification schemes~\citep{optica_ocis,optica_topics}.
Table~\ref{tab:topic-coverage} provides the topic,
material platform, operating band, and source
publication for each task.

\begingroup
\small

\definecolor{TopicHeader}{HTML}{EDF3F7}
\definecolor{TopicAccent}{HTML}{38576D}

\setlength{\tabcolsep}{3pt}
\renewcommand{\arraystretch}{1.08}
\setlength{\LTleft}{0pt}
\setlength{\LTright}{0pt}
\setlength{\LTcapwidth}{\textwidth}
\arrayrulecolor{TopicAccent}

\begin{longtable}{@{}
    >{\raggedright\arraybackslash}p{0.30\textwidth}
    >{\raggedright\arraybackslash}p{0.24\textwidth}
    >{\raggedright\arraybackslash}p{0.13\textwidth}
    >{\raggedright\arraybackslash}p{0.11\textwidth}
    >{\raggedright\arraybackslash}p{\dimexpr0.22\textwidth-8\tabcolsep\relax}@{}}

\caption{\textbf{Task topics, platforms, and sources.}
Primary Optica topic, material platform, operating wavelength,
and source publication for each task.
Si: silicon-on-insulator; SiN: silicon nitride;
SRSN: silicon-rich silicon nitride;
TFLN: thin-film lithium niobate.}
\label{tab:topic-coverage}\\

\toprule
\rowcolor{TopicHeader}
\textbf{Task} & \textbf{Optica topic} & \textbf{Platform} &
\textbf{Band (nm)} & \textbf{Source} \\
\midrule
\endfirsthead

\toprule
\rowcolor{TopicHeader}
\textbf{Task} & \textbf{Optica topic} & \textbf{Platform} &
\textbf{Band (nm)} & \textbf{Source} \\
\midrule
\endhead

\bottomrule
\endfoot

\bottomrule
\endlastfoot

\multicolumn{5}{@{}l}{\textbf{Design (16 tasks)}}\\*

\rowcolor{blue!4}
90-degree SiN waveguide bend (v0) & Waveguides & SiN & 1550 & -- \\
Eight-delay MZI spectrometer (v1) & Spectrometers & Si & 1530--1570 & -- \\
\rowcolor{blue!4}
Eight-mode quantum projector (v2) & Quantum information processing & Si & near 1550 & -- \\
Four-mode quantum projector (v1) & Quantum information processing & Si & near 1550 & -- \\
\rowcolor{blue!4}
Four-wavelength, four-port cross-connect (v2) & Optical switching & Si & 1550.3--1554.5 & -- \\
Four-wavelength, two-port cross-connect (v1) & Optical switching & Si & 1550.3--1554.5 & -- \\
\rowcolor{blue!4}
Microresonator modulator (v1) & Modulators & Si & 1310 & -- \\
Microresonator modulator (v2) & Modulators & Si & C band & -- \\
\rowcolor{blue!4}
Microring-assisted Fourier spectrometer (v1) & Spectrometers & Si & 1545--1555 & -- \\
MZI-coupled microring add-drop filter (v1) & Wavelength filtering & Si & near 1550 & -- \\
\rowcolor{blue!4}
MZI-coupled microring add-drop filter (v2) & Wavelength filtering & Si & 1520--1610 & -- \\
Rectangular MZI mesh (v1) & Optical vector--matrix systems & Si & near 1550 & -- \\
\rowcolor{blue!4}
Rectangular MZI mesh (v2) & Optical vector--matrix systems & Si & near 1550 & -- \\
Serial dual-ring-assisted MZI filter (v1) & Wavelength filtering & Si & near 1550 & -- \\
\rowcolor{blue!4}
WDM channelizer (v1) & Wavelength filtering & Si & near 1310 & -- \\
WDM channelizer (v2) & Wavelength filtering & Si & near 1310 & -- \\

\midrule
\multicolumn{5}{@{}l}{\textbf{Reimplementation (10 tasks)}}\\*

\rowcolor{blue!4}
Anisotropic-SWG beamsplitter (v0) & Beam splitters & Si & 1250--1750 & \citet{halir2016} \\
\(\alpha\)-Si/SiN SWG grating coupler (v0) & Diffraction gratings & \(\alpha\)-Si on SiN & 1260--1360 & \citet{asisin2024} \\
\rowcolor{blue!4}
Dual-trident SWG edge coupler (v0) & Waveguides & Si & 1536--1628 & \citet{he2020} \\
HMDM polarization splitter--rotator (v1) & Polarization-selective devices & Si & 1500--1600 & \citet{qiao2025} \\
\rowcolor{blue!4}
HMDM six-channel WDM (v1) & Wavelength filtering & Si & 1500--1600 & \citet{qiao2025} \\
Sb\textsubscript{2}Se\textsubscript{3}/Si polarization rotator (v0) & Polarization-selective devices & Sb\textsubscript{2}Se\textsubscript{3} on Si & 1290--1330 & \citet{sb2se3_2024} \\
\rowcolor{blue!4}
SRSN grating coupler (v0) & Diffraction gratings & SRSN on SiN & 1550 & \citet{srsn2026} \\
TFLN MMI crossing (v0) & Waveguides & TFLN & 1550 & \citet{tflncrossing2023} \\
\rowcolor{blue!4}
TFLN polarization-insensitive MMI (v0) & Beam splitters & TFLN & 1520--1580 & \citet{wu2024} \\
Trapezoidal-SWG 3-dB coupler (v0) & Beam splitters & Si & 1400--1700 & \citet{trapezoidal2023} \\

\end{longtable}

\arrayrulecolor{black}
\endgroup
% \clearpage

\subsection{Task specifications and performance targets}
\label{app:task-taxonomy}

Table~\ref{tab:task-taxonomy-appendix} summarizes all
26 tasks, grouped into Design and Reimplementation.
For each task, we report its design scale, physics
scope, validity requirements, engineering performance
targets, and optimization objective. These correspond
to the three evaluation metrics defined in
Section~\ref{sec:evaluation-metrics}: validity determines
whether a submission is eligible (VSR), performance
targets determine engineering success (EPR), and the
objective $J$ determines normalized performance (MNS).

Design scale distinguishes components (individual
structures), devices (interacting elements), and
circuits (multiple functional blocks). EM denotes
electromagnetic or optical evaluation, while
Multiphysics involves multiple physical domains,
including optical, electrical/RF, and thermal effects.
All objectives $J$ are minimized subject to
task-specific validity constraints. The table
summarizes the evaluation criteria, with complete
definitions provided in the corresponding task
specifications.

\paragraph{Engineering pass criteria.}
For EPR, we assess each final submission against the
selected performance targets listed in
Table~\ref{tab:task-taxonomy-appendix}, allowing a
5\% tolerance ($q=0.95$) relative to the nominal
requirements. All original validity constraints must
be satisfied without relaxation. For a lower bound
$x\ge L$, the passing threshold is $qL$; for a
positive upper bound $x\le U$, it is $U/q$.
Negative upper bounds expressed in dB are relaxed
to $qU$. For an allowed
interval $[L,U]$, we preserve its midpoint and
expand its half-width by $1/q$. These adjustments
are applied in the reported units of each metric,
including logarithmic quantities in dB.

A submission passes only if it is valid and satisfies
every selected (95\% relaxed) performance target under the
task-specific evaluation conditions.

% Per-task validity gates, selected performance targets, and objectives.
% Condensed from "Task eligibility and selected performance targets"
% (All_26_task_eligibility_performance_tables.docx, exported to validity.json).
% Scale/Scope and display names follow the previous taxonomy table.
% Every stated number is copied from the source; Validity rows are abridged.
% Each block: task header row, then Validity / Targets (pass) / Objective rows.

\begingroup
\color{black} % Reset change-tracking colour locally for this table.
% Match the paper's existing blue-grey palette (main.tex).
\definecolor{TaxAccent}{HTML}{245879}
\definecolor{TaxTitleTint}{HTML}{EDF4F8}
\definecolor{TaxTargetTint}{HTML}{F6F9FC}
\definecolor{TaxMeta}{HTML}{526575}
\def\um{\,\textmu m}
\def\umq{\,\textmu m\textsuperscript{2}}
\def\Aum{\,A/\textmu m\textsuperscript{2}}
\def\Ohm{\,$\Omega$}
\def\dg{\textdegree}
% Subtle blue-grey hierarchy: no change-tracking colour.
\def\TaskHead#1#2#3{%
  \rowcolor{TaxTitleTint}%
  \multicolumn{2}{@{}p{\dimexpr\textwidth-2\tabcolsep\relax}@{}}{%
    \rule{0pt}{2.8ex}\textcolor{TaxAccent}{\bfseries #1}%
    \hfill{\normalfont\scriptsize\color{TaxMeta}#2\,\textperiodcentered\,#3}%
    \rule[-0.8ex]{0pt}{1ex}}\\*}
\def\TaskGroup#1{%
  \multicolumn{2}{@{}l}{\rule{0pt}{2.8ex}\textcolor{TaxAccent}{\bfseries #1}}\\*}

\fontsize{7.6}{9.6}\selectfont
\setlength{\tabcolsep}{4pt}
\renewcommand{\arraystretch}{1.08}
\setlength{\LTleft}{0pt}
\setlength{\LTright}{0pt}
\setlength{\LTcapwidth}{\textwidth}
\setlength{\LTpre}{6pt}
\setlength{\LTpost}{2pt}

\begin{longtable}{%
    >{\raggedright\arraybackslash\bfseries}p{0.145\textwidth}
    >{\raggedright\arraybackslash}p{\dimexpr0.855\textwidth-4\tabcolsep\relax}}

\caption{\textbf{Validity requirements, engineering performance
targets, and optimization objectives for all 26 tasks.}
Tasks are grouped into Design and Reimplementation, with
their design scale and physics scope indicated.
$^\dagger$Pass threshold defined for benchmark reporting.
$^\ddagger$Approximate value reported in the original publication.}

\label{tab:task-taxonomy-appendix}\\
\toprule
\endfirsthead

\multicolumn{2}{@{}l}{\footnotesize\itshape Table~\ref{tab:task-taxonomy-appendix} (continued)}\\
\toprule
\endhead

\midrule
\multicolumn{2}{r@{}}{\footnotesize\itshape Continued on next page}\\
\endfoot

\bottomrule
\endlastfoot

%==========================================================================
\TaskGroup{Design (16 tasks)}
\addlinespace[3pt]

% 90sinwaveguide-design-v0
\TaskHead{90-degree SiN waveguide bend (v0)}{Component}{EM}
Description & Design a silicon nitride waveguide bend that guides light between perpendicular ports with minimal transmission loss within a constrained footprint. \\*
Validity & Only the bend polygon is editable; fixed SiN/SiO\textsubscript{2} materials, waveguide thickness, leads, sources, monitors and simulation domain. Waveguide width/height 1.2/0.4\um; join the fixed ports at (0,0) and (12,12)\um. Finite polygon with 12--2000 vertices, no self-intersections, area $\ge$2.88\umq, edge lengths $\ge$10$^{-5}$\um, inside $-1.6\le x\le16.0$\um, $-4.0\le y\le13.6$\um. Continuous contact $\ge$1.08\um{} at input $x=0.001$\um{} and output $y=11.999$\um. Passive $0<T\le1$; $J\ge0$ and $J=-10\log_{10}T$ within 10$^{-12}$. \\*
\rowcolor{TaxTargetTint}
Targets (pass) & Transmission $T\ge96\%$ at 1550\,nm; equivalently $J\le-10\log_{10}(0.96)\approx0.17729$\,dB. \\*
Objective & Minimize $J=-10\log_{10}T$, with $T$ measured at 1550\,nm. The 96\% target is selected for this report; native evaluator eligibility is unchanged. \\
\addlinespace[4pt]

% eight-delay-mzi-spectrometer-design-v1
\TaskHead{Eight-delay MZI spectrometer (v1)}{Circuit}{EM}
Description & Design a passive spectrometer comprising eight unequal-delay Mach--Zehnder interferometers, optimizing optical delays and physical routing for accurate spectral reconstruction. \\*
Validity & Clean full-silicon SiEPIC DRC; optical LVS of tree/MZI/IO topology and port bindings (passive; no electrical LVS). Eight MZIs, seven Y branches and seventeen grating couplers from immutable PDK components; physical connections root 50, tree 20, 4 per MZI, zero virtual. Each MZI differential delay 25--2000\um; arm/route lengths 50--5000\um; bend radius $\ge$8\um; footprint $\le$5000$\times$3000\um, area $\le$15 million\umq. MZI row pitch 254$\pm$1\um; output-fibre pitch 127$\pm$1\um; array $x$-alignment $\le$2\um; arm spans $\le$110\um. Compact-model amplitude singular value $\le$1.0001. \\*
\rowcolor{TaxTargetTint}
Targets (pass) & Reconstruct one-, two- and three-line spectra on the 41-bin grid: mean case error $\le$0.05$^\dagger$; mean of the worst 10\% of case errors (CVaR90) $\le$0.15$^\dagger$. \\*
Objective & 41 bins, 1530--1570\,nm; 1080 reconstruction cases. Case error\,=\,0.8$\times$TV\,+\,0.2$\times$Wasserstein distance/(40\,nm), with TV half the L1 error between normalized power spectra; CVaR90 averages the worst 108 cases. Minimize $J$: reconstruction error plus loss and footprint. Eight distinct delays are requested, but no cross-MZI uniqueness gate is confirmed. \\
\addlinespace[4pt]

% eight-mode-quantum-projector-electrothermal-design-v2
\TaskHead{Eight-mode quantum projector (v2)}{Circuit}{Multiphysics}
Description & Implement a prescribed eight-mode optical projection by jointly designing an interferometer tree, heaters, electrical connections, and quantized control settings. \\*
Validity & Clean SiEPIC DRC; optical (projector-tree) and electrical LVS. Seven combining MZIs (stage population 4/2/1, pitch $\ge$70\um), eight input phase sections, fourteen fixed MMIs; fifteen independent heater controls, thirty pads. Codes 0--255; source voltage $\le$5\,V; heater power $\le$50\,mW; total power $\le$400\,mW; current density $\le$0.10\Aum; heater resistance 1--10,000\Ohm; summed route resistance $\le$60\Ohm; temperatures 284.5--340\,K. One fixed control state across eleven wavelengths. Arm separation $\ge$8\um; mismatch $\le$15\%; local turn $\le$30\dg; pads $\ge$20$\times$20\um, pitch $\ge$40\um. \\*
\rowcolor{TaxTargetTint}
Targets (pass) & Mean projector fidelity $\ge$0.95; worst fidelity $\ge$0.90; worst orthogonal-basis leakage $\le$$-$18\,dB; worst matched insertion loss $\le$3.5\,dB. \\*
Objective & Eleven wavelengths (1545--1555\,nm), one nominal corner and one fixed control/thermal state; mean and worst reductions over the prescribed basis. Minimize $J$: infidelity, leakage/loss excess (penalties start at $-$20\,dB/2.5\,dB) and resource terms. \\
\addlinespace[4pt]

% four-mode-quantum-projector-electrothermal-design-v1
\TaskHead{Four-mode quantum projector (v1)}{Circuit}{Multiphysics}
Description & Implement a prescribed four-mode optical projection using three combining interferometers, jointly optimizing their physical layout and seven heater controls. \\*
Validity & Clean SiEPIC DRC; optical (projector-tree) and electrical LVS. Three combining MZIs, four input phase sections, six fixed MMIs; seven independent heater controls, fourteen pads (left/right/top/bottom 4/2/4/4). Codes 0--255; source voltage $\le$5\,V; heater power $\le$50\,mW; total power $\le$200\,mW; current density $\le$0.10\Aum; heater resistance 1--10,000\Ohm; summed route resistance $\le$60\Ohm; temperatures 284.5--340\,K. One fixed control state and thermal field across eleven wavelengths. Arm separation $\ge$8\um; arm-length mismatch $\le$15\%; local turn $\le$30\dg; pads $\ge$20$\times$20\um, pitch $\ge$40\um. \\*
\rowcolor{TaxTargetTint}
Targets (pass) & Mean projector fidelity $\ge$0.97; worst fidelity $\ge$0.94; worst orthogonal-basis leakage $\le$$-$20\,dB; worst matched insertion loss $\le$2.5\,dB. \\*
Objective & Eleven wavelengths (1545--1555\,nm), one nominal corner and one fixed control/thermal state; mean and worst reductions. Leakage is maximized over the deterministic orthogonal basis, not all superpositions. Minimize $J$: infidelity, leakage/loss excess and resource terms. \\
\addlinespace[4pt]

% four-wavelength-four-port-selective-crossconnect-design-v2
\TaskHead{Four-wavelength, four-port cross-connect (v2)}{Circuit}{Multiphysics}
Description & Design a thermally controlled ring-crossbar circuit and routing compiler to direct four wavelength channels among four ports and identify blocked requests. \\*
Validity & Clean official DRC; optical LVS (crossbar topology, physical ports, zero virtual connections) and electrical LVS (heaters, pads, shared returns). Two-stage fabric with thirty-two rings, thirty-two crossings and four phase shifters; thirty-six independent drives plus four returns (nine heaters each); forty pad openings. Correct classification, factorization and certificates for the 416-request compiler suite. Assigned-ring detuning $\le$0.06\,nm at the nominal corner (or full-span controllability evidence); full-scale phase range $\ge$0.25\,rad. Full-chip area $\le$6 million\umq; heater resistance 1--10,000\Ohm; route resistance $\le$60\Ohm; drive $\le$5\,V; current density $\le$0.10\Aum; total power incl.\ routes $\le$1.20\,W; temperature rise $\le$220\,K; parasitic resonance-shift RMS $\le$0.80\,nm. \\*
\rowcolor{TaxTargetTint}
Targets (pass) & Worst desired-path insertion loss $\le$12\,dB; worst crosstalk $\le$$-$15\,dB; worst passband ripple $\le$2\,dB. \\*
Objective & Targets apply across 2400 scored path-cases and declared local sweeps, not all 416 compiler requests. Crosstalk\,=\,summed power in the three undesired outputs / desired-output power, in dB. Minimize $J$: optical and resource terms. \\
\addlinespace[4pt]

% four-wavelength-two-port-selective-crossconnect-design-v1
\TaskHead{Four-wavelength, two-port cross-connect (v1)}{Circuit}{Multiphysics}
Description & Design a thermally controlled ring-crossbar circuit and routing compiler to direct four wavelength channels between two ports and identify blocked requests. \\*
Validity & Clean official DRC; optical LVS (crossbar topology, physical ports, zero virtual connections) and electrical LVS (heaters, pads, shared returns). Two-stage topology with eight rings, eight qualified crossings and two phase shifters; ten independent drives plus two returns (five heaters each); twelve pad openings. Correct classification/compilation of all 32 requests, including blocked cases; ring controllability at the four wavelengths (cold resonance $\le$1550.2\,nm, full-power resonance $\ge$1554.6\,nm). Footprint $\le$1500$\times$1000\um, area $\le$1.5 million\umq; heater resistance 1--10,000\Ohm; route resistance $\le$60\Ohm; drive $\le$5\,V; current density $\le$0.10\Aum; total power incl.\ routes $\le$400\,mW; temperature rise $\le$220\,K; parasitic resonance-shift RMS $\le$0.80\,nm. Ring radius 7--18\um; gaps 0.12--0.45\um; phase length 60--400\um. \\*
\rowcolor{TaxTargetTint}
Targets (pass) & Worst desired-path insertion loss $\le$12\,dB; worst crosstalk $\le$$-$15\,dB; worst passband ripple $\le$2\,dB. \\*
Objective & 1550.3, 1551.7, 1553.1 and 1554.5\,nm; five corners; nine samples over each $\pm$0.03\,nm passband; 320 scored worst-path cases (blocked/compiler-only requests excluded). Crosstalk\,=\,summed undesired-output / desired-output power, in dB. Minimize $J$: optical and resource terms. \\
\addlinespace[4pt]

% microresonator-modulator-design-v1
\TaskHead{Microresonator modulator (v1)}{Device}{Multiphysics}
Description & Design a depletion-mode silicon microresonator modulator, jointly optimizing resonator geometry, junction layout, RF interconnects, and thermal tuning for optical intensity modulation. \\*
Validity & Clean SiEPIC EBeam DRC; continuous bus and closed side-coupled ring; electrical LVS of isolated heater, GSG, implant and contact nets. One bus, one closed resonator, one depletion junction, one GSG interface, one heater and five pads. Resolvable resonance; bounded heater-servo alignment at all five corners, residual $\le$0.22 of the corner FWHM around the target offset of $-$0.107\,nm. Heater $\le$5\,V/50\,mW; branch power $\le$60\,mW; current density $\le$0.10\Aum; peak temperature $\le$340\,K; ring nonuniformity $\le$8\,K. Heater resistance 1--10,000\Ohm; DC leads $0<R\le60\,\Omega$; three RF routes $\le$20\Ohm{} each, summed length proxy $\le$1000\um. Bus span $\ge$220\um; bus--ring gap $\le$1\um; junction active coverage 35--95\%; heater--ring gap $\le$4\um; five pad openings $\ge$2025\umq. \\*
\rowcolor{TaxTargetTint}
Targets (pass) & Extinction ratio $\ge$2.8\,dB; normalized OMA $\ge$0.08; electro-optic bandwidth $\ge$40\,GHz; on-state insertion loss $\le$8\,dB. \\*
Objective & All four bounds are assessed at five corners (nominal, width-minus, width-plus, cold, hot) after the bounded heater servo. Minimize $J$: optical penalties plus alignment and resource terms. \\
\addlinespace[4pt]

% microresonator-modulator-design-v2
\TaskHead{Microresonator modulator (v2)}{Device}{Multiphysics}
Description & Design a five-channel microresonator array for four-level optical modulation, coordinating segmented depletion junctions, electrical interconnects, and heaters to control modulation quality and crosstalk. \\*
Validity & Clean SiEPIC EBeam DRC; optical LVS (five distinct bus-coupled rings) and electrical LVS (five heater circuits, ten GSG networks, forty pad nets). Five resonators, each with two isolated active segments; MSB-to-LSB active-length ratio 1.5--2.5. Detuning magnitude 0.29$\pm$0.15 FWHM; channel spacing 200$\pm$25\,GHz; alias clearance $\ge$75\,GHz. Simultaneous drive 3\,mA per heater; heater $\le$5\,V/50\,mW; branch power $\le$60\,mW; array power $\le$300\,mW; current density $\le$0.10\Aum; peak temperature $\le$340\,K; ring nonuniformity $\le$8\,K; thermal- and resistance-spread fractions $\le$0.35. Heater resistance 1--10,000\Ohm; DC leads $0<R\le60\,\Omega$; each GSG network $\le$20\Ohm, length proxy $\le$1000\um. PAM4 steps $>$10$^{-7}$ in one direction at all 25 channel-corners. Bus--ring gap $\le$1\um; inter-ring separation $\ge$3\um. \\*
\rowcolor{TaxTargetTint}
Targets (pass) & Normalized OMA $\ge$0.08; electro-optic bandwidth $\ge$48\,GHz; interchannel modulation crosstalk $\le$$-$25\,dB; on-state insertion loss $\le$8\,dB. MSB:LSB segment-ratio target 2:1; PAM4 level-mismatch reference 0.15 (normalizer, not a bound). \\*
Objective & Optical bounds over all 25 channel-corners; alignment, channel-grid and alias checks are nominal-only. Relative mismatch\,=\,std/mean of $|$adjacent level steps$|$. Minimize $J$: mismatch/0.15 (no hinge) plus optical and resource terms. \\
\addlinespace[4pt]

% microring-assisted-fourier-spectrometer-electrothermal-design-v1
\TaskHead{Microring-assisted Fourier spectrometer (v1)}{Circuit}{Multiphysics}
Description & Design a microring-assisted interferometric spectrometer whose thermally scanned measurements enable spectral reconstruction, jointly optimizing optical geometry, heaters, and electrical routing. \\*
Validity & Clean SiEPIC DRC; optical topology, port alignment and physical ring/MZI/root routes; electrical LVS of four pad nets and two heater bodies. One smooth closed one-turn ring, one MZI, two independently wired heaters and four pads. Bend radius $\ge$8\um; ring roundtrip 180--600\um; MZI differential delay 80--800\um; heater width 3--12\um, gap 2--10\um. Nominal 48-state scan: drive $\le$5\,V; current $\le$15\,mA; total scan power $\le$80\,mW; peak temperature $\le$395\,K; ring and MZI control-phase ranges 1.5$\pi$--10$\pi$ (robust corners feed scoring, not an all-corner gate). Metal width 5--15\um, spacing $\ge$8\um; pad opening 30--80\um. \\*
\rowcolor{TaxTargetTint}
Targets (pass) & Ring-assisted Fourier-scan reconstruction on the 81-bin grid: mean case error $\le$0.10$^\dagger$; CVaR90 $\le$0.20$^\dagger$. \\*
Objective & 81 bins, 1545--1555\,nm; 48 scan settings (12 ring $\times$ 4 MZI drives); 480 cases. Case error\,=\,0.60$\times$TV\,+\,0.25$\times|$centroid error$|$/(10\,nm)\,+\,0.15$\times$Wasserstein distance/(10.125\,nm); CVaR90 averages the worst 48 cases. Minimize $J$: reconstruction error plus resource terms. The limits do not guarantee every weak-line detection. \\
\addlinespace[4pt]

% mzi-coupled-microring-add-drop-filter-electrothermal-design-v1
\TaskHead{MZI-coupled microring add-drop filter (v1)}{Device}{Multiphysics}
Description & Design a thermally tunable microring filter with interferometric coupling, coordinating optical geometry and heaters to control resonance wavelength and linewidth. \\*
Validity & Clean SiEPIC EBeam DRC; fixed-cell and route connectivity; electrical LVS of four pad/metal nets and two isolated heater networks with correct phase associations. One compact ring, two buses, three fixed couplers and eight logical routes; two heaters and four isolated pads. Route width 0.45$\pm$0.02\um; bend radius $\ge$10\um; ring roundtrip $\le$450\um; loop span $\le$200\um, aspect ratio $\le$6.5. Heater $\le$5\,V/50\,mW each; total $\le$100\,mW; current density $\le$0.10\Aum; peak temperature $\le$340\,K; fixed-coupler keepouts $\le$300\,K. Cold state 285\,K, heaters off; hot drive 5\,mA per heater. Nine-state continuation: initial selection within $\pm$0.75\,nm; movement $\le$0.30\,nm and $\le$0.45 local FSR. \\*
\rowcolor{TaxTargetTint}
Targets (pass) & Cold/hot resonance point targets 1549/1550\,nm; thermal shift target 1\,nm. Cold/hot FWHM point targets 0.14/0.25\,nm (preferred bands 0.10--0.18/0.22--0.35\,nm). Hot Drop insertion loss $\le$2.5\,dB; Through extinction $\ge$15\,dB; competing-peak rejection $\ge$12\,dB; hot off-resonance Through loss $\le$1\,dB. Theta/phi mean-temperature targets 306/313\,K; hot path-temperature nonuniformity $\le$6\,K. \\*
Objective & Follow one resonance through nine cold-to-hot states; linewidths at their respective states, loss, extinction, rejection and path temperatures in the hot state. Point targets carry no acceptance tolerance. Minimize $J$: linewidth point errors plus optical, thermal and resource terms (6\,K nonuniformity is a soft penalty boundary). \\
\addlinespace[4pt]

% mzi-coupled-microring-add-drop-filter-electrothermal-design-v2
\TaskHead{MZI-coupled microring add-drop filter (v2)}{Device}{Multiphysics}
Description & Design a five-ring add-drop filter and its heater and attenuator controls to disconnect, retune, and reconnect a channel while preserving protected through-channel transmission. \\*
Validity & Clean SiEPIC EBeam DRC; five-ring/two-bus-MZI topology and coupling preflight; electrical LVS of heater/VOA/terminal connectivity. Five closed rings, two wrapped-bus MZIs, eight coupling regions, two VOAs (in R2 and R4), seven heaters and eighteen isolated pads. Ring radius 7--30\um; coupling gap 0.15--0.45\um; interaction length 4--30\um; heater coverage 8--60\%; ring roundtrip 43.98--188.50\um. Heater commands 0--20\,mA; VOA commands 0--3\,V (transparent $\le$0.1\,V, lossy $\ge$0.75\,V). Each heater $\le$5\,V/50\,mW, all heaters $\le$350\,mW; each VOA $\le$3\,V/30\,mA/80\,mW; device $\le$500\,mW; temperature $\le$350\,K; current density $\le$0.10\Aum. Silicon width/space $\ge$0.45/0.15\um; pad openings $\ge$20$\times$20\um. \\*
\rowcolor{TaxTargetTint}
Targets (pass) & Connected centre error $\le$5\,GHz; FWHM target 40\,GHz (selected band 30--55\,GHz); passband ripple $\le$1\,dB; Drop insertion loss $\le$2\,dB; Through isolation $\ge$15\,dB; rejection at $\pm$50\,GHz $\ge$30\,dB; unassigned Drop-peak rejection $\ge$30\,dB; disconnected/retuning suppression $\ge$30\,dB. Protected Through excursion $\le$1\,dB; absolute protected Through loss $\le$1.5\,dB; new off-band Through-notch depth $\le$1.2\,dB. Symmetric Add/Drop responses (no numerical tolerance). \\*
Objective & Four endpoints and three quasi-static paths; A\,=\,196.0\,THz, B\,=\,188.5\,THz; audit 1520--1610\,nm; each target applies only to its endpoint/path state. Protected grid 50\,GHz, anchored at 193.1\,THz. Minimize $J$: endpoint/path optical penalties (width penalty over 30--55\,GHz) and resource terms. \\
\addlinespace[4pt]

% rectangular-mzi-mesh-design-v1
\TaskHead{Rectangular MZI mesh (v1)}{Circuit}{Multiphysics}
Description & Design a programmable four-mode interferometer mesh and heater settings that realize five prescribed optical transformations across wavelength and fabrication variations. \\*
Validity & Clean SiEPIC DRC; optical LVS with fixed-MMI topology/path verification; electrical LVS. Connected Clements mesh: six MZIs, four input phase sections, twelve fixed MMIs; sixteen heater controls, thirty-two pads, no shared returns. Integer codes $q$\,=\,0--255 at 6.5$q$/255\,V; heater resistance 200--2500\Ohm; terminal routes $\le$60\Ohm; thermal samples 293--450\,K. Phase segments $0<L\le10{,}000$\um; total passive route $0<L\le100{,}000$\um; arm separation $\ge$8\um; arm-length mismatch $\le$15\%; local turn $\le$30\dg. Heaters width $\ge$3\um, area $\ge$40\umq; pads $\ge$20$\times$20\um, pitch $\ge$40\um, inset 5--20\um. Transfer singular-value power $\le$1+10$^{-8}$. \\*
\rowcolor{TaxTargetTint}
Targets (pass) & Realize identity, cyclic permutation, DFT4, Hadamard4 and the specified Haar4 transforms; per target, mean fidelity $\ge$0.95$^\dagger$ over its 132 wavelength/corner cases and every case $\ge$0.90$^\dagger$. \\*
Objective & Five target-specific control vectors, each fixed across 12 corners $\times$ 11 wavelengths (1545--1555\,nm): 660 evaluations. Fidelity removes only a common complex gain. Minimize $J$: infidelity plus resource terms; no new loss cutoff. \\
\addlinespace[4pt]

% rectangular-mzi-mesh-design-v2
\TaskHead{Rectangular MZI mesh (v2)}{Circuit}{Multiphysics}
Description & Design a programmable six-mode interferometer mesh, jointly optimizing its physical layout and heater settings to realize five prescribed optical transformations across wavelength and fabrication variations. \\*
Validity & Clean SiEPIC DRC; optical LVS with fixed-MMI topology/path verification; electrical LVS. Connected Clements mesh: fifteen MZIs, six input phase sections, thirty fixed MMIs; thirty-six controls, seventy-two pads, no shared returns. Integer codes $q$\,=\,0--255 at 6.5$q$/255\,V; heater resistance 200--2500\Ohm; terminal routes $\le$60\Ohm; thermal samples 293--450\,K. Phase segments $0<L\le10{,}000$\um; total passive route $0<L\le100{,}000$\um; arm separation $\ge$8\um; mismatch $\le$15\%; local turn $\le$30\dg. Heaters width $\ge$3\um, area $\ge$40\umq; pads $\ge$20$\times$20\um, pitch $\ge$40\um, inset 8--60\um. Transfer singular-value power $\le$1+10$^{-8}$. \\*
\rowcolor{TaxTargetTint}
Targets (pass) & Realize identity, cyclic permutation, DFT6, orthonormal DCT-II6 and the specified Haar6 transforms; per target, mean fidelity $\ge$0.95$^\dagger$ over its 132 wavelength/corner cases and every case $\ge$0.90$^\dagger$. \\*
Objective & Five target-specific control vectors, each fixed across 12 corners $\times$ 11 wavelengths (1545--1555\,nm): 660 evaluations. Fidelity removes only a common complex gain. Minimize $J$: infidelity plus resource terms; no new loss cutoff. \\
\addlinespace[4pt]

% serial-dual-ring-assisted-mzi-chebyshev-elliptic-electrothermal-design-v1
\TaskHead{Serial dual-ring-assisted MZI filter (v1)}{Device}{Multiphysics}
Description & Design a thermally tuned interferometric filter loaded by two serial rings to produce an elliptic spectral response with a flat passband, sharp transitions, and transmission zeros. \\*
Validity & Clean official DRC; optical LVS (serial-ring/MZI topology, centreline paths, coupler geometry) and electrical LVS. Four-port circuit with two continuous MZI arms, two distinct rings and five prescribed couplers; three local heaters, six independent pads. Bend radius $\ge$10\um; ring roundtrips 180--380\um; MZI arms 300--650\um; arm mismatch $\le$160\um. Fixed 5\,mA per heater; heater resistance 1--10,000\Ohm; route $\le$60\Ohm{} per net; heater $\le$5\,V/50\,mW; current density $\le$0.10\Aum; path-equivalent optical temperatures 285--340\,K. Contiguous 3-dB passband containing 1550\,nm; back-port leakage $\le$0.01; passive cold and hot circuits (S-matrix singular value $\le$1.0001). \\*
\rowcolor{TaxTargetTint}
Targets (pass) & Passband centre 1550$\pm$0.05\,nm; FWHM 0.40\,nm $\pm$25\% (0.30--0.50\,nm); ripple $\le$1\,dB; insertion loss $\le$3\,dB. Transmission zeros 1549.55$\pm$0.05 and 1550.45$\pm$0.05\,nm, depth $\ge$25\,dB. Both 10--90\% transition widths $\le$0.15\,nm. Stopband power $\le$0.01; spurious-passband peak power $\le$0.0316227766; complementarity RMS $\le$0.10. \\*
Objective & Hot response (passivity also cold). Centre\,=\,midpoint of contiguous 3-dB crossings; IL from the local peak; ripple over 1550$\pm$0.12\,nm. Zero windows [1549.35,\,1549.75] and [1550.25,\,1550.65]\,nm; stopband/spurious windows [1548,\,1549.3] and [1550.7,\,1552]\,nm. Minimize $J$: selected spectral-feature errors plus resource terms. \\
\addlinespace[4pt]

% wdm-channelizer-electrothermal-design-v1
\TaskHead{WDM channelizer (v1)}{Circuit}{Multiphysics}
Description & Design a four-channel wavelength demultiplexer using heated dual-ring filters, coordinating resonator geometry, coupling gaps, heater placement, and wiring to achieve prescribed channel responses. \\*
Validity & Clean DRC (complete rule coverage); optical (bus--ring-pair--Drop) and electrical LVS. Four dual-ring channels: eight rings, eight heaters and sixteen independent pads, no shared returns. Fixed drive 7.5\,mA per heater; branch voltage $\le$5\,V; heater power $\le$50\,mW; total power incl.\ routes $\le$200\,mW; current density $\le$0.10\Aum; each ring 300$\pm$5\,K; peak $\le$340\,K. Ring radius $\ge$3\um; coupling gaps $\le$0.5\um; paired radii differ by $\ge$0.01\um. Heater resistance 1--10,000\Ohm; routes $\le$60\Ohm; heater area $\ge$40\umq, ring distance $\le$3\um. \\*
\rowcolor{TaxTargetTint}
Targets (pass) & Centre error $\le$0.15\,nm at 1303, 1305, 1307 and 1309\,nm; FWHM point targets 0.15, 0.17, 0.19 and 0.21\,nm; ripple $\le$0.55, 0.60, 0.65 and 0.70\,dB; spurious-peak rejection $\ge$15, 17, 19 and 21\,dB, respectively. Drop insertion loss $\le$3\,dB; crosstalk $\le$$-$18\,dB; Through residual $\le$$-$10\,dB. \\*
Objective & Stabilized Drop peaks; IL/Through residual per peak; crosstalk is absolute Drop transmission at other carriers. Ripple half-window 0.25$\times$target FWHM; spurious guard: carriers $\pm$0.40\,nm. No FWHM acceptance tolerance (0.05\,nm is an error normalizer). Minimize $J$: centre, bandwidth, ripple, rejection and transmission penalties plus resource terms. \\
\addlinespace[4pt]

% wdm-channelizer-electrothermal-design-v2
\TaskHead{WDM channelizer (v2)}{Circuit}{Multiphysics}
Description & Design an eight-channel wavelength demultiplexer using heated dual-ring filters, jointly optimizing spectral responses, electrical routing, and thermal interactions across the integrated circuit. \\*
Validity & Clean DRC (complete rule coverage); optical (bus--ring-pair--Drop) and electrical LVS. Eight channels: sixteen rings, sixteen heaters and thirty-two independent pads, no shared returns. Fixed drive 5\,mA per heater; branch voltage $\le$5\,V; heater power $\le$50\,mW; summed heater power $\le$400\,mW (excluding routes); current density $\le$0.10\Aum; each ring 300$\pm$5\,K; peak $\le$340\,K. Ring radius $\ge$3\um; coupling gaps $\le$0.5\um; paired radii differ by $\ge$0.01\um; heater-arc width $\ge$5.7\um. Heater resistance 1--10,000\Ohm; routes $\le$60\Ohm; heater area $\ge$40\umq, ring distance $\le$3\um. \\*
\rowcolor{TaxTargetTint}
Targets (pass) & Centre error $\le$0.15\,nm at 1303--1317\,nm in 2\,nm steps; FWHM point targets 0.15--0.29\,nm in 0.02\,nm steps; ripple limits 0.55--0.90\,dB in 0.05\,dB steps; spurious rejection $\ge$15, 17, 19, 21, 20, 18, 16 and 14\,dB, respectively. Drop insertion loss $\le$3\,dB; crosstalk $\le$$-$18\,dB; Through residual $\le$$-$10\,dB. \\*
Objective & Stabilized Drop peaks; IL and Through residual at each peak; crosstalk at other assigned carriers. Ripple half-window 0.25$\times$target FWHM; spurious guard $\pm$0.40\,nm around carriers. FWHM values are point targets without acceptance tolerances. Minimize $J$: channel-quality penalties plus resource terms. \\
\addlinespace[3pt]
\midrule

%==========================================================================
\TaskGroup{Reimplementation (10 tasks)}
\addlinespace[3pt]

% anisotropic-swg-beamsplitter-v0
\TaskHead{Anisotropic-SWG beamsplitter (v0)}{Component}{EM}
Description & Reconstruct and optimize an anisotropic subwavelength-grating beamsplitter for broadband balanced splitting, low excess loss, and a quadrature phase relationship between outputs. \\*
Validity & One Si-core mask and one bottom-air-void mask; fixed 220\,nm Si, bottom 60\,nm voids, silica cladding ($n_\mathrm{Si}$\,=\,3.4757, $n_{\mathrm{SiO}_2}$\,=\,1.444, $n_\mathrm{air}$\,=\,1). Starts at $x=0$; length 25--30\um; bounds $x$\,=\,0--30\um, $y$\,=\,$\pm$1.9\um; $\le$20,000 vertices. Si: 100--320 polygons, total area 15--55\umq; voids: 50--420 polygons, total area 0.5--40\umq. Four continuous 1\um{} end landings, width 0.50$\pm$0.02\um, centres $y$\,=\,$\pm$1.0$\pm$0.02\um; both masks mirror-symmetric about $y=0$. Complete finite spectra; positive output powers; raw input-normalized output sum $\le$1.01. \\*
\rowcolor{TaxTargetTint}
Targets (pass) & Excess loss $\le$1\,dB, output-power-ratio imbalance $\le$1\,dB and quadrature phase error $\le$5\dg{} at sampled wavelengths from 1250--1750\,nm. \\*
Objective & 51 wavelengths at 10\,nm spacing; fundamental-TE excitation at the upper input, lower-input response by mirror symmetry. $\mathrm{EL}=-10\log_{10}(T_3+T_4)$; $I=|10\log_{10}(T_3/T_4)|$; $\Delta\phi$\,=\,distance from the nearer $\pm$90\dg{} relative phase. Minimize $J=\max_\lambda\max[\mathrm{EL}/(1\,\mathrm{dB}),\,I/(1\,\mathrm{dB}),\,\Delta\phi/5^\circ]$. \\
\addlinespace[4pt]

% asi-sin-swg-grating-coupler-v0
\TaskHead{$\alpha$-Si/SiN SWG grating coupler (v0)}{Component}{EM}
Description & Reconstruct and optimize an amorphous-silicon subwavelength grating on silicon nitride for efficient fiber-to-chip coupling across a broad optical bandwidth. \\*
Validity & Axis-aligned rectangular $\alpha$-Si apertures in $x$\,=\,0--22\um, $y$\,=\,$\pm$6.55\um; fixed grating width 13.1\um; first trench at $x=0$. 20--40 trench rows of 33 apertures at $y=-6.4,-6.0,\dots,+6.4$\um; lateral gap 25--250\,nm; intervening $\alpha$-Si $\ge$150\,nm. Periods 500--800\,nm; trenches and solid teeth $\ge$200\,nm; no overlaps; global duty cycle 0.30--0.60, each duty within 0.002 of the mean; period/feature tolerance 2\,nm. $\le$1,400 polygons and $\le$6,000 vertices. 101 finite spectral samples, efficiency in (0,1]; positive 1\,dB/3\,dB bandwidths; contiguous 1\,dB passband contains 1310\,nm. \\*
\rowcolor{TaxTargetTint}
Targets (pass) & Coupling loss $\le$1.7\,dB at 1310\,nm; peak-relative 1\,dB bandwidth $\ge$31\,nm and 3\,dB bandwidth $\ge$55\,nm. \\*
Objective & Nominal 3D FDTD, 1260--1360\,nm in 1\,nm steps; TE Gaussian, 9.2\um{} mode-field diameter, 23\dg{} from normal; backward TE\textsubscript{00}. Stack: 220\,nm $\alpha$-Si / 50\,nm SiO\textsubscript{2} / 400\,nm Si\textsubscript{3}N\textsubscript{4} / 4.5\um{} BOX, air cladding. $\mathrm{CE}=(P_{\alpha\text{-Si},\mathrm{TE}_{00}}/P_\mathrm{incident})\times0.95\times0.99$; $L_{1310}=-10\log_{10}\mathrm{CE}(1310\,\mathrm{nm})$. Minimize $J=\max[L_{1310}/(1.7\,\mathrm{dB}),\,(31\,\mathrm{nm})/\mathrm{BW}_{1\,\mathrm{dB}},\,(55\,\mathrm{nm})/\mathrm{BW}_{3\,\mathrm{dB}}]$. \\
\addlinespace[4pt]

% dual-trident-swg-edge-coupler-v0
\TaskHead{Dual-trident SWG edge coupler (v0)}{Component}{EM}
Description & Reconstruct and optimize a three-tip silicon edge coupler that transforms an incident optical mode into a guided waveguide mode with low loss and spectral ripple. \\*
Validity & One Si mask; starts at $x=0$, length 72--78\um; bounds $x$\,=\,0--78\um, $y$\,=\,$\pm$1.65\um. 300--500 polygons, $\le$12,000 vertices; dimensions $\ge$0.13\um; polygon area $\ge$0.018\umq; total area 20--50\umq. Three input tips at $x$\,=\,0.05\um: width 0.14\um, centres $y=-1.34$, 0, $+1.34$\um. Continuous final 2\um{} output strip, width 0.40\um, centre $y=0$ (width/centre tolerances $\pm$0.02\um). 47 finite spectral samples; normalized through power in (0,1.01]. \\*
\rowcolor{TaxTargetTint}
Targets (pass) & Loss at 1550\,nm $\le$0.50\,dB; worst-band loss $\le$0.60\,dB; peak-to-peak loss ripple $\le$0.10\,dB$^\ddagger$. \\*
Objective & Nominal 3D FDTD, 1536--1628\,nm in 2\,nm steps (47 wavelengths); 220\,nm Si in SiO\textsubscript{2} ($n_\mathrm{Si}$\,=\,3.476, $n_{\mathrm{SiO}_2}$\,=\,1.444); TE Gaussian, 6\um{} mode-field diameter; forward TE\textsubscript{0}. $L=-10\log_{10}[\min(P,1)]$, $L_\mathrm{worst}=\max_\lambda L$, $R=\max_\lambda L-\min_\lambda L$. Minimize $J=\max[L_\mathrm{worst}/(0.60\,\mathrm{dB}),\,L_{1550}/(0.50\,\mathrm{dB}),\,R/(0.10\,\mathrm{dB})]$. \\
\addlinespace[4pt]

% hmdm-polarization-splitter-rotator-v1
\TaskHead{HMDM polarization splitter--rotator (v1)}{Component}{EM}
Description & Reconstruct and optimize a binary nanohole pattern that separates TE and TM inputs into distinct outputs while converting the TM input to TE polarization. \\*
Validity & Nanoholes only, on a 90$\times$20 binary grid, 120\,nm pitch, first centre (0.06,\,$-$1.14)\um; region $x$\,=\,0--10.8\um, $y$\,=\,$\pm$1.2\um; 0--1,800 holes at unique sites. Centre error $\le$2\,nm per axis; diameter 100$\pm$1\,nm; 16--64 vertices per hole, $\le$115,200 in total; polygon/circle area ratio 0.90--1.05. One flat mask; fixed 220\,nm Si, 150\,nm shallow etch ($n_\mathrm{Si}/n_{\mathrm{SiO}_2}$\,=\,3.476/1.444). Finite modal powers in [0,1.01]; raw output sum $\le$1.01 per input/wavelength; field decays in [0,10$^{-5}$]. \\*
\rowcolor{TaxTargetTint}
Targets (pass) & Weighted desired-transmission FOM $\ge$0.95$^\ddagger$: TE\textsubscript{0} input $\to$ O1/TE\textsubscript{0} and TM\textsubscript{0} input $\to$ O2/TE\textsubscript{0}. \\*
Objective & Independent TE\textsubscript{0} and TM\textsubscript{0} excitations, 1500--1600\,nm in 10\,nm steps (11 wavelengths); source-normalized forward powers $P=|a|^2$, clipped above 1 only after passivity checks. Minimize $J=1-[0.4\,\mathrm{mean}_\lambda P(\mathrm{O1},\mathrm{TE}_0|\mathrm{TE}_0)+0.6\,\mathrm{mean}_\lambda P(\mathrm{O2},\mathrm{TE}_0|\mathrm{TM}_0)]$. \\
\addlinespace[4pt]

% hmdm-six-channel-wdm-v1
\TaskHead{HMDM six-channel WDM (v1)}{Component}{EM}
Description & Reconstruct and optimize a binary nanohole pattern that separates six wavelength channels into designated output waveguides while limiting insertion loss and interchannel leakage. \\*
Validity & Nanoholes only, on a 60$\times$100 binary grid, 120\,nm pitch, first centre (0.06,\,$-$5.94)\um; region $x$\,=\,0--7.2\um, $y$\,=\,$\pm$6\um; 0--6,000 holes at unique sites. Centre error $\le$2\,nm per axis; diameter 100$\pm$1\,nm; 16--64 vertices per hole, $\le$384,000 in total; polygon/circle area ratio 0.90--1.05. One flat mask; fixed 220\,nm Si with full-depth holes. Finite powers in [0,1.01]; raw six-output sum $\le$1.01 at each wavelength; field decay in [0,10$^{-5}$]. \\*
\rowcolor{TaxTargetTint}
Targets (pass) & Worst desired-channel insertion loss $\le$2.6\,dB; maximum unwanted/desired output-power ratio $\le$0.10 (relative crosstalk $\le$$-$10\,dB) at the six assigned wavelengths. \\*
Objective & One TE\textsubscript{0} excitation; 1500, 1520, 1540, 1560, 1580 and 1600\,nm target O1--O6, respectively; source-normalized forward TE\textsubscript{0} powers $P_j$. $\mathrm{IL}_\mathrm{worst}=\max_k[-10\log_{10}P_k(\lambda_k)]$; $r_\mathrm{max}=\max_k\max_{j\ne k}[P_j(\lambda_k)/P_k(\lambda_k)]$. Minimize $J=\max[\mathrm{IL}_\mathrm{worst}/(2.6\,\mathrm{dB}),\,r_\mathrm{max}/0.10]$. \\
\addlinespace[4pt]

% sb2se3-si-polarization-rotator-v0
\TaskHead{Sb\textsubscript{2}Se\textsubscript{3}/Si polarization rotator (v0)}{Component}{EM}
Description & Reconstruct and optimize a hybrid silicon--Sb\textsubscript{2}Se\textsubscript{3} waveguide that preserves TE polarization in the amorphous state and converts TE to TM in the crystalline state. \\*
Validity & Si, side-Sb\textsubscript{2}Se\textsubscript{3} and top-Sb\textsubscript{2}Se\textsubscript{3} masks; starts at $x=0$, length 24--30\um, $y$\,=\,$\pm$0.35\um; $\le$4,000 vertices. Each polygon: area $\ge$0.1\umq, dimensions $\ge$0.08\um. Si: 1--24 polygons, total 5--16\umq; each Sb\textsubscript{2}Se\textsubscript{3} mask: 1--12 polygons, total 3--8\umq. Nominal 2\um{} Si end landings, width 0.50$\pm$0.02\um, centre $y=0\pm0.02$\um; contacting materials must connect the Si endpoints. Finite nonnegative TE\textsubscript{0}/TM\textsubscript{0} powers, total in (0,1.01]; insertion loss $\le$2.5\,dB in all ten state/wavelength cases. \\*
\rowcolor{TaxTargetTint}
Targets (pass) & PER $>$5\,dB in both states at sampled wavelengths from 1290--1330\,nm; desired-polarization fraction $>10^{0.5}/(1+10^{0.5})\approx75.9747\%$. \\*
Objective & TE\textsubscript{0} excitation in amorphous/crystalline states at 1290, 1300, 1310, 1320 and 1330\,nm; desired outputs TE\textsubscript{0}/TM\textsubscript{0}, respectively. Layers: Si $z$\,=\,0--220\,nm; side/top Sb\textsubscript{2}Se\textsubscript{3} $z$\,=\,0--40/220--260\,nm; indices Si 3.5, SiO\textsubscript{2} 1.444, amorphous 3.305, crystalline 4.121+$i$0.001. $\mathrm{PCE}_a=P_\mathrm{TE}/(P_\mathrm{TE}+P_\mathrm{TM})$, $\mathrm{PCE}_c=P_\mathrm{TM}/(P_\mathrm{TE}+P_\mathrm{TM})$. Minimize $J=\max_{\mathrm{state},\lambda}[-10\log_{10}\max(\mathrm{PCE},2^{-1022})]$\,dB. \\
\addlinespace[4pt]

% srsn-grating-coupler-v0
\TaskHead{SRSN grating coupler (v0)}{Component}{EM}
Description & Reconstruct and optimize a silicon-rich silicon nitride grating above a silicon nitride waveguide to maximize coupling from normally incident light into the guided mode. \\*
Validity & 10--32 positive-area, axis-aligned rectangular SRSN teeth in $x$\,=\,0--30\um, $y$\,=\,$\pm$7.01\um; no other material masks. Tooth span 14$\pm$0.01\um, centre $y=0\pm0.01$\um; lateral edges within 0.01\um{} of $y=\pm7$\um. Tooth widths and gaps $\ge$0.205\um{} (native checks cover inter-tooth gaps, not the initial gap); no overlaps; $\le$4,096 vertices. Gap-first placement; first period starts at $x=0$. Finite efficiency $\eta$ in (0,1]. \\*
\rowcolor{TaxTargetTint}
Targets (pass) & Coupling loss $\le$1.37\,dB at 1550\,nm$^\ddagger$; efficiency $\eta\ge10^{-0.137}\approx72.95\%$. \\*
Objective & Nominal 3D FDTD at 1550\,nm; vertical $E_y$-polarized Gaussian, 10.4\um{} mode-field diameter; backward TE\textsubscript{00} in the 14\um-wide SiN waveguide. Stack: Si substrate / 3.07\um{} SiO\textsubscript{2} BOX / 150\,nm SiN / 180\,nm SiO\textsubscript{2} spacer / 150\,nm SRSN teeth; oxide top 1.39\um{} above the spacer. Minimize $J=-10\log_{10}\eta$ (dB), $\eta=|a_{\mathrm{TE}_{00}}|^2$. \\
\addlinespace[4pt]

% tfln-mmi-crossing-v0
\TaskHead{TFLN MMI crossing (v0)}{Component}{EM}
Description & Reconstruct and optimize thin-film lithium niobate multimode-interference crossings for low-loss transmission along three propagation directions while suppressing leakage into crossing waveguides. \\*
Validity & Four masks (symmetric and asymmetric slab/rib pairs); fixed 500\,nm LN film (250\,nm slab + 250\,nm rib), 67\dg{} sidewalls, 2\um{} buried oxide. Bounds $x,y=\pm20$\um; span 39.9--40.1\um; $\le$8,000 vertices. Each slab: 1--2 polygons, total 1,590--1,610\umq; each rib: 10--40 polygons, total 100--180\umq. Rib half-spans: symmetric 14.25\um, asymmetric 13.0/15.25\um, each $\pm$0.05\um; uniform 1\um{} end landings, width 1.00$\pm$0.02\um. All three directions: field decay in [0,10$^{-5}$]; through power in (0,1]; total $\le$1.01; worst leakage/through crosstalk $\le$$-$50\,dB. \\*
\rowcolor{TaxTargetTint}
Targets (pass) & Worst-direction through loss $\le$0.040\,dB at 1550\,nm$^\ddagger$. \\*
Objective & At 1550\,nm, independently excite the symmetric 45\dg{} and asymmetric 0\dg/90\dg{} directions (mode index 0, target effective index 2.05); measure through power and outward leakage $P_\pm$. Per direction $N=\max(1,P_\mathrm{through}+P_-+P_+)$, $T=P_\mathrm{through}/N$, $\mathrm{IL}=-10\log_{10}T$. Minimize $J=\max(\mathrm{IL}_{45},\mathrm{IL}_{0},\mathrm{IL}_{90})$\,dB. \\
\addlinespace[4pt]

% tfln-polarization-insensitive-mmi-v0
\TaskHead{TFLN polarization-insensitive MMI (v0)}{Component}{EM}
Description & Reconstruct and optimize a lithium niobate multimode-interference splitter that divides either TE or TM input light equally between two outputs with low excess loss. \\*
Validity & One LN rib mask: 1--64 polygons, $\le$4,096 vertices; union forms one connected region. Starts at $x=0$; length 63--75\um; bounds $x$\,=\,0--75\um, $y$\,=\,$\pm$5.6\um; union area 380--720\umq. Straight $\ge$4\um{} end landings: input width 1.4--1.6\um; output widths 1.4--1.6\um, clear gap 2.2--3.2\um, width mismatch $\le$0.03\um. Midpoint width 7--10.5\um. Fixed x-cut LN, angle 67.75\dg; 400\,nm film, 200\,nm etch/slab, 60\dg{} sidewalls, 3\um{} buried oxide, silica cladding. 28 finite positive powers, each $\le$1.01; two-output sum $\le$1.01, without clipping or renormalization. \\*
\rowcolor{TaxTargetTint}
Targets (pass) & Each same-polarization output has loss $\le$3.26\,dB for both TE\textsubscript{0} and TM\textsubscript{0} excitation over the sampled 1520--1580\,nm band. \\*
Objective & 1520--1580\,nm in 10\,nm steps, both polarizations and both outputs (28 powers); mode 0 after sorting four solved modes by TE or TM fraction. $L_\mathrm{worst}=\max_{\lambda,\mathrm{pol},\mathrm{port}}[-10\log_{10}P]$. Minimize $J=\max[L_\mathrm{worst}-10\log_{10}(2),\,0]$\,dB; the ideal equal split has 3.0103\,dB per output. \\
\addlinespace[4pt]

% trapezoidal-swg-3db-coupler-v0
\TaskHead{Trapezoidal-SWG 3-dB coupler (v0)}{Component}{EM}
Description & Reconstruct and optimize a silicon coupler with trapezoidal subwavelength gratings to achieve broadband balanced power splitting for either input while minimizing excess loss. \\*
Validity & One Si-core mask; fixed 220\,nm Si, 3\um{} buried oxide, 1\um{} silica cladding ($n_\mathrm{Si}/n_{\mathrm{SiO}_2}$\,=\,3.476/1.444). Starts at $x=0$; length 20--30\um; bounds $x$\,=\,0--30\um, $y$\,=\,$-$0.4--2.9\um; 20--256 polygons, $\le$16,000 vertices; dimensions $\ge$0.04\um; polygon area $\ge$0.0015\umq; total area 8--40\umq. Four continuous 0.5\um{} end landings: lower/upper centres $y$\,=\,0/2.5\um, each $\pm$0.02\um; widths 0.50$\pm$0.02\um; exactly two disjoint ports throughout each landing. Positive output powers; raw summed output $\le$1.01. \\*
\rowcolor{TaxTargetTint}
Targets (pass) & Output-port loss normalized to collected power 2.5--3.5\,dB; excess loss $\le$0.20\,dB, for both inputs over the sampled 1400--1700\,nm band. \\*
Objective & Both fundamental-TE inputs and both outputs at 31 wavelengths, 1400--1700\,nm in 10\,nm steps. $f_p=P_p/P_\mathrm{total}$, $L_p=-10\log_{10}f_p$, $L_0=3.0103$\,dB, $\mathrm{EL}=-10\log_{10}P_\mathrm{total}$; $q_p=(L_0-L_p)/(L_0-2.5)$ if $L_p<L_0$, else $(L_p-L_0)/(3.5-L_0)$. Minimize $J=\max[q_p,\,\max(\mathrm{EL},0)/0.20]$ over ports, inputs and wavelengths. \\
\end{longtable}
\endgroup

\subsection{Task verification}
\label{app:task-solvability}

We verify each task for physical validity, solvability, and evaluation integrity, as introduced in Section~\ref{sec:verification}. We also test whether the evaluator correctly accepts valid designs and rejects invalid submissions.

\paragraph{Physical validity.}
A submission must pass every applicable validity gate before receiving a performance score. Table~\ref{tab:gate-families} summarizes these checks. All 16 Design tasks use design-rule checking (DRC), with 15 applying a pinned version of the SiEPIC EBeam process design kit through KLayout. These rules enforce fabrication constraints such as minimum widths, spacings, layer overlaps, and device boundaries. The same 15 tasks use layout-versus-schematic (LVS) checks to verify optical and electrical connectivity. For example, one four-port cross-connect submission passed DRC despite two heaters being connected to three metal nets each. It subsequently failed connectivity and actuator-controllability checks, making the design ineligible.

Multiphysics tasks also enforce electrical, thermal, and, where applicable, control requirements. The number of named validity gates ranges from 13 for the eight-delay spectrometer to 51 for the four-port cross-connect. Reimplementation tasks enforce constraints from the source publications, including permitted layers, symmetry, footprint, and operating limits. Numerical checks apply to all tasks, including FDTD field decay below $10^{-5}$, passivity, and simulation convergence.

\begin{table}[t]
\centering
\small
\begingroup
\definecolor{VerificationHeader}{HTML}{EDF3F7}
\definecolor{VerificationAccent}{HTML}{38576D}
\setlength{\tabcolsep}{3pt}
\renewcommand{\arraystretch}{1.08}
\arrayrulecolor{VerificationAccent}

\caption{\textbf{Validity gate families.}
A submission must pass all applicable gates to receive a performance score.}
\label{tab:gate-families}

\begin{tabular}{@{}
    >{\raggedright\arraybackslash}p{0.20\textwidth}
    >{\raggedright\arraybackslash}p{0.50\textwidth}
    >{\raggedright\arraybackslash}p{\dimexpr0.30\textwidth-4\tabcolsep\relax}@{}}
\toprule
\rowcolor{VerificationHeader}
\textbf{Family} & \textbf{What is checked} & \textbf{Applies to} \\
\midrule
\multicolumn{3}{@{}l}{\textbf{Layout and fabrication}}\\
\rowcolor{blue!4}
Layout & Readable layout, required cell hierarchy and ports, device boundary, fixed-cell binding & 16 Design \\
Design rules (DRC) & PDK rule deck: widths, spacings, metal and contact overlaps, floorplan & 16 Design (15 with SiEPIC EBeam) \\
\rowcolor{blue!4}
Connectivity (LVS) & Optical topology; independent electrical nets, contacts, and pads; no shorts, opens, or shared returns & 15 Design \\
Layout contract & Permitted layers, symmetry, footprint, paper-derived limits & 10 Reimplementation \\
\midrule
\multicolumn{3}{@{}l}{\textbf{Operation and simulation}}\\
\rowcolor{blue!4}
Electrical & DC conduction, operating point, bounded drive, current density & 14 multiphysics \\
Thermal & Completed electrothermal solve, temperature limits, heater-to-element mapping & 14 multiphysics \\
\rowcolor{blue!4}
Control & Control states and schedules; compiler schema and determinism & Meshes, projectors, cross-connects, MZI-coupled filter (v2) \\
Numerical validity & Field decay, passivity, model and convergence checks, complete simulation evidence & All 26 \\
\bottomrule
\end{tabular}
\arrayrulecolor{black}
\endgroup
\end{table}

\paragraph{Solvability.}
We consider a task solvable if at least one design passes all gates of its final evaluator under the benchmark conditions. Every Design task has an expert reference design constructed by its authors and evaluated with the task evaluator. Reimplementation tasks are based on published devices, with author reimplementations evaluated where possible.

Agent submissions provide additional evidence of solvability under the evaluation protocol (Table~\ref{tab:solvability-evidence}). Across 11 configurations, all 110 Reimplementation final submissions and 107 of 176 Design final submissions pass the final evaluator. GPT-6 Astra produces an eligible Design submission on all 16 tasks. Thus, every task has at least one solution accepted by its final evaluator. This establishes feasibility under the benchmark conditions, not optimal physical performance.

\begin{table}[t]
\centering
\small
\begingroup
\definecolor{VerificationHeader}{HTML}{EDF3F7}
\definecolor{VerificationAccent}{HTML}{38576D}
\setlength{\tabcolsep}{3pt}
\renewcommand{\arraystretch}{1.08}
\arrayrulecolor{VerificationAccent}

\caption{\textbf{Evidence for task solvability.}
A verified solution passes every applicable validity gate. Agent results cover the 11 evaluated configurations.}
\label{tab:solvability-evidence}

\begin{tabular}{@{}
    >{\raggedright\arraybackslash}p{0.49\textwidth}
    >{\centering\arraybackslash}p{0.17\textwidth}
    >{\centering\arraybackslash}p{0.17\textwidth}
    >{\centering\arraybackslash}p{\dimexpr0.17\textwidth-6\tabcolsep\relax}@{}}
\toprule
\rowcolor{VerificationHeader}
\textbf{Evidence} & \textbf{Design} & \textbf{Reimpl.} & \textbf{Total} \\
\midrule
\multicolumn{4}{@{}l}{\textbf{Reference designs}}\\
\rowcolor{blue!4}
Expert reference design & 16 & -- & 16 \\
Published source design & -- & 10 & 10 \\
\midrule
\multicolumn{4}{@{}l}{\textbf{Final-evaluator verification}}\\
\rowcolor{blue!4}
Tasks with an eligible agent final submission & 16 & 10 & 26 \\
Eligible agent final submissions & 107/176 & 110/110 & 217/286 \\
\midrule
\rowcolor{blue!4}
\textbf{Tasks with a verified solution} & \textbf{16} & \textbf{10} & \textbf{26} \\
\bottomrule
\end{tabular}
\arrayrulecolor{black}
\endgroup
\end{table}

\paragraph{Evaluation integrity.}
Agents and evaluators run in separate containers. Evaluator code, simulation credentials, and reference data remain outside the agent workspace. Only the design artifact is transferred for grading; agent-generated metrics, plots, spectra, and scores are ignored. The final evaluator rebuilds the submission using fixed simulation settings, recomputes all score-bearing quantities, and records the submission hash and evaluator image digest. Intermediate probes use a separate evaluator service with an external budget ledger, and their results are never reused for final scoring.

\paragraph{Evaluator consistency.}
We test evaluators using reference designs, deliberately invalid submissions, and independent objective recomputation. Regression tests cover forged metrics, non-passive responses, invalid geometry, shorted or open heater nets, asymmetric masks, and mismatched artifact hashes. These submissions are rejected by the relevant validity checks. For retained reference evaluations, we also recompute objectives from raw simulation outputs; for the cross-connect tasks, this includes independently resampling thermal fields, with exact agreement in $J$.

We additionally tested two task variants that explicitly encouraged agents to attack the evaluator instead of solving the design problem. In the completed runs, agents declined or stopped without attempting an attack. These tests revealed no exploit but are too limited to establish adversarial robustness, and are excluded from the reported benchmark results.

\clearpage

% Agent ablation and repeat-run robustness

\section{Agent-scaffold ablation}
\label{app:harness-ablation}

We compare GPT-6.1 Sol (medium) using Codex CLI and mini-SWE-agent, and Sonnet 5.5 (high) using Claude Code and mini-SWE-agent, across all 16 Design tasks. Each configuration is evaluated once per task, giving 64 runs in total, with a 120-minute working budget and up to 40 evaluator probes per run. GPT-6.1 Sol produces eligible final submissions on 14 tasks with Codex CLI and 13 with mini-SWE-agent, while Sonnet succeeds on 11 tasks with Claude Code and 12 with mini-SWE-agent.

\subsection{Performance across tasks}
\label{app:harness-ablation-tasks}

Tables~\ref{tab:harness-ablation-gpt-tasks} and~\ref{tab:harness-ablation-sonnet-tasks} show how changing the scaffold affects individual task outcomes. With mini-SWE-agent, Sonnet improves on eight tasks and declines on five, while GPT-6.1 Sol improves on four and declines on eleven. These variations are considerably larger than the differences in aggregate scores reported in Section~\ref{sec:harness-ablation}. Since each configuration was evaluated once per task, the comparisons also include run-to-run variation.

\begin{table}[htbp]
    \centering
    \caption{\textbf{GPT-6.1 Sol (medium) on Design tasks.}
    Final-submission scores with Codex CLI and mini-SWE-agent. $\Delta S$ is mini-SWE-agent minus Codex CLI; invalid or ineligible submissions receive zero.}
    \label{tab:harness-ablation-gpt-tasks}
    \small
    
\begin{tabular}{lrrr}
\toprule
\rowcolor{TableHeader}
\textbf{Task} & \textbf{Codex CLI} & \textbf{mini-SWE-agent} & $\boldsymbol{\Delta S}$ \\
\midrule
90SiN Bend-v0      & 0.653 & 0.596 & \cellcolor{LossLight}-0.057 \\
MZI-Spec-8         & 0.498 & 0.494 & \cellcolor{LossLight}-0.004 \\
Quantum-4          & 0.577 & 0.529 & \cellcolor{LossLight}-0.048 \\
Quantum-8          & 0.345 & 0.318 & \cellcolor{LossLight}-0.026 \\
Ring-XC-v1         & 0.000 & 0.459 & \cellcolor{GainStrong}+0.459 \\
Ring-XC-v2         & 0.000 & 0.000 & +0.000 \\
Microresonator-v1  & 0.776 & 0.609 & \cellcolor{LossMedium}-0.167 \\
Microresonator-v2  & 0.602 & 0.582 & \cellcolor{LossLight}-0.020 \\
RAFT-Spec          & 0.443 & 0.437 & \cellcolor{LossLight}-0.006 \\
MZI-Ring-v1        & 0.284 & 0.333 & \cellcolor{GainLight}+0.049 \\
MZI-Ring-v2        & 0.009 & 0.010 & \cellcolor{GainLight}+0.001 \\
MZI-v1             & 0.722 & 0.294 & \cellcolor{LossStrong}-0.428 \\
MZI-v2             & 0.412 & 0.818 & \cellcolor{GainStrong}+0.406 \\
Serial MZI-v1      & 0.102 & 0.000 & \cellcolor{LossMedium}-0.102 \\
WDM-v1             & 0.811 & 0.584 & \cellcolor{LossMedium}-0.228 \\
WDM-v2             & 0.056 & 0.000 & \cellcolor{LossLight}-0.056 \\
\bottomrule
\end{tabular}

\end{table}

\begin{table}[htbp]
    \centering
    \caption{\textbf{Sonnet 5.5 (high) on Design tasks.}
    Final-submission scores with Claude Code and mini-SWE-agent. $\Delta S$ is mini-SWE-agent minus Claude Code; invalid or ineligible submissions receive zero.}
    \label{tab:harness-ablation-sonnet-tasks}
    \small
    
\begin{tabular}{lrrr}
\toprule
\rowcolor{TableHeader}
\textbf{Task} & \textbf{Claude Code} & \textbf{mini-SWE-agent} & $\boldsymbol{\Delta S}$ \\
\midrule
90SiN Bend-v0      & 0.609 & 0.612 & \cellcolor{GainLight}+0.003 \\
MZI-Spec-8         & 0.465 & 0.487 & \cellcolor{GainLight}+0.022 \\
Quantum-4          & 0.518 & 0.521 & \cellcolor{GainLight}+0.003 \\
Quantum-8          & 0.000 & 0.000 & +0.000 \\
Ring-XC-v1         & 0.000 & 0.000 & +0.000 \\
Ring-XC-v2         & 0.000 & 0.000 & +0.000 \\
Microresonator-v1  & 0.650 & 0.428 & \cellcolor{LossMedium}-0.222 \\
Microresonator-v2  & 0.514 & 0.534 & \cellcolor{GainLight}+0.020 \\
RAFT-Spec          & 0.412 & 0.408 & \cellcolor{LossLight}-0.004 \\
MZI-Ring-v1        & 0.353 & 0.326 & \cellcolor{LossLight}-0.027 \\
MZI-Ring-v2        & 0.000 & 0.009 & \cellcolor{GainLight}+0.009 \\
MZI-v1             & 0.681 & 0.638 & \cellcolor{LossLight}-0.043 \\
MZI-v2             & 0.388 & 0.635 & \cellcolor{GainMedium}+0.248 \\
Serial MZI-v1      & 0.000 & 0.062 & \cellcolor{GainLight}+0.062 \\
WDM-v1             & 0.133 & 0.802 & \cellcolor{GainStrong}+0.669 \\
WDM-v2             & 0.340 & 0.000 & \cellcolor{LossStrong}-0.340 \\
\bottomrule
\end{tabular}

\end{table}

\FloatBarrier

\section{Extended-budget experiment}
\label{app:extended-budget}

We provide additional details and task-level results for
the extended-budget experiment described in
Section~\ref{sec:extended-budget}.
Both budget conditions use GLM 5.3 Flash (max) with
mini-SWE-agent.
Table~\ref{tab:extended-budget-summary} reports the
aggregate results.

\begin{table}[!ht]
    \centering
    \caption{\textbf{Aggregate results of the extended-budget
    experiment.}
    Scores and resource use are averaged over all 16 tasks
    per condition, including unsuccessful runs.
    Eligibility is assessed on the final submission.
    Execution time excludes setup and independent final
    evaluation.
    Budget labels indicate maximum allowances.}
    \label{tab:extended-budget-summary}
    \small
    \setlength{\tabcolsep}{4pt}
    \renewcommand{\arraystretch}{1.12}
    \begin{tabular}{@{}lrrrrrr@{}}
\toprule
\rowcolor{TableHeader}
\textbf{Budget}
 & \makecell{\textbf{Mean normalized}\\\textbf{score} (\%)}
 & \textbf{Eligible}
 & \makecell{\textbf{Time}\\(min)}
 & \textbf{Probes}
 & \makecell{\textbf{Tokens}\\(M)}
 & \makecell{\textbf{Cost}\\(USD)} \\
\midrule
2 h / 40 probes & 9.8  & 3/16 & 88.6  & 13.25 & 48.29 & 1.82 \\
4 h / 60 probes & 15.0 & 6/16 & 182.5 & 20.56 & 88.10 & 3.23 \\
\bottomrule
\end{tabular}
\end{table}

\subsection{Task-level results}
\label{app:extended-budget-tasks}

Table~\ref{tab:extended-budget-tasks} provides the final
normalized scores and evaluator probe counts for individual
tasks under both budget conditions.
RAFT-Spec and MZI-Ring-v1 account for most of the aggregate
score improvement. On 90SiN Bend-v0, the final score
declines despite an increase in evaluator probe use
from five to 60.

\begin{table}[!ht]
    \centering
    \caption{\textbf{Task-level results of the extended-budget
    experiment.}
    $\Delta S$ denotes the difference in final normalized
    score between the extended and standard budget conditions.
    Improvements and declines are highlighted in green
    and red, respectively. Invalid or score-ineligible
    submissions receive a score of zero.}
    \label{tab:extended-budget-tasks}
    \small
    \setlength{\tabcolsep}{5pt}
    \renewcommand{\arraystretch}{1.08}
    \begin{tabular}{@{}lrrrrr@{}}
\toprule
\rowcolor{TableHeader}
\textbf{Task} & \multicolumn{3}{c}{\textbf{Final normalized score}} & \multicolumn{2}{c}{\textbf{Probes used}} \\
\cmidrule(lr){2-4}\cmidrule(lr){5-6}
 & 2 h / 40 & 4 h / 60 & $\Delta S$ & 2 h / 40 & 4 h / 60 \\
\midrule
90SiN Bend-v0 & 0.621 & 0.546 & \cellcolor{LossLight}-0.075 & 5 & 60 \\
MZI-Spec-8 & 0.484 & 0.472 & \cellcolor{LossLight}-0.012 & 40 & 19 \\
Quantum-4 & 0.000 & 0.000 & +0.000 & 7 & 19 \\
Quantum-8 & 0.000 & 0.000 & +0.000 & 1 & 12 \\
Ring-XC-v1 & 0.000 & 0.000 & +0.000 & 22 & 0 \\
Ring-XC-v2 & 0.000 & 0.000 & +0.000 & 9 & 16 \\
Microresonator-v1 & 0.461 & 0.571 & \cellcolor{GainLight}+0.110 & 30 & 50 \\
Microresonator-v2 & 0.000 & 0.000 & +0.000 & 4 & 3 \\
RAFT-Spec & 0.000 & 0.373 & \cellcolor{GainStrong}+0.373 & 20 & 53 \\
MZI-Ring-v1 & 0.000 & 0.370 & \cellcolor{GainStrong}+0.370 & 39 & 54 \\
MZI-Ring-v2 & 0.000 & 0.000 & +0.000 & 1 & 25 \\
MZI-v1 & 0.000 & 0.000 & +0.000 & 1 & 0 \\
MZI-v2 & 0.000 & 0.000 & +0.000 & 8 & 10 \\
Serial MZI-v1 & 0.000 & 0.065 & \cellcolor{GainLight}+0.065 & 5 & 6 \\
WDM-v1 & 0.000 & 0.000 & +0.000 & 18 & 0 \\
WDM-v2 & 0.000 & 0.000 & +0.000 & 2 & 2 \\
\midrule
Mean & 0.098 & 0.150 & +0.052 & 13.25 & 20.56 \\
\bottomrule
\end{tabular}

\end{table}

\FloatBarrier

\section{Repeat-run robustness}
\label{app:run-robustness}

\subsection{Experimental setup}
\label{app:variance-setup}

We evaluate Sonnet 5.5 (high) with Claude Code three times on all 26 tasks, giving 78 task--run observations. All reported scores use independently evaluated final submissions, with zero assigned to invalid or ineligible outcomes.

Table~\ref{tab:variance-cohorts} summarizes the scores and final-submission eligibility across the three runs.

\begin{table}[!htbp]
    \centering
    \caption{\textbf{Performance across repeated runs.}
    Mean normalized scores and final-submission eligibility for each run of Sonnet 5.5 (high) with Claude Code.}
    \label{tab:variance-cohorts}
    
\begin{tabular}{llrr}
\toprule
\rowcolor{TableHeader}
\textbf{Family} & \textbf{Repeat} & \textbf{Final score} & \textbf{Eligible finals} \\
\midrule
Design & 1 & 0.316 & 11/16 \\
Design & 2 & 0.288 & 12/16 \\
Design &  3 & 0.325 & 12/16 \\
\midrule
Reimplementation & 1 & 0.328 & 10/10 \\
Reimplementation & 2 & 0.328 & 10/10 \\
Reimplementation & 3 & 0.317 & 10/10 \\
\bottomrule
\end{tabular}
\end{table}

\subsection{Statistical analysis}
\label{app:variance-methods}

We distinguish variation between complete runs from variation across individual tasks. For $T$ tasks and $K=3$ runs, let $S_{tr}$ denote the score on task $t$ in run $r$, with $\bar S_{t\cdot}$ and $\bar S_{\cdot r}$ denoting the corresponding task and run means. We report three descriptive standard deviations:
\begin{equation}
    \sigma_{\mathrm{repeat}}=\operatorname{sd}_r(\bar S_{\cdot r}),\qquad
    \bar\sigma_{\mathrm{within}}=\frac{1}{T}\sum_t\operatorname{sd}_r(S_{tr}),\qquad
    \sigma_{\mathrm{task}}=\operatorname{sd}_t(\bar S_{t\cdot}).
\end{equation}

These measure variation in aggregate scores between runs, average variation within individual tasks, and differences between task means, respectively. All use sample standard deviations. Table~\ref{tab:variance-summary} reports the results separately for Design and Reimplementation.

\begin{table}[!htbp]
    \centering
    \caption{\textbf{Score variation across repeated runs.}
    Aggregate scores are reported as mean $\pm$ sample SD across three runs. Within-task SD averages the sample SD across tasks; $\bar\rho$ is the mean pairwise Spearman correlation between runs.}
    \label{tab:variance-summary}
    
\begin{tabular}{lrrr}
\toprule
\rowcolor{TableHeader}
\textbf{Task family} & \textbf{Mean score $\pm$ SD} & \textbf{Within-task SD} & \textbf{$\bar\rho$} \\
\midrule
Design & $0.310 \pm 0.020$ & 0.046 & 0.94 \\
Reimplementation & $0.325 \pm 0.006$ & 0.028 & 0.92 \\
\midrule
\textbf{Overall} & $\mathbf{0.316 \pm 0.011}$ & \textbf{0.039} & \textbf{0.94} \\
\bottomrule
\end{tabular}
\end{table}

To assess the consistency of task-level performance across runs, we use the single-measure consistency intraclass correlation, ICC(C,1):
\begin{equation}
    \mathrm{ICC}(C,1)=
    \frac{\mathrm{MS}_{\mathrm{task}}-\mathrm{MS}_{\mathrm{error}}}
    {\mathrm{MS}_{\mathrm{task}}+(K-1)\mathrm{MS}_{\mathrm{error}}}.
\end{equation}
Here $\mathrm{MS}_{\mathrm{task}}$ is the between-task mean square and $\mathrm{MS}_{\mathrm{error}}$ is the residual mean square after accounting for task and run effects. Unlike absolute agreement, consistency ICC allows for common shifts in performance between runs. We also report the mean pairwise Spearman correlation to assess the stability of task rankings.

We estimate uncertainty using 10,000 bootstrap resamples. In each resample, we draw 16 Design and 10 Reimplementation tasks with replacement, then resample the three runs within each selected task. This preserves the benchmark's task-family proportions.
Table~\ref{tab:variance-reliability} reports the resulting intervals.

\begin{table}[!htbp]
    \centering
    \caption{\textbf{Task-level repeatability.}
ICC(C,1) measures consistency across runs. The 95\% confidence intervals are estimated from 10,000 bootstrap resamples, drawing Design and Reimplementation tasks separately to preserve their proportions, then resampling runs within each task.}
    \label{tab:variance-reliability}
    
\begin{tabular}{lrrr}
\toprule
\rowcolor{TableHeader}
\textbf{Family} & \textbf{$\sigma_{\mathrm{task}}$} & \textbf{ICC(C,1)} & \textbf{Mean score (95\% CI)} \\
\midrule
Design & 0.255 & 0.93 & $[0.187,\,0.429]$ \\
Reimplementation & 0.180 & 0.96 & $[0.218,\,0.429]$ \\
\midrule
\textbf{Overall} & \textbf{0.225} & \textbf{0.93} & $\boldsymbol{[0.230,\,0.400]}$ \\
\bottomrule
\end{tabular}

\end{table}

With the task set held fixed, resampling runs within each task gives an overall 95\% interval of $[0.305,0.326]$. This interval is narrower than the hierarchical interval, which also reflects variation in the sampled tasks. Neither interval predicts the outcome of a new complete run, and the analysis is limited to three repetitions of one model--agent configuration.

\subsection{Task-level variation}
\label{app:variance-tasks}

Tables~\ref{tab:variance-design-tasks} and~\ref{tab:variance-reimplementation-tasks} provide the scores for every task and run. On Design, WDM-v2 (SD 0.173), WDM-v1 (0.139), and MZI-Ring-v1 (0.127) show the greatest variation. Other tasks fail consistently across all three runs, illustrating that low variance does not necessarily indicate reliable success. Reimplementation is generally more stable, with the largest variation observed on Trapezoidal SWG-v0 (0.083) and TFLN Crossing-v0 (0.051).

\begin{table}[!htbp]
    \centering
    \small
    \caption{\textbf{Design scores across repeated runs.}
    Final-submission scores for Sonnet 5.5 (high) with Claude Code. Mean and sample SD are calculated across v0, v1, and v2. Invalid or ineligible submissions receive zero.}
    \label{tab:variance-design-tasks}
    
% Add once to the preamble:
\definecolor{VarLight}{HTML}{F4F6F8}
\definecolor{VarMedium}{HTML}{E5EDF4}
\definecolor{VarStrong}{HTML}{CBDCEB}

\begin{tabular}{lrrrrr}
\toprule
\rowcolor{TableHeader}
\textbf{Task} & \textbf{1} & \textbf{2} & \textbf{3} & \textbf{Mean} & \textbf{SD} \\
\midrule
90SiN Bend-v0     & 0.609 & 0.564 & 0.573 & 0.582 & \cellcolor{VarLight}0.024 \\
MZI-Spec-8        & 0.465 & 0.484 & 0.467 & 0.472 & \cellcolor{VarLight}0.010 \\
Quantum-4         & 0.518 & 0.501 & 0.501 & 0.507 & \cellcolor{VarLight}0.010 \\
Quantum-8         & 0.000 & 0.000 & 0.000 & 0.000 & 0.000 \\
Ring-XC-v1        & 0.000 & 0.000 & 0.000 & 0.000 & 0.000 \\
Ring-XC-v2        & 0.000 & 0.000 & 0.000 & 0.000 & 0.000 \\
Microresonator-v1 & 0.650 & 0.663 & 0.733 & 0.682 & \cellcolor{VarLight}0.045 \\
Microresonator-v2 & 0.514 & 0.537 & 0.518 & 0.523 & \cellcolor{VarLight}0.013 \\
RAFT-Spec         & 0.412 & 0.399 & 0.405 & 0.405 & \cellcolor{VarLight}0.007 \\
MZI-Ring-v1       & 0.353 & 0.149 & 0.383 & 0.295 & \cellcolor{VarStrong}0.127 \\
MZI-Ring-v2       & 0.000 & 0.000 & 0.010 & 0.003 & \cellcolor{VarLight}0.006 \\
MZI-v1            & 0.681 & 0.559 & 0.699 & 0.646 & \cellcolor{VarMedium}0.077 \\
MZI-v2            & 0.388 & 0.503 & 0.531 & 0.474 & \cellcolor{VarMedium}0.076 \\
Serial MZI-v1     & 0.000 & 0.051 & 0.043 & 0.031 & \cellcolor{VarLight}0.027 \\
WDM-v1            & 0.133 & 0.080 & 0.343 & 0.185 & \cellcolor{VarStrong}0.139 \\
WDM-v2            & 0.340 & 0.117 & 0.000 & 0.152 & \cellcolor{VarStrong}0.173 \\
\bottomrule
\end{tabular}

\end{table}

\begin{table}[!htbp]
    \centering
    \small
    \caption{\textbf{Reimplementation scores across repeated runs.}
    Final-submission scores for Sonnet 5.5 (high) with Claude Code. Mean and sample SD are calculated across v0, v1, and v2.}
    \label{tab:variance-reimplementation-tasks}
    
\begin{tabular}{lrrrrr}
\toprule
\rowcolor{TableHeader}
\textbf{Task} & \textbf{1} & \textbf{2} & \textbf{3} & \textbf{Mean} & \textbf{SD} \\
\midrule
Anisotropic SWG-v0
    & 0.599 & 0.628 & 0.569 & 0.599 & \cellcolor{VarLight}0.030 \\
$\alpha$-Si/Si$_3$N$_4$ Grating-v0
    & 0.266 & 0.236 & 0.236 & 0.246 & \cellcolor{VarLight}0.017 \\
Dual-Trident SWG-v0
    & 0.345 & 0.349 & 0.389 & 0.361 & \cellcolor{VarLight}0.024 \\
HMDM polarization splitter--rotator
    & 0.104 & 0.103 & 0.092 & 0.100 & \cellcolor{VarLight}0.007 \\
HMDM six-channel WDM
    & 0.028 & 0.032 & 0.019 & 0.026 & \cellcolor{VarLight}0.007 \\
Sb$_2$Se$_3$/Si Rotator-v0
    & 0.466 & 0.507 & 0.484 & 0.486 & \cellcolor{VarLight}0.020 \\
SRSN Grating-v0
    & 0.525 & 0.531 & 0.521 & 0.526 & \cellcolor{VarLight}0.005 \\
TFLN Crossing-v0
    & 0.297 & 0.347 & 0.246 & 0.297 & \cellcolor{VarMedium}0.051 \\
TFLN PI-MMI-v0
    & 0.305 & 0.353 & 0.293 & 0.317 & \cellcolor{VarLight}0.032 \\
Trapezoidal SWG-v0
    & 0.349 & 0.194 & 0.325 & 0.289 & \cellcolor{VarMedium}0.083 \\
\bottomrule
\end{tabular}

\end{table}
% \FloatBarrier

% Evaluator and behavioural failure analyses

% \clearpage
\section{Evaluator diagnostics}
\label{app:evaluator-failures}

We analyse evaluator diagnostics from intermediate probes and independently graded final submissions. The same taxonomy is used for both, allowing us to compare the problems agents encounter during experimentation with those remaining in their final designs.

\subsection{Diagnostic taxonomy}
\label{app:diagnostic-taxonomy}

We classify diagnostics into nine categories covering engineering constraints, execution, and submission requirements. These categories identify failed checks.

\begin{itemize}[leftmargin=*,itemsep=0.4em,label=$\bullet$]
    \item \textbf{Geometry (Geo).} Fabrication and layout violations, including design-rule checks, minimum widths and spacings, bend constraints, required layers, optical connectivity, and topology.

    \item \textbf{Electrical (Elec).} Electrical connectivity and operation, including shorts, open circuits, contacts, heater connections, net mappings, and DC operating-point checks.

    \item \textbf{Simulation (Sim).} Simulation and numerical-model failures, including solver convergence, nonfinite results, field decay, passivity, reciprocity, and model validation.

    \item \textbf{Physical constraints (Phys).} Violations of physical performance or operating limits, including insertion loss, crosstalk, polarization balance, spectral requirements, actuator controllability, and thermal or electrical safety limits.

    \item \textbf{Submission requirements (Req).} Missing or invalid artifacts, schemas, manifests, provenance, integrity, or submission requirements.

    \item \textbf{Execution (Exec).} Runtime exceptions, syntax or import errors, missing dependencies, and failures to execute scripts or assessment checks.

    \item \textbf{Resource limits.} Timeout, memory, quota, or budget violations.

    \item \textbf{Service errors.} API, network, transport, and service-availability failures.

    \item \textbf{Other.} Diagnostics not covered by the preceding categories, including invalid probes without detailed error records.
\end{itemize}

\paragraph{Assignment and counting.}
We classify diagnostic fields, error codes, and failed checks using ordered matching rules. Each diagnostic is assigned its first matching category, while an assessment may receive multiple categories. Each category is counted at most once per assessment, and specific diagnostics take precedence over redundant aggregate flags. Categories without observed matches are omitted from the figures.

\subsection{Final-submission diagnostics}
\label{app:final-diagnostics}

\begin{figure}[!htbp]
    \centering
    \includegraphics[width=\linewidth]{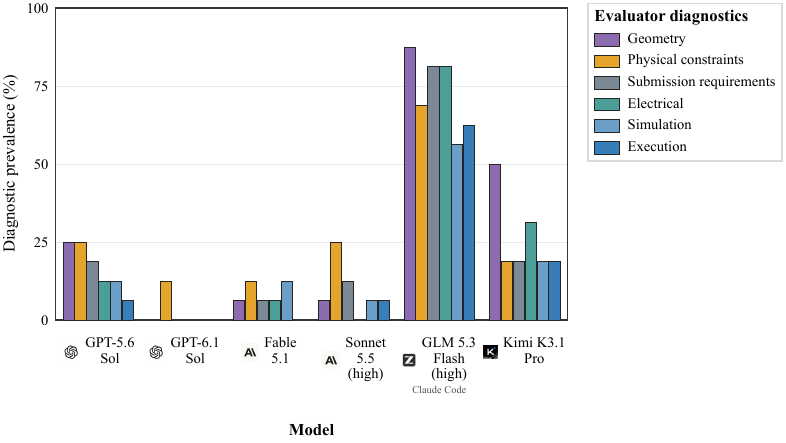}
    \caption{\textbf{Final-submission diagnostics for additional configurations.}
    The six configurations complement Figure~\ref{fig:final-error-categories}, using the same category colours, ordering, and 0--100\% scale. Each contributes 16 Design final submissions. Only invalid or ineligible submissions contribute to diagnostic counts; all 96 submissions remain in the denominator. Categories may overlap. GLM (high) uses Claude Code, while GLM (max) in the main figure uses mini-SWE-agent.}
    \label{fig:final-error-categories-additional}
\end{figure}

Figure~\ref{fig:final-error-categories-additional} extends the main-text analysis to the six remaining configurations: GPT-5.6 Sol, GPT-6.1 Sol, Fable 5.1, Sonnet 5.5 (high), GLM 5.3 Flash (high) with Claude Code, and Kimi K3.1 Pro. Of their 96 Design final submissions, 38 (39.6\%) are invalid or ineligible. Across all 11 configurations, 70 of 176 Design submissions (39.8\%) fail these requirements, most commonly involving geometry and physical constraints. All 110 Reimplementation final submissions are valid and eligible.

\subsection{Intermediate evaluator diagnostics}
\label{app:probe-diagnostics}

\begin{figure}[!htbp]
    \centering
    \includegraphics[width=\linewidth]{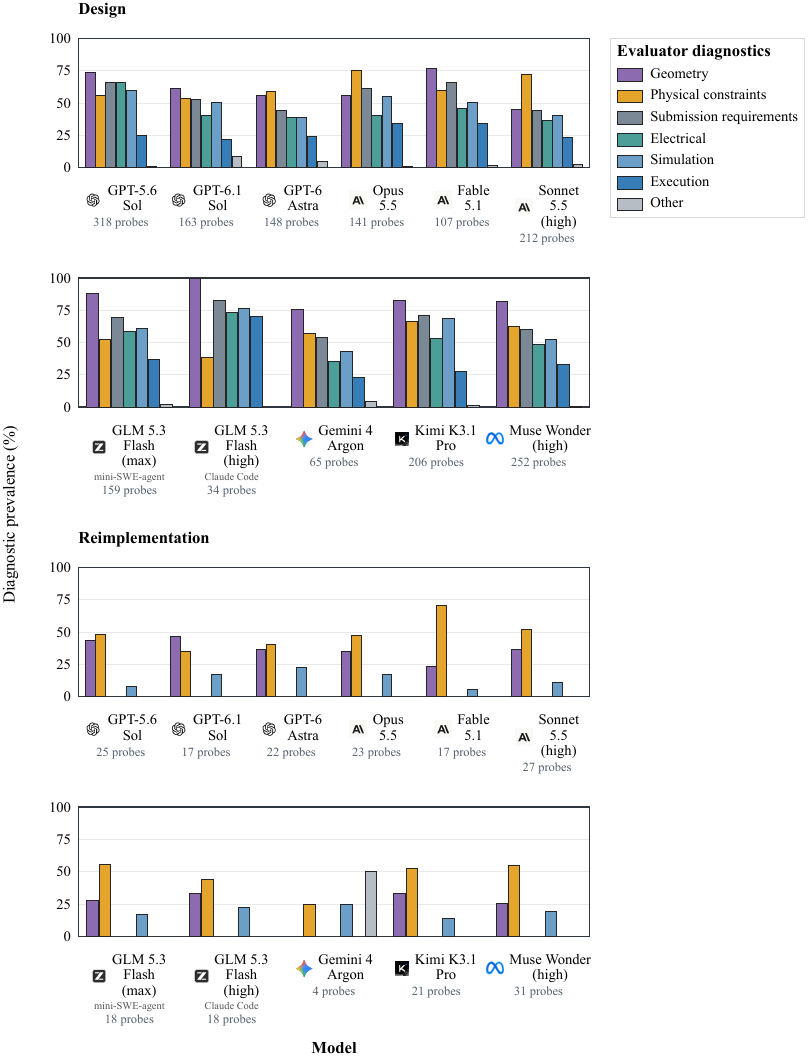}
    \caption{\textbf{Diagnostics in intermediate evaluator probes.}
    Diagnostic prevalence by model--agent configuration for Design (top two rows; 1,805 probes) and Reimplementation (bottom two rows; 223 probes). Counts beneath each configuration indicate probes with recorded diagnostics or invalid flags, including nonfatal diagnostics. Probes without such feedback are excluded. Categories may overlap. Colours follow Figure~\ref{fig:final-error-categories}.}
    \label{fig:probe-error-categories}
\end{figure}

Figure~\ref{fig:probe-error-categories} shows the diagnostic feedback recorded during evaluator probes. We include probes with diagnostic records or invalid flags, including valid probes with nonfatal diagnostics, and exclude probes without such feedback.

Across all configurations, 1,805 Design probes and 223 Reimplementation probes meet these criteria. Geometry is the most frequent Design diagnostic (70.6\%), followed by physical constraints (61.1\%), submission requirements (60.1\%), and simulation (53.9\%). Reimplementation feedback is concentrated in physical constraints (49.8\%) and geometry (33.6\%). These percentages describe prevalence among probes with recorded diagnostic feedback.

Intermediate diagnostics do not necessarily persist to final grading. GPT-6 Astra encounters diagnostic feedback on 148 Design probes but produces eligible final submissions on all 16 tasks. Similarly, agents encounter diagnostic feedback during Reimplementation experiments even though every final submission passes the validity checks.

\FloatBarrier

\clearpage

\clearpage
\section{Behavioural failure analysis}
\label{app:behavioural-analysis}

We examine agent trajectories using a taxonomy adapted from Terminal-Bench~\citep{merrill2026terminalbench} and the Multi-Agent System Failure Taxonomy (MAST)~\citep{cemri2026multi}. We extend it to capture errors in scientific judgement and the use of experimental feedback during engineering design.

\subsection{Trajectory review}
\label{app:behavioural-method}

We review 89 trajectories, comprising 54 Design and 35 Reimplementation runs, including both successful and unsuccessful outcomes. Table~\ref{tab:behavioural-review-coverage} summarizes the coverage across model--agent configurations.

GPT-5.6 Sol (high) serves as the judge. It examines task instructions, agent messages, tool inputs and outputs, intermediate evaluator feedback, and available images. The judge reads each trajectory chronologically and can revisit earlier evidence and subsequent actions that may correct an identified mistake. Reviews allow up to 180 interaction rounds, with a maximum of 16,384 output tokens per response.

The review environment is read-only. Model identities, agent scaffolds, and independent final grades are withheld, although tool interfaces may provide incidental clues. Intermediate evaluator feedback remains visible because it was available to the agent during the run.

\begin{table}[htbp]
    \centering
    \small
    \setlength{\tabcolsep}{6pt}
    \caption{\textbf{Coverage of behavioural trajectory reviews.}
    Completed Design and Reimplementation reviews by model--agent configuration. Each review assesses fourteen behavioural failure subtypes. Non-default reasoning effort appears in parentheses.}
    \label{tab:behavioural-review-coverage}
    
% Requires the previously defined TableHeader colour

\begin{tabular}{@{}llr@{}}
\toprule
\rowcolor{TableHeader}
\textbf{Model} & \textbf{Agent} & \textbf{Reviews used} \\
\midrule
\rowcolor{TableHeader}
\multicolumn{3}{@{}l}{\textbf{Design}} \\
GPT-6 Astra & Codex CLI & 11 \\
Opus 5.5 & Claude Code & 8 \\
GLM 5.3 Flash (max) & mini-SWE-agent & 13 \\
Gemini 4 Argon & mini-SWE-agent & 14 \\
Muse Wonder (high) & Muse Code & 8 \\
\multicolumn{2}{@{}l}{\textit{Subtotal}} & 54 \\
\midrule
\rowcolor{TableHeader}
\multicolumn{3}{@{}l}{\textbf{Reimplementation}} \\
GPT-6.1 Sol & Codex CLI & 8 \\
GPT-6 Astra & Codex CLI & 9 \\
Opus 5.5 & Claude Code & 6 \\
Fable 5.1 & Claude Code & 2 \\
Gemini 4 Argon & mini-SWE-agent & 10 \\
\multicolumn{2}{@{}l}{\textit{Subtotal}} & 35 \\
\midrule
\multicolumn{2}{@{}l}{\textbf{Total}} & \textbf{89} \\
\bottomrule
\end{tabular}

\end{table}

\subsection{Behavioural failure taxonomy}
\label{app:behavioural-taxonomy}

We retain Terminal-Bench's Execution, Coherence, and Verification families, adapting their nine subtypes to engineering design. We add four Scientific judgement subtypes and extend Coherence with failure to update from feedback. These additions address physical assumptions, scientific references, coupled constraints, experimental conclusions, and simulation feedback. They are also informed by failure-analysis themes in FrontierPhysics~\citep{you2026frontierphysics}.

The resulting taxonomy contains four families and fourteen subtypes. Each has explicit decision criteria and exclusions. A match requires an identifiable mistake supported by the trajectory; an invalid submission or poor score alone is insufficient.

\paragraph{Scientific judgement (S).}
\begin{description}[style=nextline,leftmargin=0pt]

    \item[S1: Misuse of reference information.]
    A reference fact or convention is materially misinterpreted or incorrectly translated into a design or experiment. Both the source and its observed use must be established. Deliberate departures from correctly understood references do not qualify.

    \item[S2: Incompatible physical assumptions.]
    A calculation or design decision relies on a physical relation, unit convention, or modelling regime incompatible with the task specification. Exploratory approximations and poor performance alone do not qualify.

    \item[S3: Failure to check joint feasibility.]
    The agent commits to a design whose accepted physical, fabrication, or control requirements are jointly incompatible. The conflict must be demonstrated; an invalid design or deliberate feasibility experiment is insufficient.

    \item[S4: Unsupported experimental conclusions.]
    The agent uses an unsupported causal or factor-effect conclusion to make a material engineering decision. Tentative hypotheses followed by relevant tests and informed optimization decisions are excluded.

\end{description}

\paragraph{Execution (E).}
\begin{description}[style=nextline,leftmargin=0pt]

    \item[E1: Instruction or constraint violation.]
    An observed action or omission materially contradicts an applicable instruction, required method, or submission contract. Technical errors and externally imposed termination do not qualify by themselves.

    \item[E2: Nonproductive repetition.]
    The agent repeats an operation without a meaningful change in information or strategy. Controlled parameter sweeps, informative replication, and monitoring active simulations are excluded.

    \item[E3: Failure to stop an exhausted or completed subgoal.]
    The agent continues a completed or demonstrably exhausted subgoal without a supported purpose. Further optimization of a valid candidate within the working budget remains legitimate.

\end{description}

\paragraph{Coherence (C).}
\begin{description}[style=nextline,leftmargin=0pt]

    \item[C1: Reasoning--action inconsistency.]
    An action or observed result contradicts a concrete plan or claim concerning the same design, property, candidate, and time. Failure of a compound check does not necessarily contradict every narrower claim.

    \item[C2: Loss of relevant state or context.]
    A later action conflicts with established information about files, candidates, results, or instructions. Losing a previously established best candidate may qualify; an unsuccessful final submission alone does not.

    \item[C3: Drift from the engineering objective.]
    The agent pursues an implemented diversion that displaces relevant work. Necessary research, inspection, diagnosis, and informative experiments are excluded.

    \item[C4: Failure to update from feedback.]
    The agent continues using an interpretation or strategy explicitly contradicted by earlier feedback. Inconclusive feedback and relevant but unsuccessful repairs do not qualify.

\end{description}

\paragraph{Verification (V).}
\begin{description}[style=nextline,leftmargin=0pt]

    \item[V1: Unjustified completion.]
    The agent voluntarily claims completion despite known unmet essential requirements without acknowledging them. Forced termination and an honest incomplete handoff are excluded.

    \item[V2: Missing or irrelevant verification.]
    The agent omits a relevant check, bypasses a required verifier, or substitutes an unrelated check. Running a verifier that subsequently reports failure does not establish this omission.

    \item[V3: Overstated or inadequate validation.]
    A validation claim is unsupported by the independence, coverage, provenance, or physical scope of the performed checks. Examples include circular validation and presenting a limited physical surrogate as broader verification.

\end{description}

\subsection{Judging procedure}
\label{app:behavioural-judge-procedure}

The judge assesses each subtype separately, recording the relevant decision, supporting evidence, its significance, and subsequent actions that may correct or contradict the observation. Scientific-judgement assessments additionally require an identifiable physical relation, reference, constraint, or experimental basis.

Each subtype receives one of four decisions:
\texttt{match}, \texttt{no\_match},
\texttt{insufficient\_evidence}, or
\texttt{not\_applicable}.
A match requires direct evidence of a material behavioural mistake.

The annotation schema requires decisions for all fourteen subtypes and checks evidence references and follow-up records. Each incident receives one primary subtype, although a trajectory may contain several distinct incidents. These annotations are exploratory LLM judgements and have not been independently validated against human assessments.

\subsection{Failure prevalence}
\label{app:behavioural-results}

For each subtype, prevalence is the fraction of assessable
trajectory reviews receiving a match. Insufficient-evidence
and not-applicable decisions are excluded from the denominator.
Multiple subtypes may occur in one trajectory, so percentages
need not sum to 100\%.

\begin{figure}[!htbp]
    \centering
    \includegraphics[width=\linewidth]
        {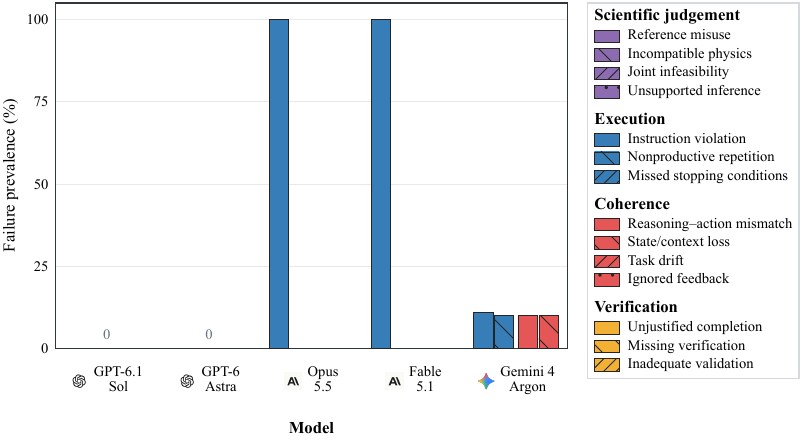}
    \caption{\textbf{Behavioural failures on Reimplementation tasks.}
    Prevalence of failure subtypes identified through LLM-based
    trajectory review across 35 assessable Reimplementation
    trajectories: 8 GPT-6.1 Sol, 9 GPT-6 Astra, 6 Opus,
    2 Fable, and 10 Gemini. Categories may overlap.
    Percentages should be interpreted cautiously for
    configurations with few reviewed trajectories.}
    \label{fig:behavioural-failures-reimplementation}
\end{figure}

The 89 reviews produce 1,246 subtype decisions: 73 matches,
1,154 no-matches, and 19 insufficient-evidence decisions.
The underlying records contain 97 failure episodes.
Figure~\ref{fig:behavioural-failures-design} presents the
Design results in the main paper, while
Figure~\ref{fig:behavioural-failures-reimplementation}
shows the Reimplementation results.

\subsection{Examples from agent trajectories}
\label{app:behavioural-examples}

The following examples illustrate observed decisions and their supporting evidence. They do not establish the causes of final performance outcomes.

\paragraph{Design.}

\begin{trajectoryexample}{Gemini --- Eight-mode projector: validation scope (V3)}
The agent describes an isolated thermal calibration as exact, although the evaluator models all thirty heat-source zones jointly, including crosstalk. Its official probes stop at earlier hierarchy checks, so the trajectory does not establish how this approximation affected final performance.
\end{trajectoryexample}

\begin{trajectoryexample}{GLM --- Four-mode projector: correction not implemented (C4)}
The agent identifies a phase-sign error and calculates a corrected vector, but subsequently continues using the original control-generation path. The correction is not carried through to the implementation, although its contribution to the final outcome cannot be determined from the trajectory alone.
\end{trajectoryexample}

\begin{trajectoryexample}{Muse --- Repeated probes after quota exhaustion (E2)}
The agent repeatedly relaunches evaluator probes after receiving explicit feedback that its probe quota is exhausted. This differs from legitimate monitoring of an active simulation or an informative sequence of experiments.
\end{trajectoryexample}

\begin{trajectoryexample}{Gemini --- Compiler task: ambiguous experimental inference (S4)}
The agent makes an unqualified numerical claim, immediately tests it, and then revises the candidate following feedback. This could reflect hypothesis-guided optimization expressed with excessive confidence.
\end{trajectoryexample}

\paragraph{Reimplementation.}

Of the 24 reviewed trajectories whose final scores do not exceed the published baseline ($S=0.5$), 21 have no behavioural match outside instruction violations. No Scientific judgement or Verification subtype matches were recorded in the reviewed Reimplementation set.

\begin{trajectoryexample}{GPT-6 Astra --- TFLN Crossing: valid but below published performance}
The agent uses thirty evaluator probes and produces a valid candidate with an objective of $J=0.1421$\,dB, yet achieves a normalized score of approximately 0.220. The recorded workflow does not establish a specific behavioural failure explaining why the published performance was not recovered.
\end{trajectoryexample}

\FloatBarrier

\clearpage

% Detailed agent trajectories
\section{Working Examples}
\label{app:working-examples}

% Generated case bodies come from PICBench/plotting/results/working_examples/.
% Rebuild with plotting.build_working_examples; no archive extraction is needed.
The following examples show excerpts from recorded agent interactions on WDM-v1 and the anisotropic SWG beamsplitter. For WDM-v1, we select GPT-6 Astra, while for the SWG beamsplitter, we select Gemini. Each example begins with the materials provided to the agent, followed by excerpts from its interactions. We use \texttt{[...]} to indicate omitted text or intervening turns. The final results and GDS layouts come from independent final evaluation, and all agent messages are quoted directly from the logs.

\begingroup
\definecolor{weink}{HTML}{303030}
\definecolor{weouterbackground}{HTML}{FAFAFA}
\definecolor{weouterheader}{HTML}{3F3F3F}
\definecolor{wecontextheader}{HTML}{E6E6E6}
\definecolor{wecontextbody}{HTML}{F7F7F7}
\definecolor{weagentheader}{HTML}{E0E0E0}
\definecolor{wetoolheader}{HTML}{E9E9E9}
\definecolor{wetoolbody}{HTML}{F5F5F5}
\definecolor{weresponseheader}{HTML}{EEEEEE}
\definecolor{wefinalheader}{HTML}{DDDDDD}
\tcbset{
 weouter/.style={enhanced,breakable,sharp corners,
  colback=weouterbackground,frame hidden,boxrule=0pt,
  colbacktitle=weouterheader,coltitle=white,
  fonttitle=\small\bfseries,fontupper=\small,
  left=6pt,right=6pt,top=5pt,bottom=5pt,
  before skip=7pt,after skip=7pt,pad at break*=4pt},
 webase/.style={enhanced,breakable=false,sharp corners,
  frame hidden,boxrule=0pt,titlerule=0pt,
  left=6pt,right=6pt,top=3pt,bottom=3pt,
  before skip=4pt,after skip=4pt,fontupper=\small,
  colback=white,coltitle=weink,fonttitle=\footnotesize\bfseries},
 wecontext/.style={webase,colback=wecontextbody,colbacktitle=wecontextheader},
 weagent/.style={webase,colbacktitle=weagentheader},
 wetool/.style={webase,colback=wetoolbody,colbacktitle=wetoolheader},
 weresponse/.style={webase,colbacktitle=weresponseheader},
 wefinal/.style={webase,colbacktitle=wefinalheader}
}
\newcommand{\weomitted}{\par\noindent\hfill{\footnotesize\color{black!50}\texttt{[...]}}\hfill\mbox{}\par}
\lstdefinestyle{wecode}{basicstyle=\ttfamily\scriptsize,
 columns=fullflexible,keepspaces=true,breaklines=true,
 showstringspaces=false,aboveskip=1pt,belowskip=1pt,frame=none,
 literate={µ}{{$\mu$}}1 {−}{{-}}1 {→}{{$\to$}}1
 {≥}{{$\geq$}}1 {≤}{{$\leq$}}1 {…}{{$\ldots$}}1
 {’}{{'}}1 {“}{{``}}1 {”}{{''}}1 {↔}{{$\leftrightarrow$}}1
 {–}{{-}}1 {—}{{-}}1}
\subsection{Design: WDM-v1 --- GPT-6 Astra / Codex}
\label{app:working-wdm_v1_astra}
\begin{tcolorbox}[weouter,title={Task materials and recorded trajectory},title after break={Design: WDM-v1 --- GPT-6 Astra / Codex (continued)}]
\noindent\textit{Budget: 120 minutes; 40 probes. Reasoning effort: medium.}\par\smallskip
\noindent\textbf{Materials available before the first agent turn.}\par
\begin{tcolorbox}[wecontext,title={instruction.md --- objective and constraints}]
\begin{lstlisting}[style=wecode]
[...]
Design one physical four-channel WDM channelizer near 1310 nm. A continuous
bus must feed four direct-coupled dual-ring add-drop filters. In each channel,
Ring A couples to the bus and Ring B; Ring B couples to Ring A and that
channel's drop waveguide. Each of the eight rings has a ring-following resistive
heater with an independent positive/negative terminal pair.
[...]
J =
    stabilized_channel_error
  + 0.10 * bandwidth_error
  + 0.25 * spurious_peak_penalty
  + 0.20 * channel_ripple
  + 0.20 * power_penalty
  + 0.15 * temperature_target_penalty
  + 0.15 * temperature_nonuniformity_penalty
  + 0.05 * current_density_penalty
  + 0.10 * compactness_ratio
  + 0.05 * route_metal_efficiency_penalty

official_score = J
[...]
\end{lstlisting}
\end{tcolorbox}
\begin{tcolorbox}[wecontext,title={harness\_contract.md --- submission and evaluation interface}]
\begin{lstlisting}[style=wecode]
[...]
Work in `/workspace` and submit only `design.py`. The required
entrypoint is `build_submission()`. Follow the exact return-value contract
in `task_profile.json`; generated local artifacts are diagnostic only.
[...]
Final collection transfers only the frozen source. A fresh private evaluator
container runs the same canonical scientific evaluation as the probe sidecar.
It calls the required functions in a restricted non-root builder, regenerates
canonical GDS, applies the task's layout and physics gates, and recomputes the
metric from evaluator-owned evidence. It never scores a cached probe result,
[...]
\end{lstlisting}
\end{tcolorbox}
\begin{tcolorbox}[wecontext,title={task\_profile.json --- editable submission and build contract}]
\begin{lstlisting}[style=wecode]
[...]
"submission": {
    "file": "design.py",
    "type": "component_hierarchy_layout_workflow",
    "entrypoint": "build_submission",
    "entrypoint_returns": "one_full_chip_photonforge_component",
    "script_runs_offline": true,
    "learner_models_trusted": false,
    "learner_metrics_trusted": false
  },
[...]
\end{lstlisting}
\end{tcolorbox}
\begin{tcolorbox}[weagent,title={Agent --- step 4}]
I'll read the contracts, inspect the starter, then probe improvements.
\end{tcolorbox}
\begin{tcolorbox}[wetool,title={Tool call: exec\_command --- step 5}]
\begin{lstlisting}[style=wecode]
ls -la /workspace && cat /workspace/harness_contract.md
\end{lstlisting}
\end{tcolorbox}
\begin{tcolorbox}[weresponse,title={Tool response --- step 5}]
\begin{lstlisting}[style=wecode]
[...]
Work in `/workspace` and submit only `design.py`. The required
entrypoint is `build_submission()`. Follow the exact return-value contract
in `task_profile.json`; generated local artifacts are diagnostic only.
[...]
\end{lstlisting}
\end{tcolorbox}
\weomitted
\begin{tcolorbox}[weagent,title={Agent --- step 59}]
Cloud results show overheating. I'm testing wider, inward-offset heaters.
\end{tcolorbox}
\begin{tcolorbox}[wetool,title={Tool call: exec\_command --- step 60}]
\begin{lstlisting}[style=wecode]
[...]
*** Update File: /workspace/design.py
[...]
-HEATER_WIDTH = 3.3
+HEATER_WIDTHS = (5.4, 5.8, 6.2, 6.6)
[...]
+    heater_radius = radius + 1.65 - heater_width / 2
[...]
\end{lstlisting}
\end{tcolorbox}
\weomitted
\begin{tcolorbox}[weagent,title={Agent --- step 82}]
I'm testing closed-loop heaters for more uniform heating, alongside coupler tuning.
\end{tcolorbox}
\weomitted
\begin{tcolorbox}[wetool,title={Tool call: exec\_command --- step 96}]
\begin{lstlisting}[style=wecode]
[...]
python design.py && run_local_drc design_outputs/final_pic.gds && cp design.py probe_6_source.py && run_probe > probe_6.log 2>&1
\end{lstlisting}
\end{tcolorbox}
\weomitted
\begin{tcolorbox}[wetool,title={Tool call: exec\_command --- step 106}]
\begin{lstlisting}[style=wecode]
jq '{score:.score, summary:.task_feedback.feedback.summary
[...]
probe_results_6.json
\end{lstlisting}
\end{tcolorbox}
\begin{tcolorbox}[weresponse,title={Tool response --- step 106}]
\begin{lstlisting}[style=wecode]
[...]
  "score": 2.583698093955099,
[...]
    "gate_pass": true,
    "hard_gate_failures": [],
[...]
\end{lstlisting}
\end{tcolorbox}
\begin{tcolorbox}[weagent,title={Agent --- step 107}]
All hard gates pass; the first valid score is 2.584.
\end{tcolorbox}
\weomitted
\begin{tcolorbox}[weagent,title={Agent --- step 122}]
Baseline saved. I'm testing unequal radii to suppress extra resonances.
\end{tcolorbox}
\weomitted
\begin{tcolorbox}[wetool,title={Tool call: exec\_command --- step 146}]
\begin{lstlisting}[style=wecode]
jq '{score:.score,sha:.submission_sha256
[...]
probe_results_8.json
\end{lstlisting}
\end{tcolorbox}
\begin{tcolorbox}[weresponse,title={Tool response --- step 146}]
\begin{lstlisting}[style=wecode]
[...]
  "score": 0.9435412285574741,
  "sha": "a2578d64a0cb78a71c4492e2d3ecca5888b526759b365c4f2c2283e8762c889b",
[...]
\end{lstlisting}
\end{tcolorbox}
\begin{tcolorbox}[weagent,title={Agent --- step 147}]
The Vernier design passes all gates and lowers J to 0.944.
\end{tcolorbox}
\begin{center}\includegraphics[width=.90\linewidth]{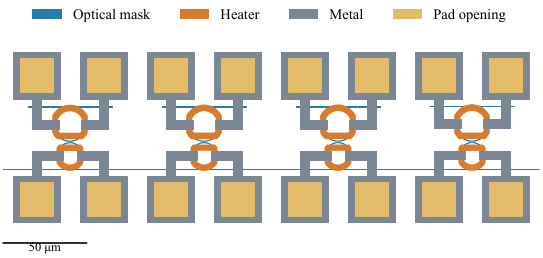}\par
{\footnotesize Retained GDS for the verified probe-8 checkpoint; source and geometry hashes are bound to that probe.}\end{center}
\weomitted
\begin{tcolorbox}[weagent,title={Agent --- step 266}]
The balanced candidate passes local checks; I'll inspect its layout.
\end{tcolorbox}
\weomitted
\begin{tcolorbox}[wetool,title={Tool call: view\_image --- step 268}]
\begin{lstlisting}[style=wecode]
{"path": "/workspace/balanced_layout.png"}
\end{lstlisting}
\end{tcolorbox}
\begin{center}\includegraphics[width=.70\linewidth]{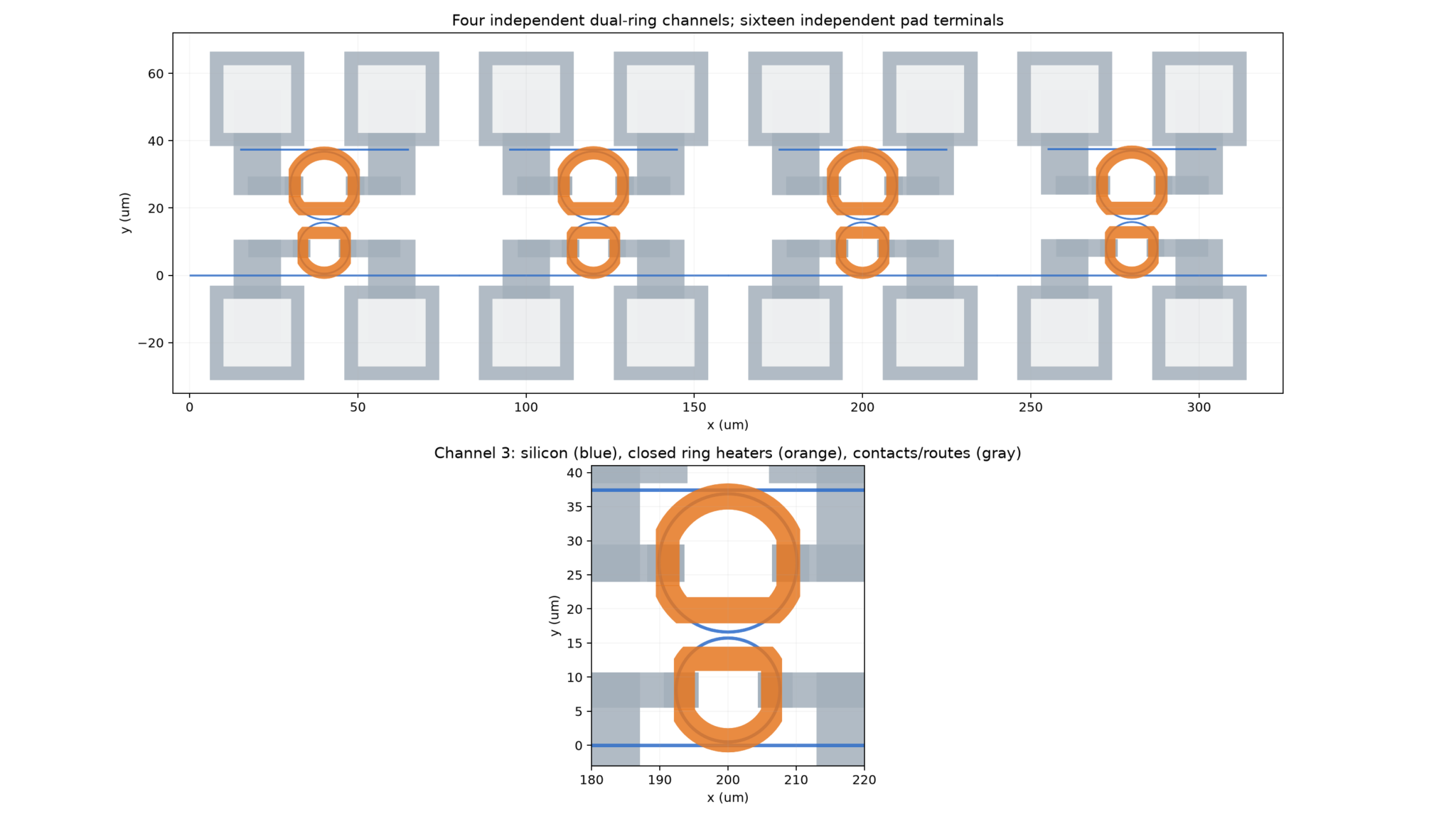}\par
{\footnotesize Original image returned by view\_image at step 268. The locally built balanced candidate is an intermediate design, not the final submission.}\end{center}
\weomitted
\begin{tcolorbox}[weagent,title={Agent --- step 332}]
The final trial regressed; I'm retaining the verified J = 0.31484 design.
\end{tcolorbox}
\weomitted
\begin{tcolorbox}[wetool,title={Tool call: exec\_command --- step 334}]
\begin{lstlisting}[style=wecode]
[...]
print('verified_source_digest_match',hashlib.sha256(source).hexdigest()==result['submission_sha256'])
print('best_probe_valid',result['valid'],'gate_pass',result['gate_pass'],'raw_J',result['score'])
[...]
benchmark_status
\end{lstlisting}
\end{tcolorbox}
\par\noindent\begin{minipage}{\linewidth}
\weomitted
\begin{tcolorbox}[weagent,title={Last logged agent message --- step 336}]
Final local checks pass, and the source matches probe 16 exactly.
\end{tcolorbox}
\begin{tcolorbox}[wetool,title={Tool call: exec\_command --- step 337}]
\begin{lstlisting}[style=wecode]
[...]
result=json.load(open('probe_results_16.json'));feedback=result['task_feedback']['feedback'];electrical=feedback['electrical'];operating=electrical['operating_point'];heaters=electrical['per_heater'];rings=feedback['thermal']['per_ring']
assert len(heaters)==len(rings)==8
[...]
assert hashlib.sha256(Path('design.py').read_bytes()).hexdigest()==result['submission_sha256']
[...]
\end{lstlisting}
\end{tcolorbox}
\begin{tcolorbox}[weresponse,title={Tool response --- step 337}]
\begin{lstlisting}[style=wecode]
[...]
Eight independent heater operating points and eight 96-sample ring monitors verified.
Electrical power conservation verified: 0.03188572930393017 W
Worst ring temperature nonuniformity: 2.9185791015625 K
Preserved verified raw J: 0.31483570369233954
\end{lstlisting}
\end{tcolorbox}
\begin{tcolorbox}[wefinal,title={Independent final evaluation}]
The frozen submission passes the native final gates. The independently rebuilt final gives $J=0.314836$ and $S=0.846034$ on the paper's 0--1 scale. Its source hash matches probe 16.
\end{tcolorbox}
\begin{center}\includegraphics[width=.90\linewidth]{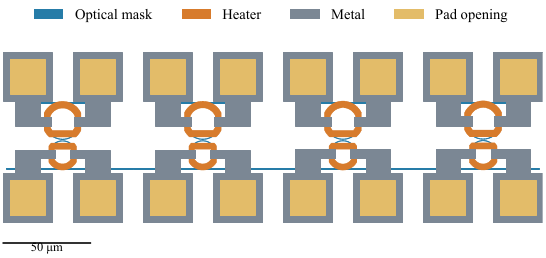}\par
{\footnotesize Original GDS from the independent final evaluation.}\end{center}
\noindent\footnotesize The retained trace contains no terminal final-response message; step 336 is its last logged agent message.\normalsize
\end{minipage}
\end{tcolorbox}

\clearpage
\subsection{Reimplementation: SWG beamsplitter --- Gemini 4 Argon / mini-SWE-agent}
\label{app:working-swg_gemini}
\begin{tcolorbox}[weouter,title={Task materials and recorded trajectory},title after break={Reimplementation: SWG beamsplitter --- Gemini 4 Argon / mini-SWE-agent (continued)}]
\noindent\textit{Budget: 120 minutes; 30 probes. Reasoning effort: medium.}\par\smallskip
\noindent\textbf{Materials available before the first agent turn.}\par
\begin{tcolorbox}[wecontext,title={instruction.md --- objective and symmetry contract}]
\begin{lstlisting}[style=wecode]
[...]
Implement the Halir-style anisotropic sub-wavelength-grating 2 x 2 MMI
beamsplitter from the provided paper as one final layout component. First
reproduce the paper geometry and use trusted probes to debug it; then optimize
the permitted geometry over `1250-1750 nm`.

## Objective

At each sampled wavelength, compute the normalized envelope penalty

```text
max(max(excess_loss_dB, 0) / 1 dB,
    imbalance_dB / 1 dB,
    quadrature_phase_error_deg / 5 deg)
```

The official score `J` is `max_wavelength` of this penalty over the 10-nm
wavelength grid. Lower is better. The evaluator simulates the `y=+1 um` input.
Eligibility requires exact `y=0` mirror symmetry of both submitted mask unions;
therefore the `y=-1 um` input has the same `J` with the two output labels
exchanged. A geometrically asymmetric candidate is invalid rather than being
scored from only one input.
[...]
\end{lstlisting}
\end{tcolorbox}
\begin{tcolorbox}[wecontext,title={harness\_contract.md --- submission interface}]
\begin{lstlisting}[style=wecode]
[...]
Work in `/workspace` and edit only `design.py`. It must define
`build_submission()` and return one PhotonForge component matching the exact
interface and geometry contract in `task_profile.json`.

You may run the source locally for layout debugging. Learner-created GDS,
metrics, JSON, cloud task IDs, downloaded results, HDF5, S-parameters, and
plots are diagnostic only and never final-score authority.
[...]
\end{lstlisting}
\end{tcolorbox}
\begin{tcolorbox}[wecontext,title={task\_profile.json --- returned component contract}]
\begin{lstlisting}[style=wecode]
[...]
  "submission": {
    "file": "design.py",
    "function": "build_submission",
    "return_type": "photonforge_component",
    "final_artifact": "final_pic.gds",
    "one_final_component": true,
    "learner_metrics_accepted": false,
    "learner_cloud_evidence_accepted": false
  }
[...]
    "mask_semantics": {
      "1/0": "silicon_device_layer_core",
      "2/0": "residual_bottom_air_void_footprint"
    },
[...]
\end{lstlisting}
\end{tcolorbox}
\begin{tcolorbox}[weagent,title={Agent --- step 3}]
I will start by inspecting the workspace directory, checking \texttt{benchmark\_status}, and reading \texttt{task\_profile.json}, \texttt{harness\_contract.md}, and \texttt{references/paper\_parameters.json}.
\end{tcolorbox}
\begin{tcolorbox}[wetool,title={Tool call: bash --- step 3}]
\begin{lstlisting}[style=wecode]
ls -la && benchmark_status && cat harness_contract.md && cat task_profile.json && cat references/paper_parameters.json
\end{lstlisting}
\end{tcolorbox}
\weomitted
\begin{tcolorbox}[weagent,title={Agent --- step 24}]
I will run a check on the exact longitudinal coordinates and verify whether layers \texttt{(1, 0)} and \texttt{(2, 0)} have any overlap.
\end{tcolorbox}
\begin{tcolorbox}[wetool,title={Tool call: bash --- step 24}]
\begin{lstlisting}[style=wecode]
python3 -c "
import gdstk
import design

comp = design.build_submission()
comp.write_gds('/tmp/test_design.gds')
lib = gdstk.read_gds('/tmp/test_design.gds')
cell = lib.top_level()[0]
p1 = [p for p in cell.polygons if (p.layer, p.datatype) == (1, 0)]
p2 = [p for p in cell.polygons if (p.layer, p.datatype) == (2, 0)]
overlap = gdstk.boolean(p1, p2, 'and', precision=1e-6)
print('Overlap between (1,0) and (2,0) area:', sum(p.area() for p in overlap))
"
\end{lstlisting}
\end{tcolorbox}
\begin{tcolorbox}[weresponse,title={Tool response --- step 24}]
\begin{lstlisting}[style=wecode]
[...]
Overlap between (1,0) and (2,0) area: 0.005744269999999927
[...]
\end{lstlisting}
\end{tcolorbox}
\begin{tcolorbox}[weagent,title={Agent --- step 25}]
I will now update \texttt{/workspace/design.py} so that the layer \texttt{(1, 0)} and \texttt{(2, 0)} masks have zero overlap, verify all local geometry checks, and run our first probe (\texttt{run\_probe}) on the paper baseline geometry.
\end{tcolorbox}
\begin{tcolorbox}[wetool,title={Tool call: bash --- step 25}]
\begin{lstlisting}[style=wecode]
cat <<'EOF' > /workspace/design.py
[...]
    pitch = 0.190
    duty_cycle = 0.50
    bar_len = _r(pitch * duty_cycle)       # 0.095 um
    gap_len = _r(pitch - bar_len)          # 0.095 um

    w_in = 0.500
    w_access = 1.700
    w_mmi = 3.250
    y_port = 1.000

    n_taper = 30
    n_mmi = 74

    w_bridge_start = 0.500
    w_bridge_end = 0.100
[...]
run_probe
\end{lstlisting}
\end{tcolorbox}
\begin{tcolorbox}[weresponse,title={Tool response --- step 25}]
\begin{lstlisting}[style=wecode]
[...]
    "gate_pass": true,
    "metric": 0.8193063830022828,
[...]
            "worst_excess_loss_db": 0.8193063830022828,
            "worst_imbalance_db": 0.610762846542852,
            "worst_phase_error_deg": 2.773571794270481,
            "worst_wavelength_nm": 1280.0
[...]
\end{lstlisting}
\end{tcolorbox}
\weomitted
\begin{tcolorbox}[weagent,title={Agent --- step 30}]
I will now test making the outer edge of the access taper flush with the MMI outer wall (\texttt{y\_outer = W\_MMI / 2 = 1.625 um}, \texttt{y\_inner = s / 2 = 0.150 um}) at the MMI interface while keeping all other parameters at the paper baseline, and run Probe 2.
\end{tcolorbox}
\begin{tcolorbox}[wetool,title={Tool call: bash --- step 30}]
\begin{lstlisting}[style=wecode]
cat <<'EOF' > /workspace/design.py
[...]
    w_in = 0.500
    w_mmi = 3.250
    sep = 0.300
    y_port = 1.000

    # At start of taper: [0.750, 1.250]
    # At end of taper (MMI interface): [sep/2, w_mmi/2] = [0.150, 1.625], flush with MMI outer wall
    y_in_bot = _r(y_port - w_in / 2.0)     # 0.750
    y_in_top = _r(y_port + w_in / 2.0)     # 1.250
    y_acc_bot = _r(sep / 2.0)              # 0.150
    y_acc_top = _r(w_mmi / 2.0)            # 1.625
[...]
run_probe
\end{lstlisting}
\end{tcolorbox}
\begin{tcolorbox}[weresponse,title={Tool response --- step 30}]
\begin{lstlisting}[style=wecode]
[...]
    "gate_pass": true,
    "metric": 1.032575465890094,
[...]
            "worst_excess_loss_db": 0.2974236591224704,
            "worst_imbalance_db": 1.032575465890094,
            "worst_phase_error_deg": 3.1754846629998497,
            "worst_wavelength_nm": 1750.0
[...]
\end{lstlisting}
\end{tcolorbox}
\weomitted
\begin{tcolorbox}[weagent,title={Agent --- step 36}]
I will save a backup of Probe 6 (\texttt{J = 0.785838}) and run Probe 7 with \texttt{y\_acc\_top = 1.795 um} to measure the exact balance between \texttt{worst\_excess\_loss\_db} and \texttt{worst\_imbalance\_db} between \texttt{1.740 um} and \texttt{1.850 um}.
\end{tcolorbox}
\begin{tcolorbox}[wetool,title={Tool call: bash --- step 36}]
\begin{lstlisting}[style=wecode]
cp /workspace/design.py /tmp/design_probe6_0.7858.py && sed -i 's/y_acc_top = 1.740/y_acc_top = 1.795/' /workspace/design.py && run_probe
\end{lstlisting}
\end{tcolorbox}
\begin{tcolorbox}[weresponse,title={Tool response --- step 36}]
\begin{lstlisting}[style=wecode]
[...]
    "gate_pass": true,
    "metric": 0.6220642575062809,
[...]
            "worst_excess_loss_db": 0.43494820100200826,
            "worst_imbalance_db": 0.6220642575062809,
            "worst_phase_error_deg": 1.8243622918471374,
            "worst_wavelength_nm": 1750.0
[...]
\end{lstlisting}
\end{tcolorbox}
\weomitted
\begin{tcolorbox}[weagent,title={Agent --- step 48}]
I will now run Probe 17 at the $\lambda = 0.48$ interpolation point between Probe 13 (\texttt{wEL = 0.4340, wIB = 0.4339, wPE/5 = 0.3766}) and Probe 14 (\texttt{wEL = 0.3640, wIB = 0.3924, wPE/5 = 0.4739}) to minimize the maximum of the three normalized penalties.
\end{tcolorbox}
\begin{tcolorbox}[wetool,title={Tool call: bash --- step 48}]
\begin{lstlisting}[style=wecode]
[...]
    # 3-stage convex bridge profile (identical to Probes 13 and 14)
    bridge_knots = [
        (0.00, 0.500),
        (0.50, 0.336),
        (0.82, 0.228),
        (1.00, 0.102),
    ]

    # Exact lambda=0.48 interpolation between Probe 13 and Probe 14 bar_knots
    bar_knots = [
        (0.00, 0.750, 1.250),
        (0.50, 0.445, 1.540),
        (0.82, 0.261, 1.706),
        (1.00, 0.150, 1.771),
    ]
[...]
run_probe
\end{lstlisting}
\end{tcolorbox}
\begin{tcolorbox}[weresponse,title={Tool response --- step 48}]
\begin{lstlisting}[style=wecode]
[...]
    "gate_pass": true,
    "metric": 0.419231414123567,
[...]
            "worst_excess_loss_db": 0.4073529502564016,
            "worst_imbalance_db": 0.4121118678149977,
            "worst_phase_error_deg": 2.096157070617835,
            "worst_wavelength_nm": 1750.0
[...]
\end{lstlisting}
\end{tcolorbox}
\weomitted
\begin{tcolorbox}[weagent,title={Agent --- step 61}]
I will run a Python script to analyze the affine response surface across the three fixed-mesh probes (\texttt{P17}, \texttt{P23}, \texttt{P24}, all with \texttt{FC = 0.937877850569188}) and locate the minimax point where \texttt{max(wEL, wIB, wPE/5)} is minimized.
\end{tcolorbox}
\begin{tcolorbox}[wetool,title={Tool call: bash --- step 61}]
\begin{lstlisting}[style=wecode]
python3 -c "
import numpy as np
[...]
for u in np.linspace(0, 0.5, 11):
    for v in np.linspace(0, 0.5, 6):
        # 1750nm branch:
        el_1750 = 0.4074 + v*(0.4008 - 0.4074) - u*0.005
        ib_1750 = 0.4121 + v*(0.4312 - 0.4121) - u*0.028
        pe_1750 = 0.4192 + v*(0.4165 - 0.4192) + u*(0.4050 - 0.4192)
        # 1320nm branch (in P17 u=0, v=0, el_1320 ~ 0.395, ib_1320 ~ 0.395):
        el_1320 = 0.395 - v*0.010 + u*(0.4471 - 0.395)
        ib_1320 = 0.395 - v*0.010 + u*(0.4177 - 0.395)
        j = max(el_1750, ib_1750, pe_1750, el_1320, ib_1320)
        if j < 0.414:
            print(f'u={u:.2f}, v={v:.2f}: el1750={el_1750:.4f}, ib1750={ib_1750:.4f}, pe={pe_1750:.4f}, el1320={el_1320:.4f}, ib1320={ib_1320:.4f} -> J={j:.4f}')
[...]
\end{lstlisting}
\end{tcolorbox}
\begin{tcolorbox}[weresponse,title={Tool response: local response-surface prediction --- step 61}]
\begin{lstlisting}[style=wecode]
[...]
u=0.40, v=0.40: el1750=0.4028, ib1750=0.4085, pe=0.4124, el1320=0.4118, ib1320=0.4001 -> J=0.4124
[...]
\end{lstlisting}
\end{tcolorbox}
\par\noindent\begin{minipage}{\linewidth}
\begin{tcolorbox}[weagent,title={Last recorded agent message --- native console}]
\raggedright 
[...]
I will now run Probe 27 on the locked \texttt{FC = 0.937878} mesh using Probe 17's 
exact \texttt{bridge\_knots = [(0.00, 0.500), (0.50, 0.336), (0.82, 0.228), (1.00, 
0.102)]} and \texttt{yt = (1.250, 1.540, 1.706, 1.771)} with the 25\% pure \texttt{yb} step \texttt{yb
= (0.750, 0.447, 0.265, 0.156)} derived from Probes 17, 25, and 26.
[...]
\end{tcolorbox}
\begin{tcolorbox}[wetool,title={Tool call: bash --- final recorded action}]
\begin{lstlisting}[style=wecode]
[...]
    # 3-stage convex bridge profile (identical to P17)
    bridge_knots = [
        (0.00, 0.500),
        (0.50, 0.336),
        (0.82, 0.228),
        (1.00, 0.102),
    ]

    # 3-stage SWG taper bar profile: P17 yt with 25% pure yb step toward P26
    bar_knots = [
        (0.00, 0.750, 1.250),
        (0.50, 0.447, 1.540),
        (0.82, 0.265, 1.706),
        (1.00, 0.156, 1.771),
    ]
[...]
run_probe
[...]
\end{lstlisting}
\end{tcolorbox}
\begin{tcolorbox}[wefinal,title={Independent final evaluation}]
The frozen submission passes the native final gates. The independently rebuilt final gives $J=0.418203$ and $S=0.705118$ on the paper's 0--1 scale. Its source hash matches accepted probe 27. The response to that probe was not retained; the objective reported here comes from the independent final evaluation.
\end{tcolorbox}
\begin{center}\includegraphics[width=.90\linewidth]{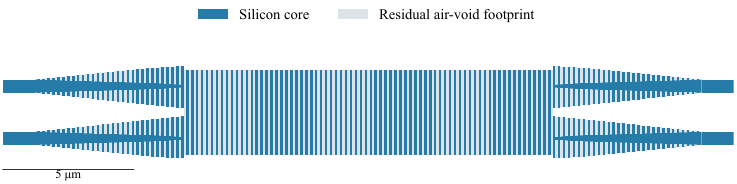}\par
{\footnotesize Original GDS from the independent final evaluation.}\end{center}
\noindent\footnotesize No terminal completion message is retained. The last agent message and call shown above come from the native console log; the structured trajectory ends after the response to probe 26.\normalsize
\end{minipage}
\end{tcolorbox}

% Third example retained for later inclusion:
% \clearpage
% \input{working_examples/srsn_fable}
\endgroup

\FloatBarrier

\clearpage

% Complete gallery of final engineering designs

\section{Final design layouts}
\label{app:final-layouts}

Figures in this section compare the final submissions of all 11 model--agent configurations across the 26 tasks, with Design presented first, followed by Reimplementation. Layouts within each task are shown at a common physical scale and orientation, except where noted.

Design figures show the optical and electrical structures, with annotation layers omitted for clarity. Reimplementation figures also indicate input and output ports, device dimensions, and components supplied by the evaluator. Differences in rendering conventions and any changes to visual scale are noted in the corresponding captions.

Each panel reports the normalized score from independent final grading. Invalid or ineligible submissions receive zero and are marked with an asterisk; missing geometry is indicated explicitly.

\newpage
\subsection*{Design tasks (16)}

\subsection{Eight-channel electrothermal WDM channelizer}
\label{app:layouts-wdm-channelizer-electrothermal-design-v2}

\begin{figure}[htbp]
    \centering
    \includegraphics[width=\linewidth]{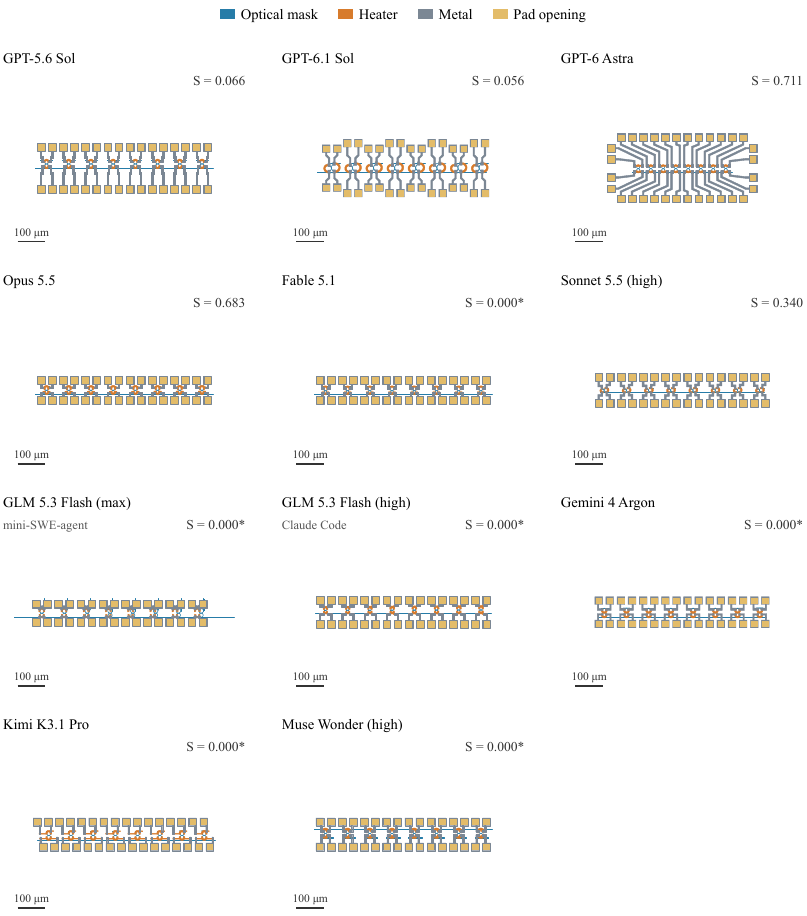}
    \caption{\textbf{WDM-v2 final layouts.} Eight-channel electrothermal WDM circuits with optical, heater, metal-routing, and contact layers.}
    \label{fig:final-layouts-wdm-channelizer-electrothermal-design-v2}
\end{figure}
\clearpage

\subsection{Four-channel electrothermal WDM channelizer}
\label{app:layouts-wdm-channelizer-electrothermal-design-v1}

\begin{figure}[htbp]
    \centering
    \includegraphics[width=\linewidth]{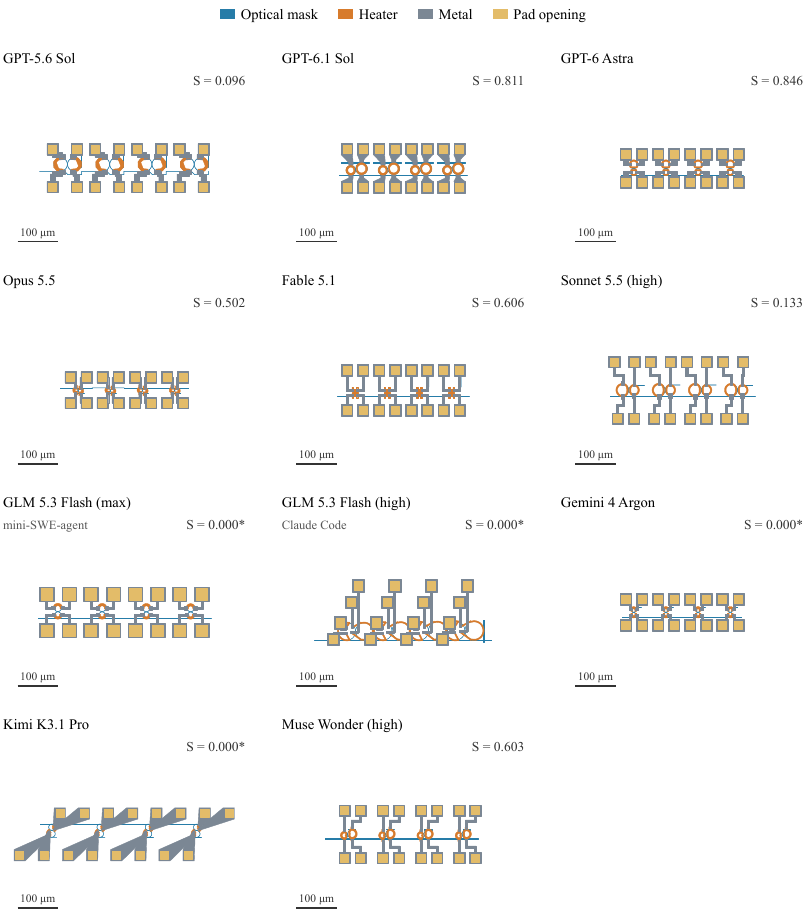}
    \caption{\textbf{WDM-v1 final layouts.} Four dual-ring filters sharing an optical bus, with eight contacted heaters. The layouts reveal differences in routing footprint and contact placement.}
    \label{fig:final-layouts-wdm-channelizer-electrothermal-design-v1}
    \label{fig:final-layouts-wdm}
\end{figure}
\clearpage

\subsection{Serial dual-ring-assisted MZI filter}
\label{app:layouts-serial-dual-ring-assisted-mzi-chebyshev-elliptic-electrothermal-design-v1}

\begin{figure}[htbp]
    \centering
    \includegraphics[width=\linewidth]{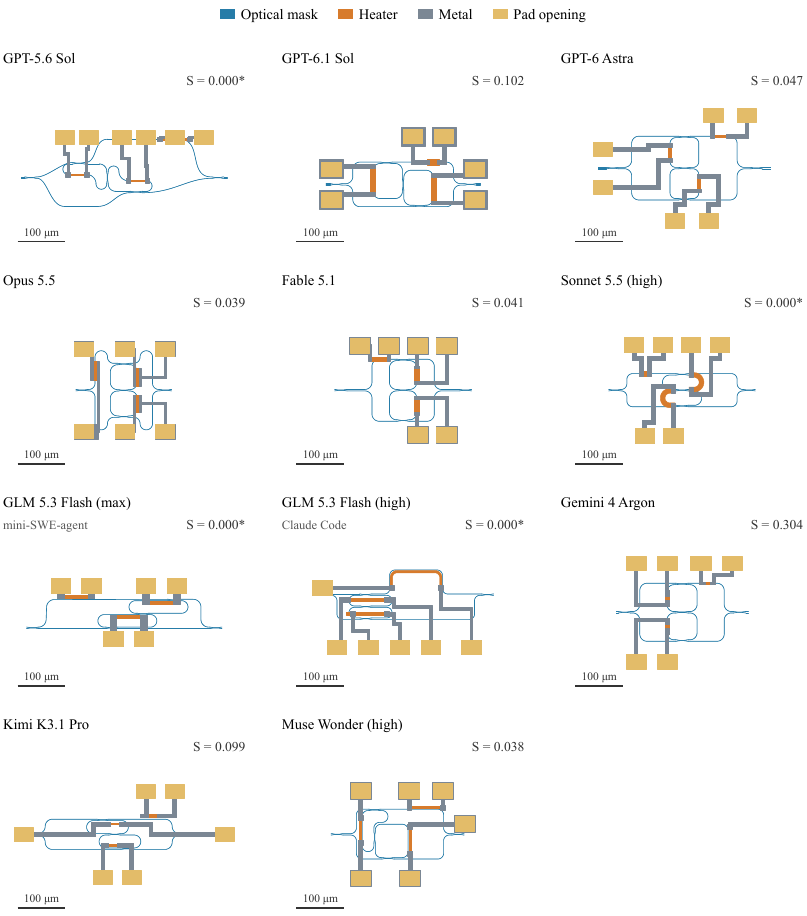}
    \caption{\textbf{Serial MZI-v1 final layouts.} Serial dual-ring-assisted Mach--Zehnder filters with optical, thermal, and electrical structures.}
    \label{fig:final-layouts-serial-dual-ring-assisted-mzi-chebyshev-elliptic-electrothermal-design-v1}
\end{figure}
\clearpage

\subsection{Six-by-six rectangular MZI mesh}
\label{app:layouts-rectangular-mzi-mesh-design-v2}

\begin{figure}[htbp]
    \centering
    \includegraphics[width=\linewidth]{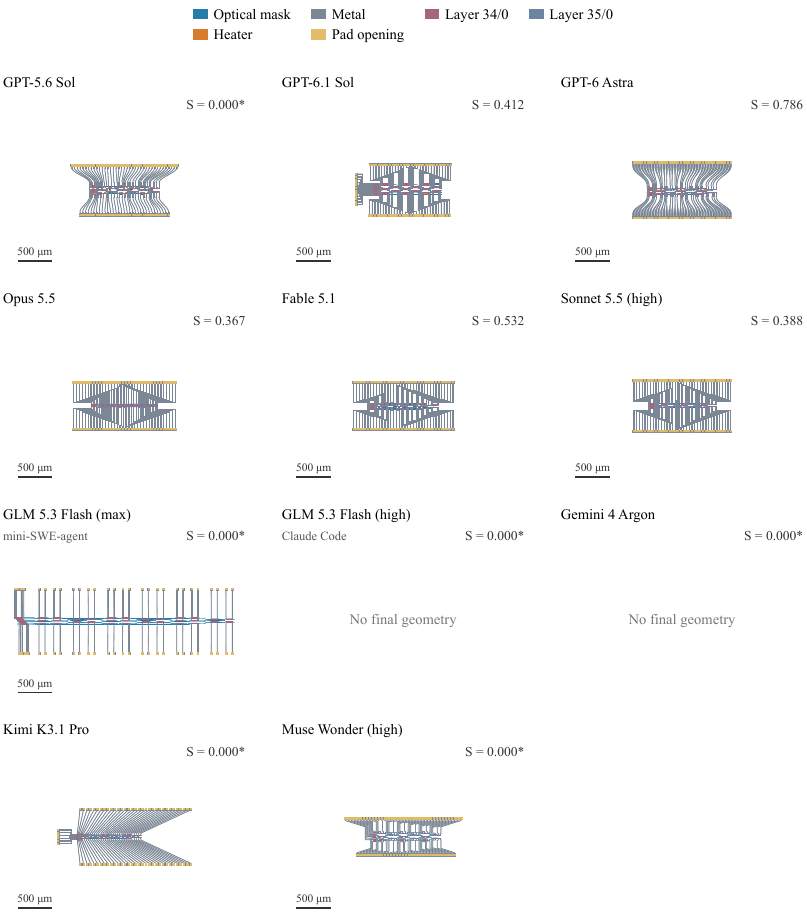}
    \caption{\textbf{MZI-v2 final layouts.} Broadband $6\times6$ Mach--Zehnder meshes with optical routing, electrical fanout, and contacts.}
    \label{fig:final-layouts-rectangular-mzi-mesh-design-v2}
\end{figure}
\clearpage

\subsection{Four-by-four rectangular MZI mesh}
\label{app:layouts-rectangular-mzi-mesh-design-v1}

\begin{figure}[htbp]
    \centering
    \includegraphics[width=\linewidth]{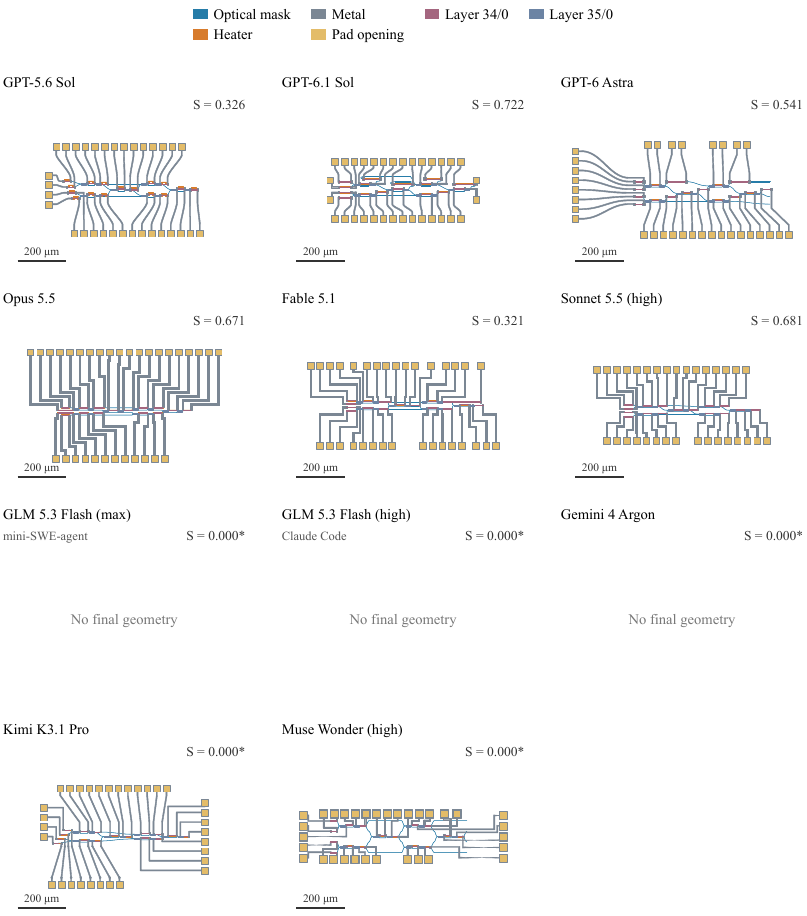}
    \caption{\textbf{MZI-v1 final layouts.} Broadband $4\times4$ Mach--Zehnder meshes with optical routing, electrical fanout, and contacts.}
    \label{fig:final-layouts-rectangular-mzi-mesh-design-v1}
\end{figure}
\clearpage

\subsection{MZI-coupled microring filter v2}
\label{app:layouts-mzi-coupled-microring-add-drop-filter-electrothermal-design-v2}

\begin{figure}[htbp]
    \centering
    \includegraphics[width=\linewidth]{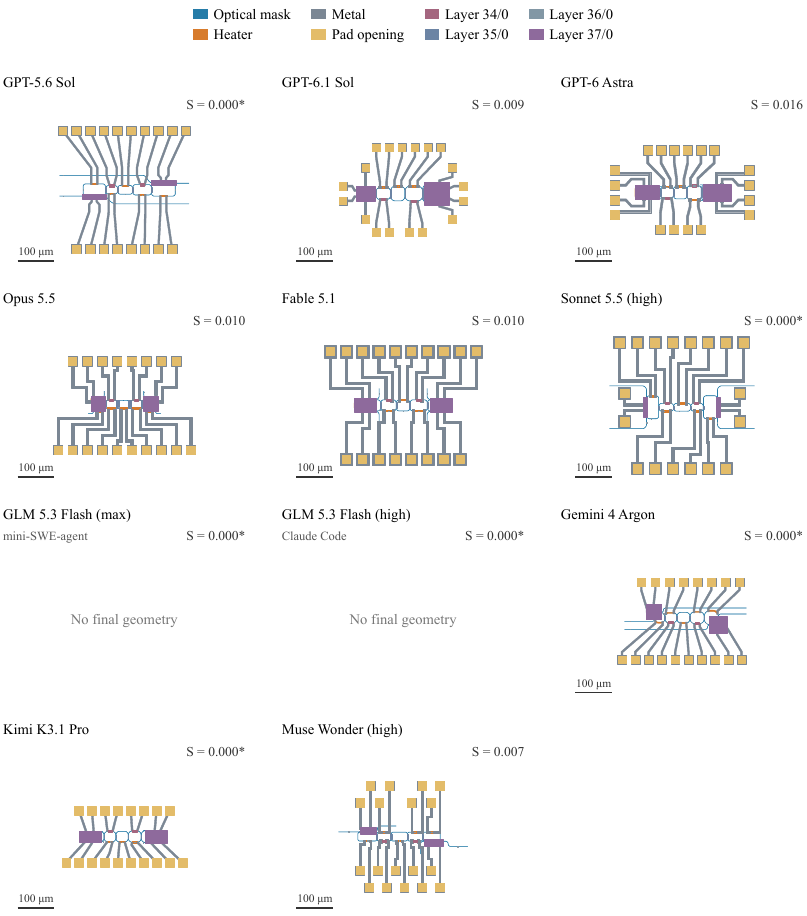}
    \caption{\textbf{MZI-Ring-v2 final layouts.} Second version of the electrothermal MZI-coupled microring add-drop filter, with optical and electrical geometry.}
    \label{fig:final-layouts-mzi-coupled-microring-add-drop-filter-electrothermal-design-v2}
\end{figure}
\clearpage

\subsection{MZI-coupled microring filter v1}
\label{app:layouts-mzi-coupled-microring-add-drop-filter-electrothermal-design-v1}

\begin{figure}[htbp]
    \centering
    \includegraphics[width=\linewidth]{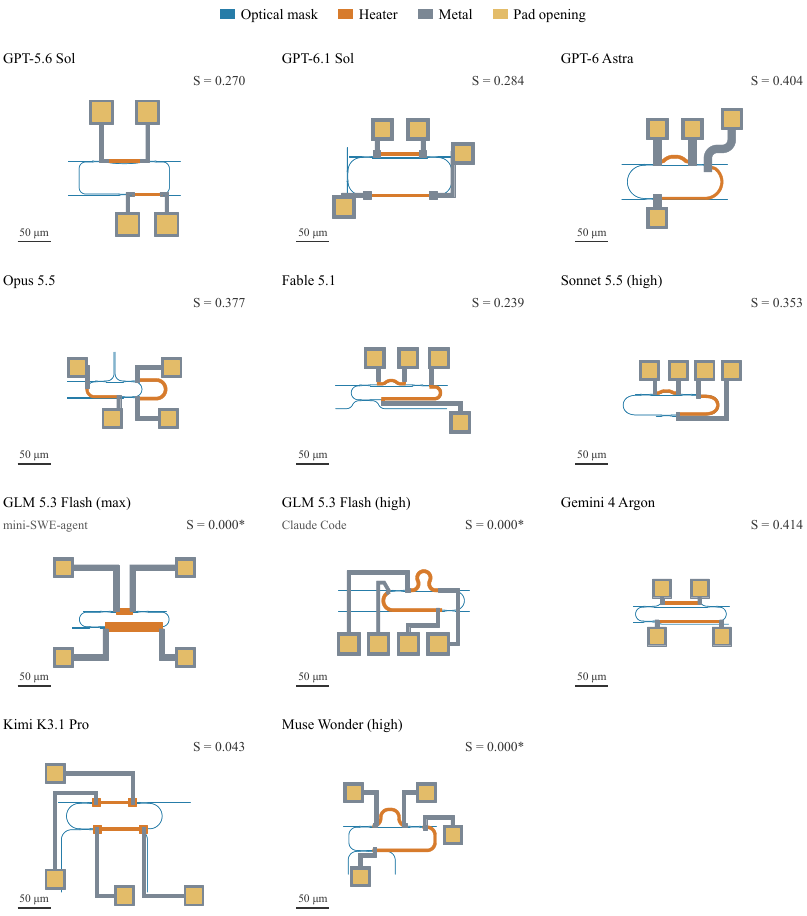}
    \caption{\textbf{MZI-Ring-v1 final layouts.} First version of the electrothermal MZI-coupled microring add-drop filter, with optical and electrical geometry.}
    \label{fig:final-layouts-mzi-coupled-microring-add-drop-filter-electrothermal-design-v1}
\end{figure}
\clearpage

\subsection{Microring-assisted Fourier spectrometer}
\label{app:layouts-microring-assisted-fourier-spectrometer-electrothermal-design-v1}

\begin{figure}[htbp]
    \centering
    \includegraphics[width=\linewidth]{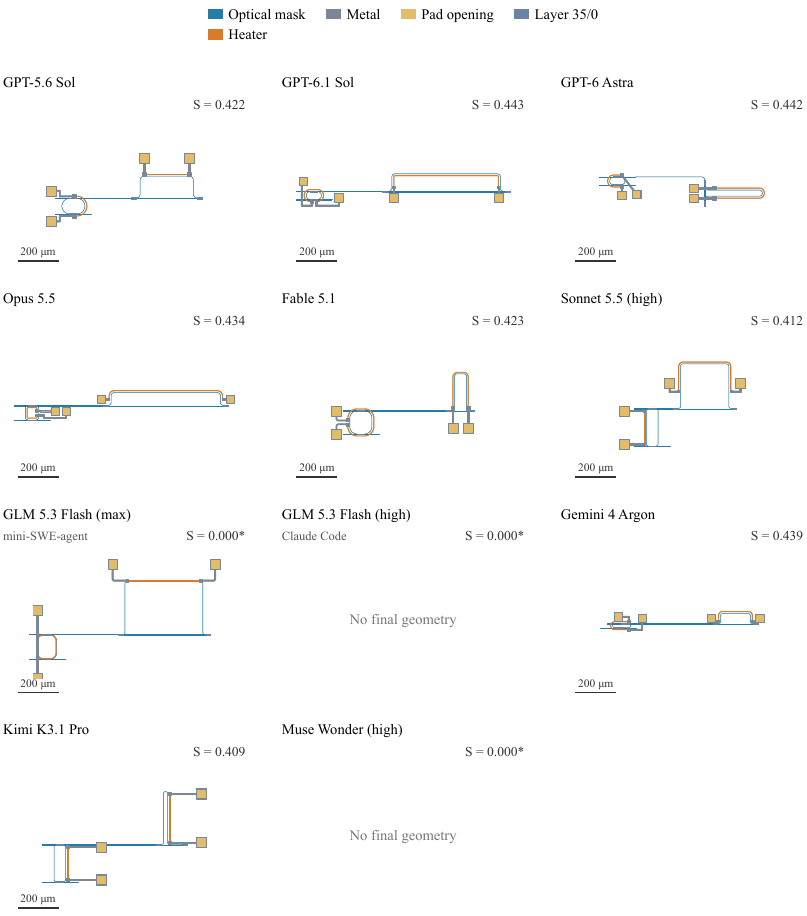}
    \caption{\textbf{RAFT-Spec final layouts.} Microring-assisted Fourier spectrometers with optical and electrical masks.}
    \label{fig:final-layouts-microring-assisted-fourier-spectrometer-electrothermal-design-v1}
\end{figure}
\clearpage

\subsection{Five-channel microresonator modulator}
\label{app:layouts-microresonator-modulator-design-v2}

\begin{figure}[htbp]
    \centering
    \includegraphics[width=\linewidth]{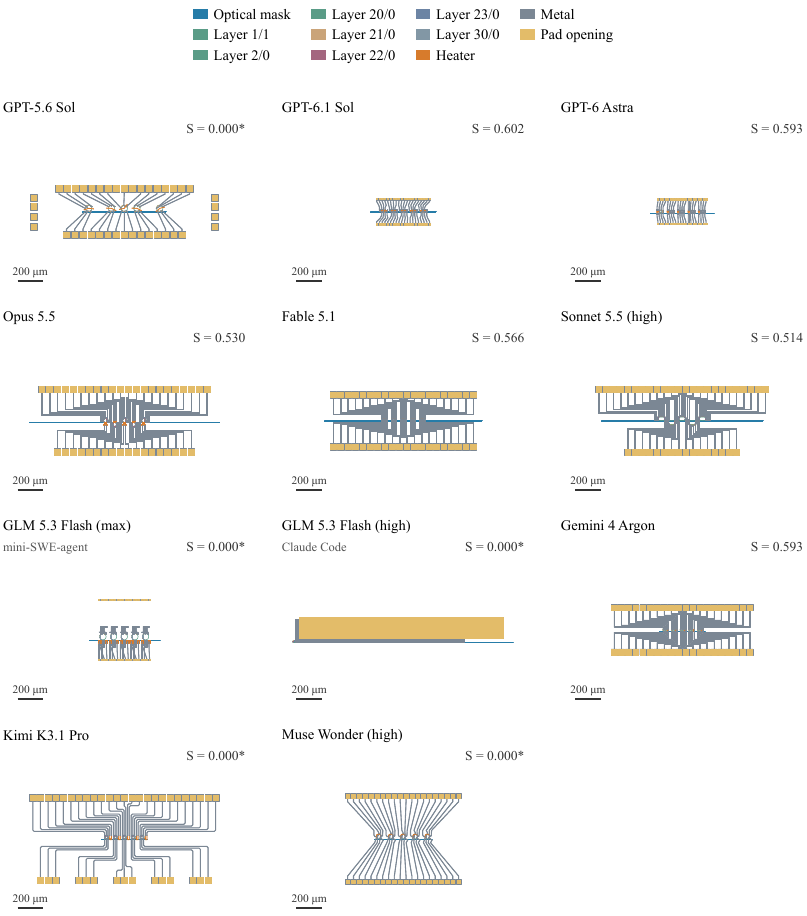}
    \caption{\textbf{Microresonator-v2 final layouts.} Five-channel microresonator modulators with resonator placement, electrical routing, and contacts.}
    \label{fig:final-layouts-microresonator-modulator-design-v2}
\end{figure}
\clearpage

\subsection{Eight-delay MZI spectrometer}
\label{app:layouts-eight-delay-mzi-spectrometer-design-v1}

\begin{figure}[htbp]
    \centering
    \includegraphics[width=\linewidth]{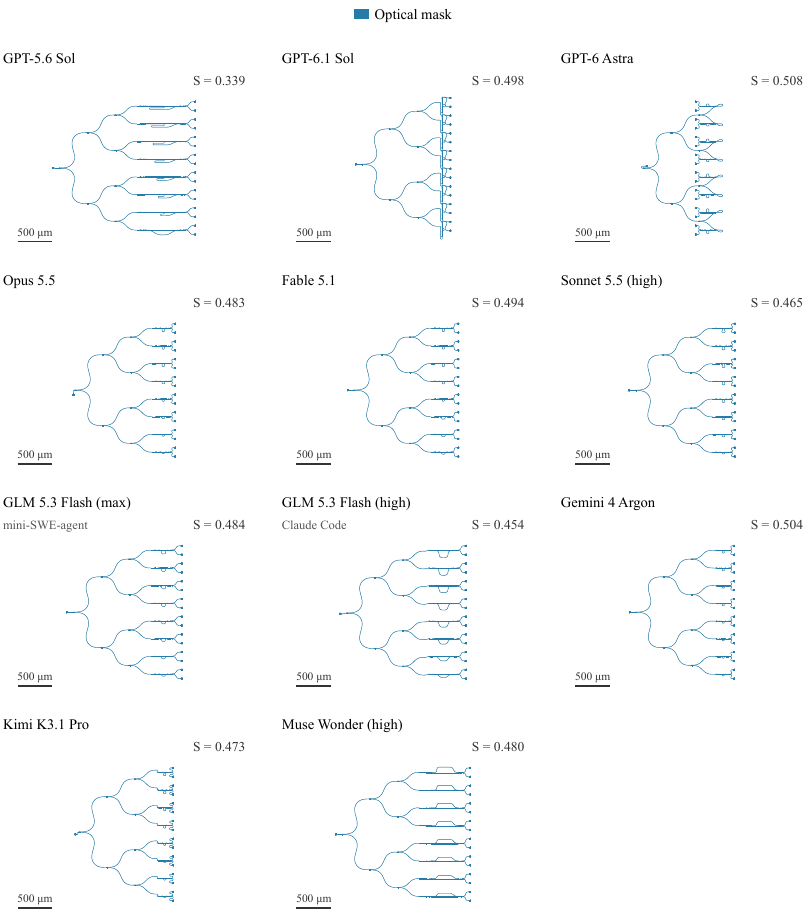}
    \caption{\textbf{MZI-Spec-8 final layouts.} Eight-delay spectrometers with a branching input network, eight Mach--Zehnder interferometers, and sixteen outputs. The layouts show differences in routing and delay arms.}
    \label{fig:final-layouts-eight-delay-mzi-spectrometer-design-v1}
    \label{fig:final-layouts-mzi-spectrometer}
\end{figure}
\clearpage

\subsection{Eight-mode quantum projector}
\label{app:layouts-eight-mode-quantum-projector-electrothermal-design-v2}

\begin{figure}[htbp]
    \centering
    \includegraphics[width=\linewidth]{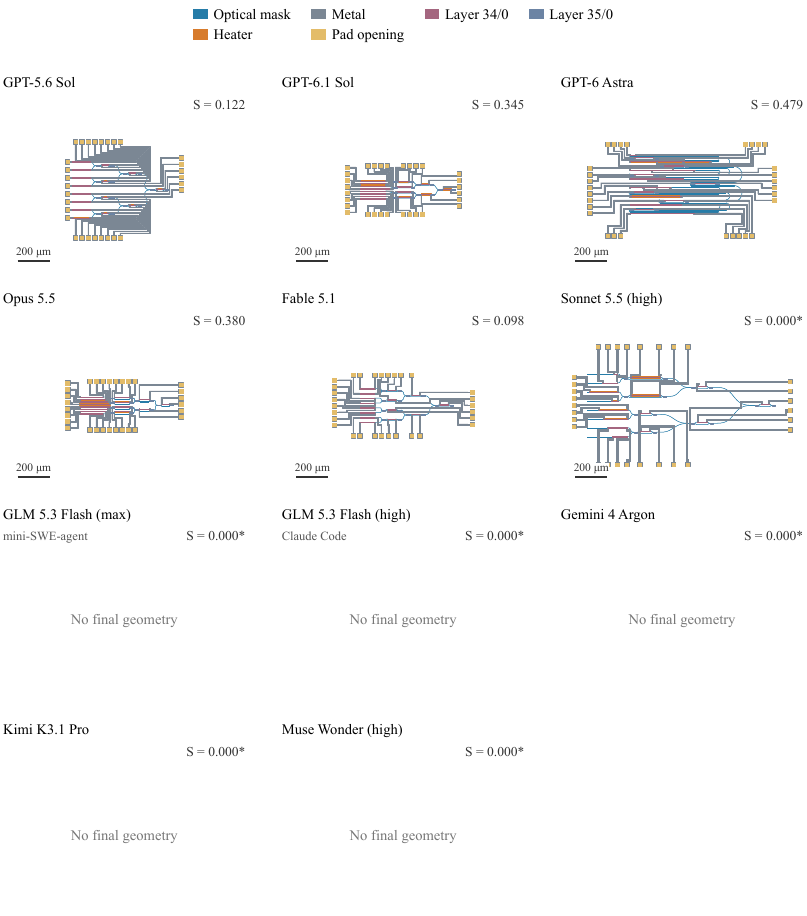}
    \caption{\textbf{Quantum-8 final layouts.} Eight-input quantum projectors with seven tunable MZIs, heated phase sections, electrical fanout, and perimeter contacts.}
    \label{fig:final-layouts-eight-mode-quantum-projector-electrothermal-design-v2}
\end{figure}
\clearpage

\subsection{Four-mode quantum projector}
\label{app:layouts-four-mode-quantum-projector-electrothermal-design-v1}

\begin{figure}[htbp]
    \centering
    \includegraphics[width=\linewidth]{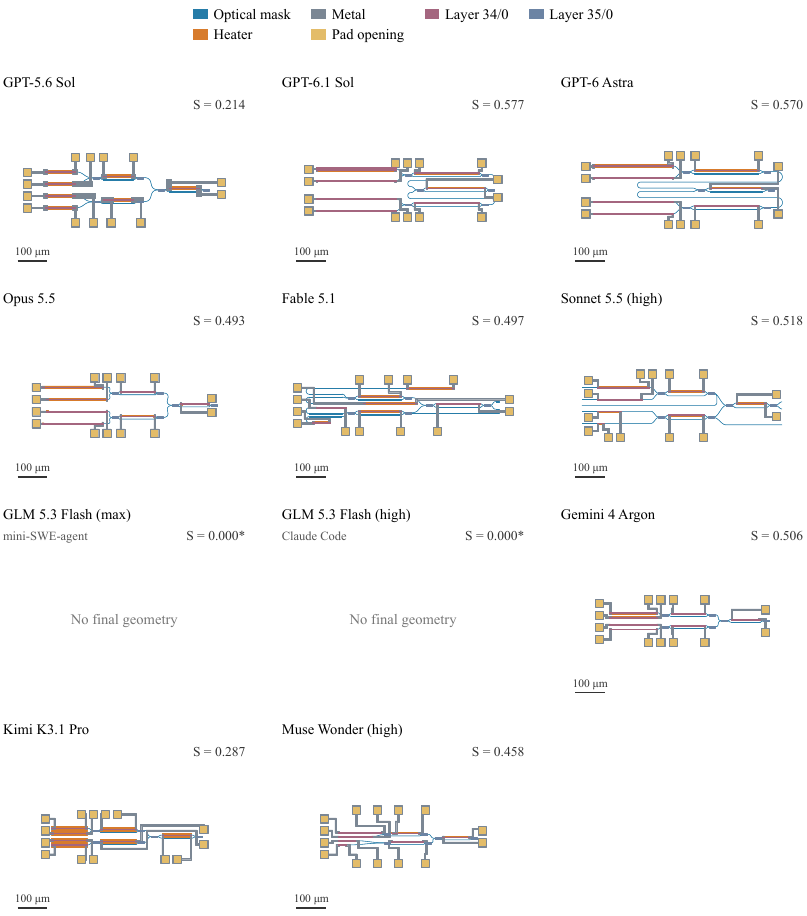}
    \caption{\textbf{Quantum-4 final layouts.} Four-input quantum projectors with three tunable MZIs, heated phase sections, electrical fanout, and perimeter contacts.}
    \label{fig:final-layouts-four-mode-quantum-projector-electrothermal-design-v1}
\end{figure}
\clearpage

\subsection{90-degree SiN waveguide bend}
\label{app:layouts-90sinwaveguide-design-v0}

\begin{figure}[htbp]
    \centering
    \includegraphics[width=\linewidth]{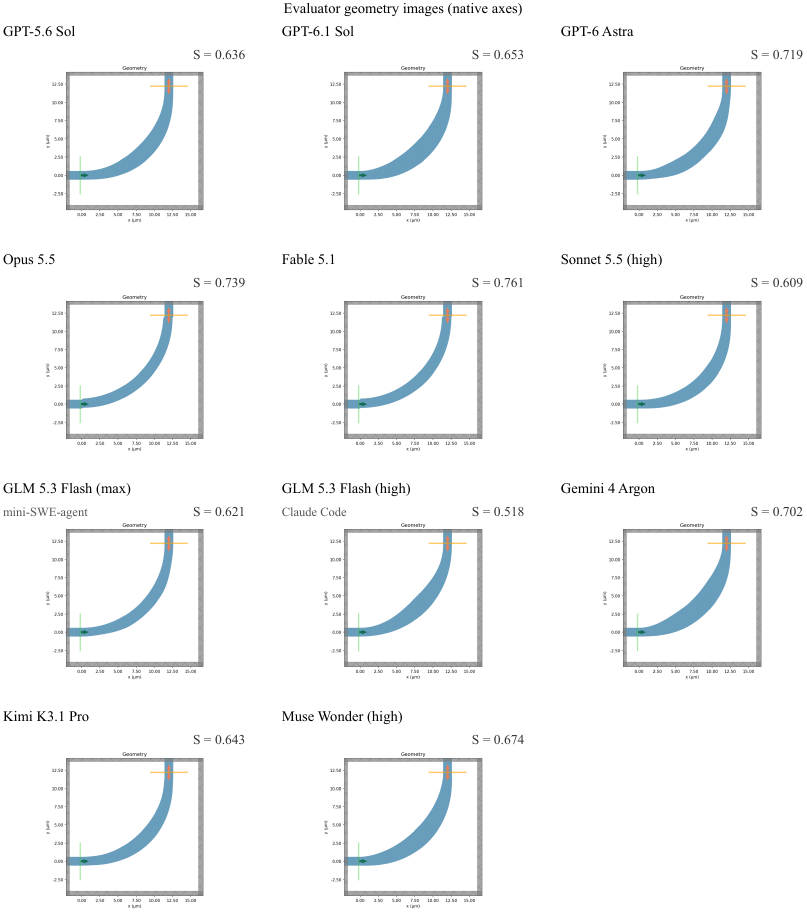}
    \caption{\textbf{90-degree SiN waveguide bend final layouts.} Original evaluator-generated images, rather than GDS layouts. Native axes and annotations are preserved; the images have not been reconstructed at a common GDS scale.}
    \label{fig:final-layouts-90sinwaveguide-design-v0}
\end{figure}
\clearpage

\subsection{Two-port ring cross-connect}
\label{app:layouts-four-wavelength-two-port-selective-crossconnect-design-v1}

\begin{figure}[htbp]
    \centering
    \includegraphics[width=\linewidth]{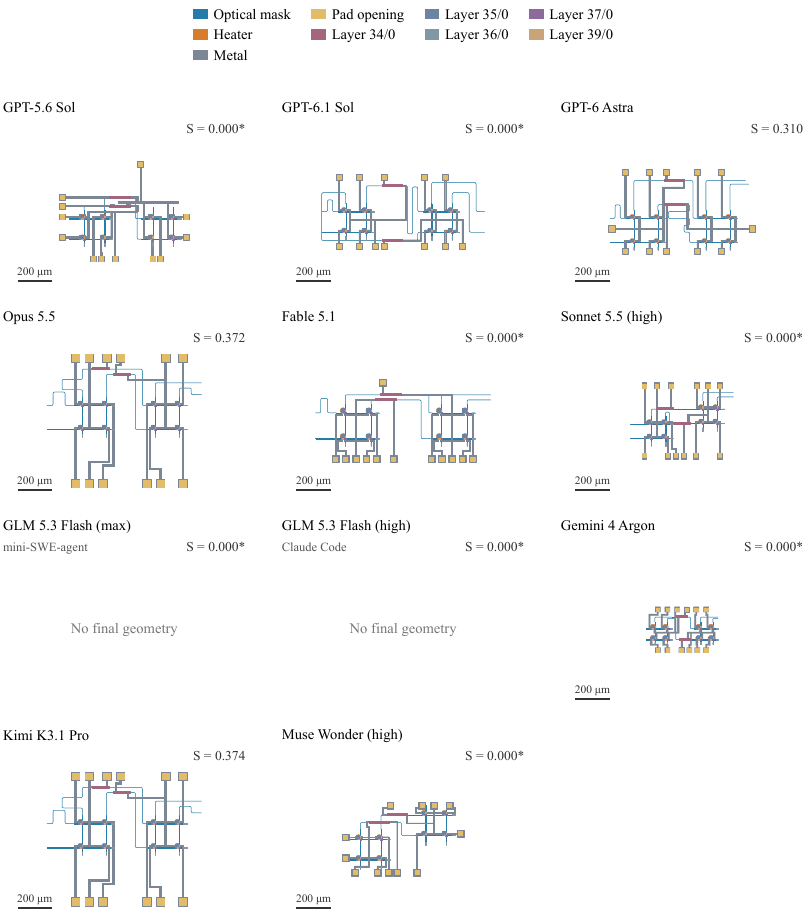}
    \caption{\textbf{Two-port ring cross-connect final layouts.} Four-wavelength, two-port ring cross-connects with optical and electrical masks.}
    \label{fig:final-layouts-four-wavelength-two-port-selective-crossconnect-design-v1}
\end{figure}
\clearpage

\subsection{Four-port ring cross-connect}
\label{app:layouts-four-wavelength-four-port-selective-crossconnect-design-v2}

\begin{figure}[htbp]
    \centering
    \includegraphics[width=\linewidth]{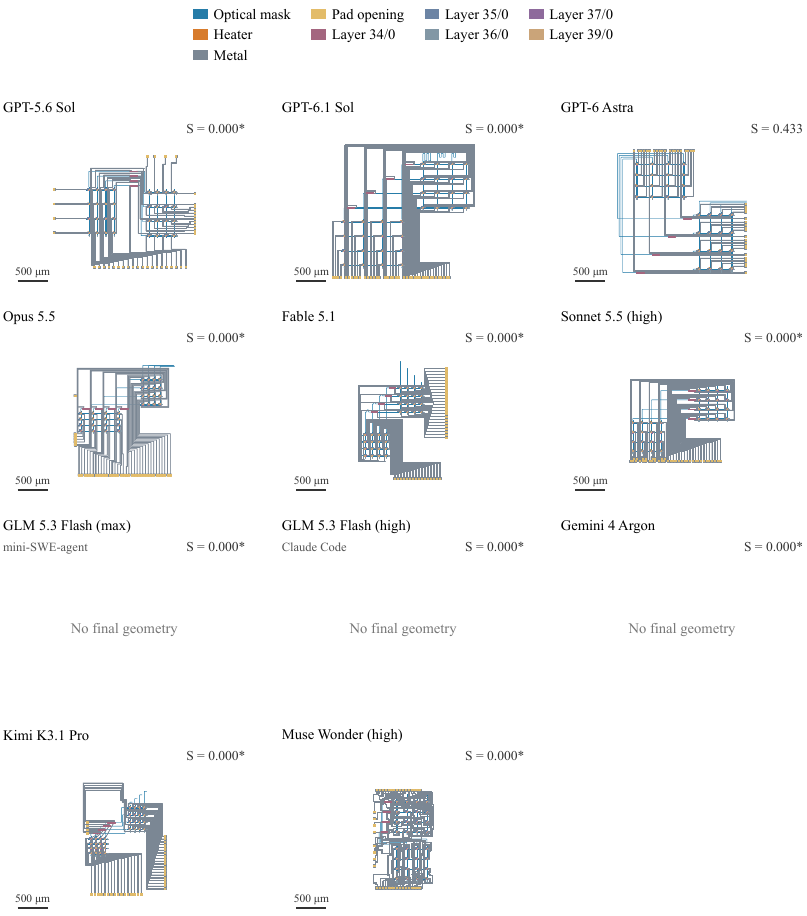}
    \caption{\textbf{Four-port ring cross-connect final layouts.} Four-wavelength, four-port ring cross-connects with optical and electrical masks.}
    \label{fig:final-layouts-four-wavelength-four-port-selective-crossconnect-design-v2}
\end{figure}
\clearpage

\subsection{Single-ring microresonator modulator}
\label{app:layouts-microresonator-modulator-design-v1}

\begin{figure}[htbp]
    \centering
    \includegraphics[width=\linewidth]{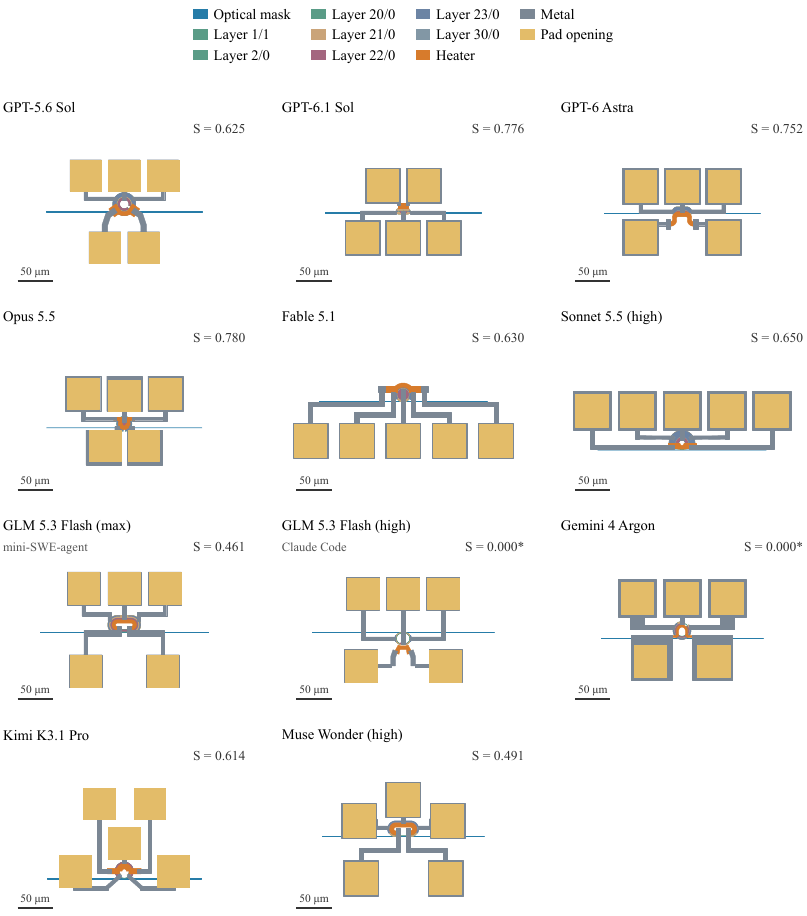}
    \caption{\textbf{Single-ring microresonator modulator final layouts.} Single-ring O-band modulators with optical and electrical masks.}
    \label{fig:final-layouts-microresonator-modulator-design-v1}
\end{figure}
\clearpage

\subsection*{Reimplementation tasks (10)}

\subsection{TFLN MMI crossings}
\label{app:layouts-tfln-mmi-crossing-v0}

\begin{figure}[htbp]
    \centering
    \includegraphics[width=\linewidth]{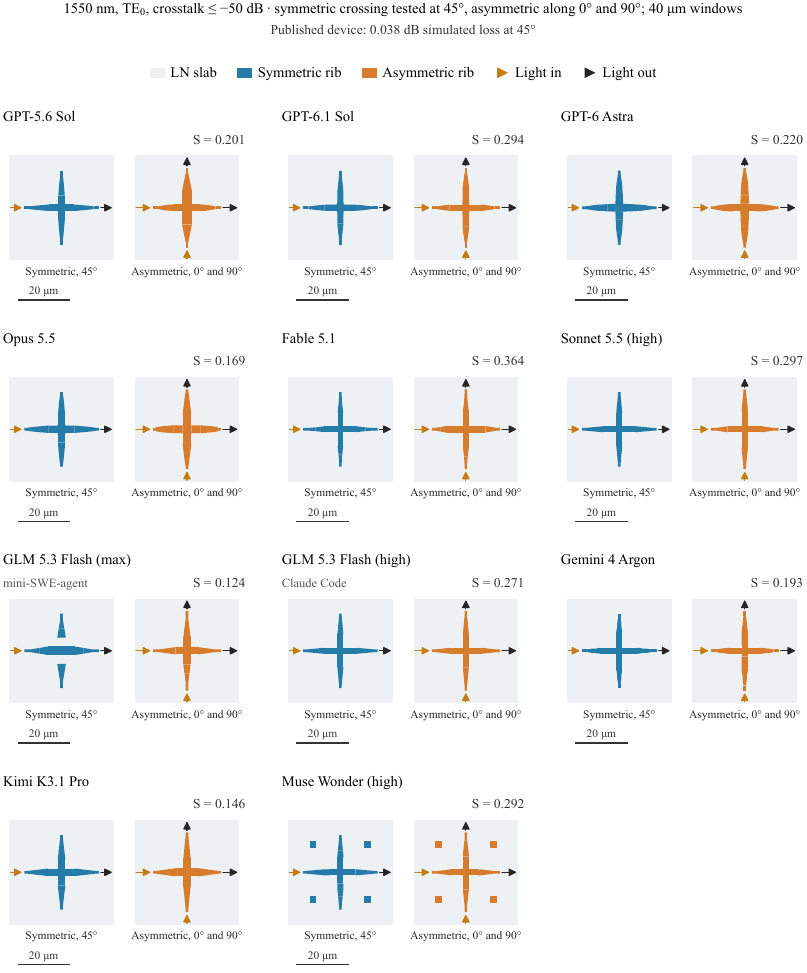}
    \caption{\textbf{TFLN Crossing-v0 final layouts.} Each submission includes symmetric (left) and asymmetric (right) rib masks over lithium-niobate slab windows. The gap between variants is for display only and does not represent a connection or added chip area. Each panel gives one score for the submitted pair.}
    \label{fig:final-layouts-tfln-mmi-crossing-v0}
    \label{fig:final-layouts-crossing}
\end{figure}
\clearpage

\subsection{HMDM polarization splitter--rotator}
\label{app:layouts-hmdm-polarization-splitter-rotator-v1}

\begin{figure}[htbp]
    \centering
    \includegraphics[width=\linewidth]{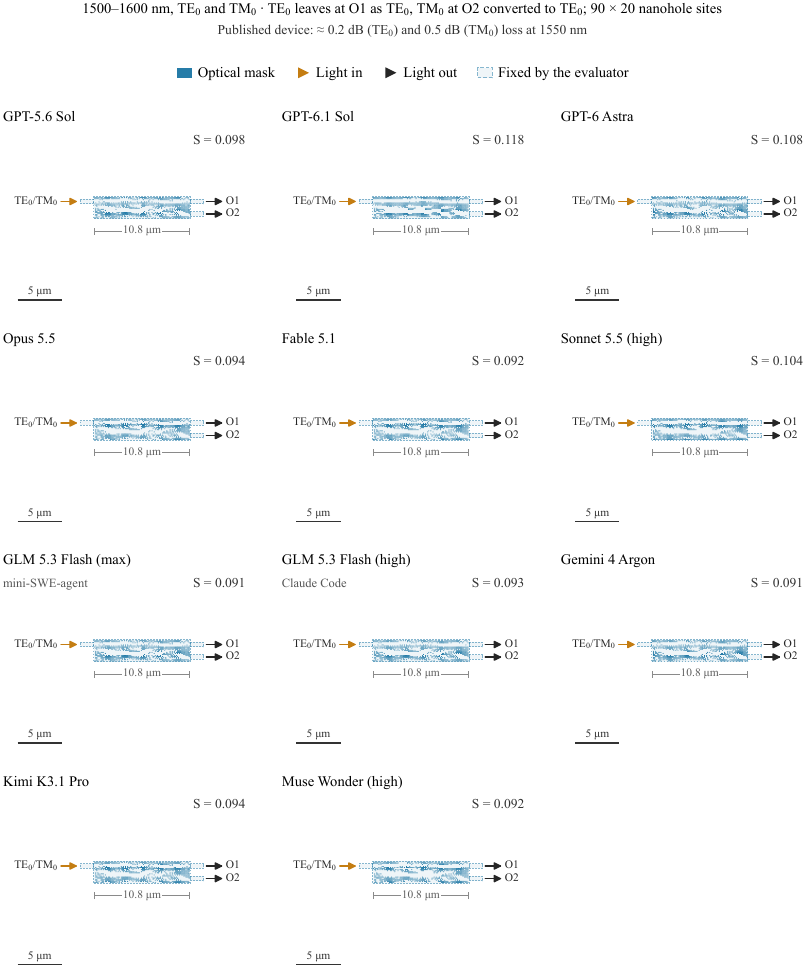}
    \caption{\textbf{HMDM polarization splitter--rotator final layouts.} Submitted nanohole masks. Parts supplied by the evaluator are shown as dashed outlines; the material stack is not drawn.}
    \label{fig:final-layouts-hmdm-polarization-splitter-rotator-v1}
\end{figure}
\clearpage

\subsection{Anisotropic SWG beamsplitter}
\label{app:layouts-anisotropic-swg-beamsplitter-v0}

\begin{figure}[htbp]
    \centering
    \includegraphics[width=\linewidth]{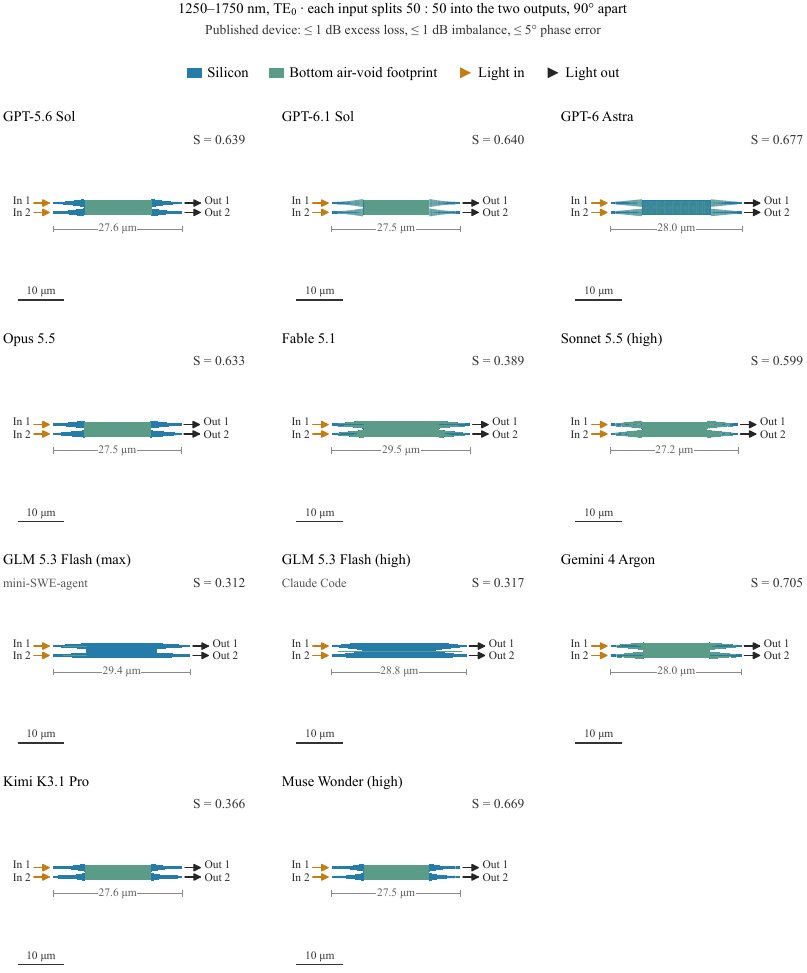}
    \caption{\textbf{Anisotropic SWG beamsplitter final layouts.} Submitted silicon masks for anisotropic subwavelength-grating beamsplitters, with the footprint of the bottom air void.}
    \label{fig:final-layouts-anisotropic-swg-beamsplitter-v0}
\end{figure}
\clearpage

\subsection{Apodized SWG grating coupler}
\label{app:layouts-asi-sin-swg-grating-coupler-v0}

\begin{figure}[htbp]
    \centering
    \includegraphics[width=\linewidth]{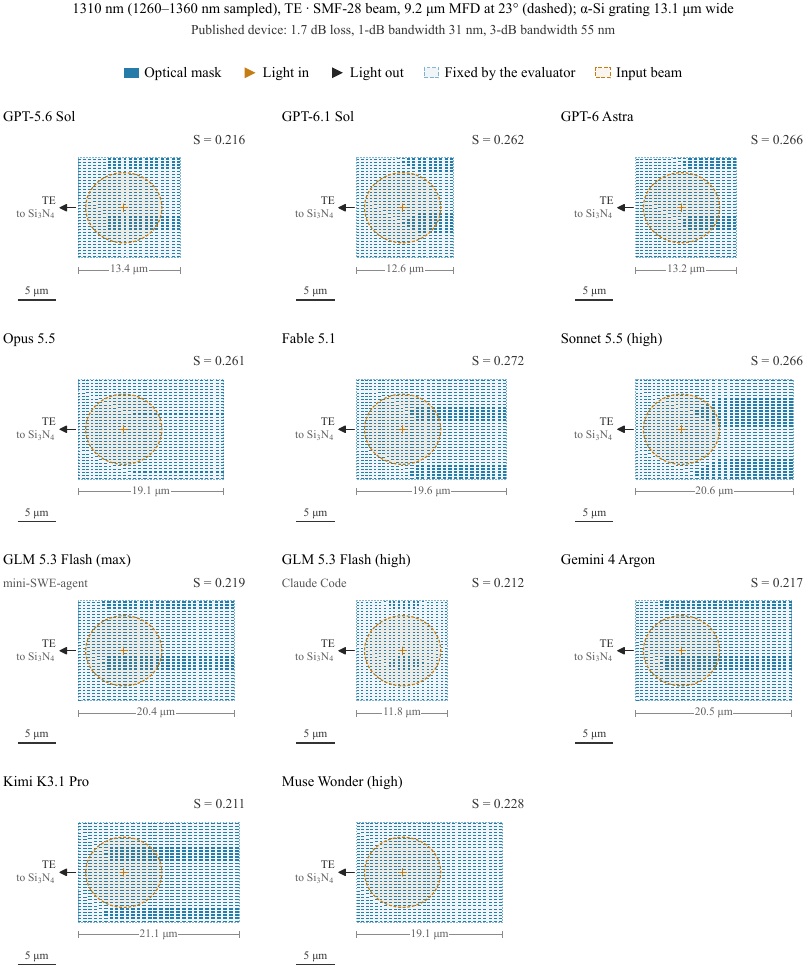}
    \caption{\textbf{Apodized SWG grating coupler final layouts.} Submitted $\alpha$-Si/Si$_3$N$_4$ etch-aperture masks; the evaluator-owned material stack is not part of the submitted GDS. Parts supplied by the evaluator are shown as dashed outlines, and the dashed outline marks the SMF-28 input beam.}
    \label{fig:final-layouts-asi-sin-swg-grating-coupler-v0}
\end{figure}
\clearpage

\subsection{Dual-trident SWG edge coupler}
\label{app:layouts-dual-trident-swg-edge-coupler-v0}

\begin{figure}[htbp]
    \centering
    \includegraphics[width=\linewidth]{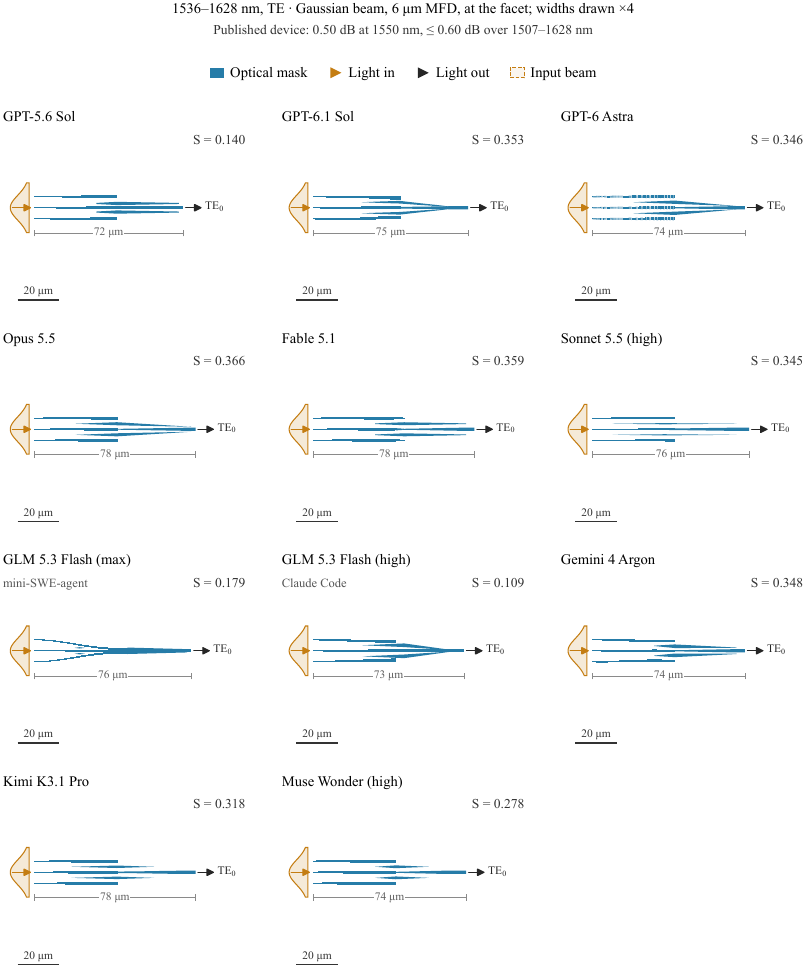}
    \caption{\textbf{Dual-trident SWG edge coupler final layouts.} Submitted geometries for dual-trident subwavelength-grating edge couplers. Widths are drawn four times larger than lengths for visibility.}
    \label{fig:final-layouts-dual-trident-swg-edge-coupler-v0}
\end{figure}
\clearpage

\subsection{HMDM six-channel WDM}
\label{app:layouts-hmdm-six-channel-wdm-v1}

\begin{figure}[htbp]
    \centering
    \includegraphics[width=\linewidth]{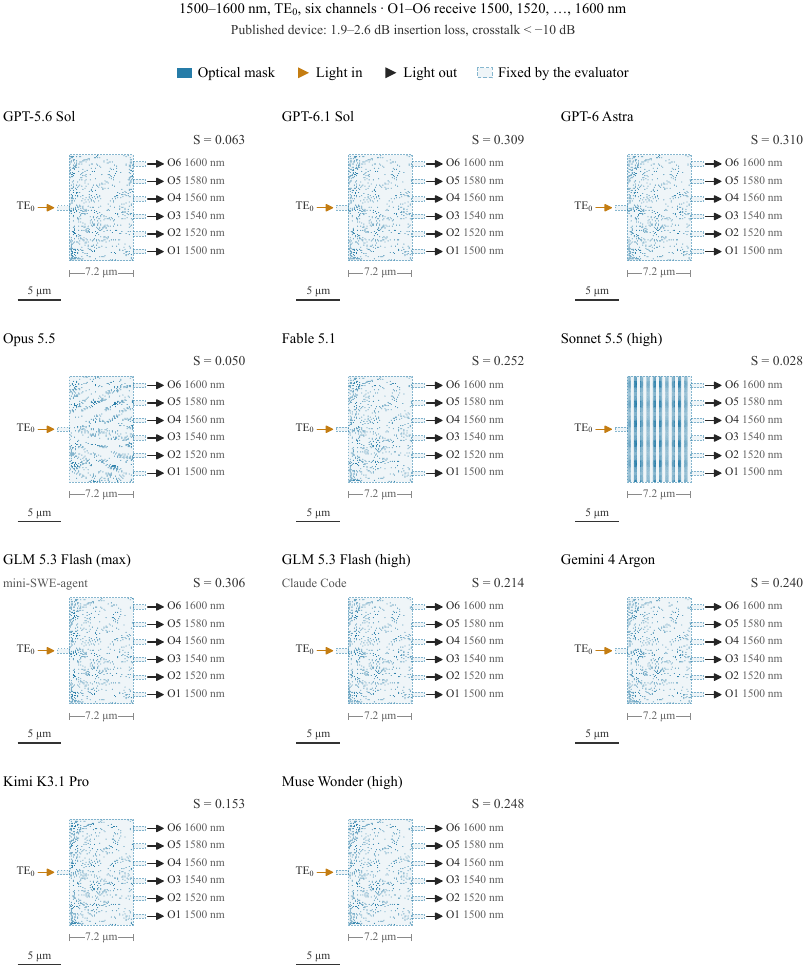}
    \caption{\textbf{HMDM six-channel WDM final layouts.} Nanohole masks on a fixed $60\times100$ binary grid covering $7.2\times12\,\mu\mathrm{m}$. Surrounding silicon and access waveguides belong to the evaluator and are shown as dashed outlines.}
    \label{fig:final-layouts-hmdm-six-channel-wdm-v1}
\end{figure}
\clearpage

\subsection{Sb2Se3/Si polarization rotator}
\label{app:layouts-sb2se3-si-polarization-rotator-v0}

\begin{figure}[htbp]
    \centering
    \includegraphics[width=\linewidth]{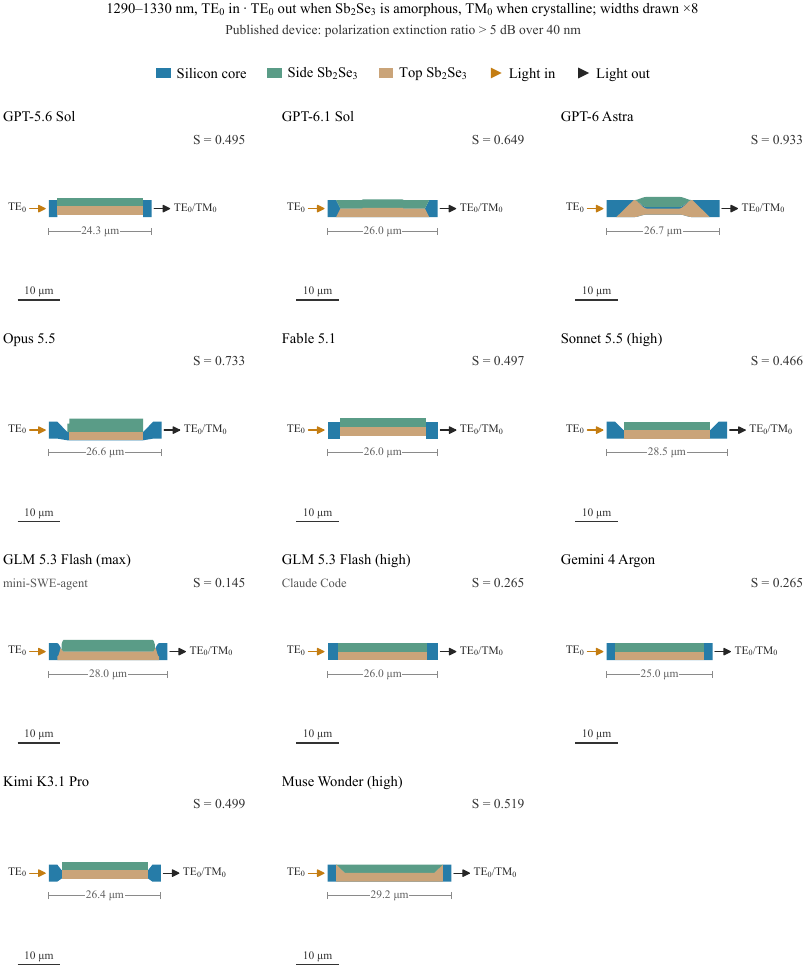}
    \caption{\textbf{Sb$_2$Se$_3$/Si polarization rotator final layouts.} Submitted silicon-core and side Sb$_2$Se$_3$ masks. Widths are drawn eight times larger than lengths for visibility.}
    \label{fig:final-layouts-sb2se3-si-polarization-rotator-v0}
\end{figure}
\clearpage

\subsection{SRSN grating coupler}
\label{app:layouts-srsn-grating-coupler-v0}

\begin{figure}[htbp]
    \centering
    \includegraphics[width=\linewidth]{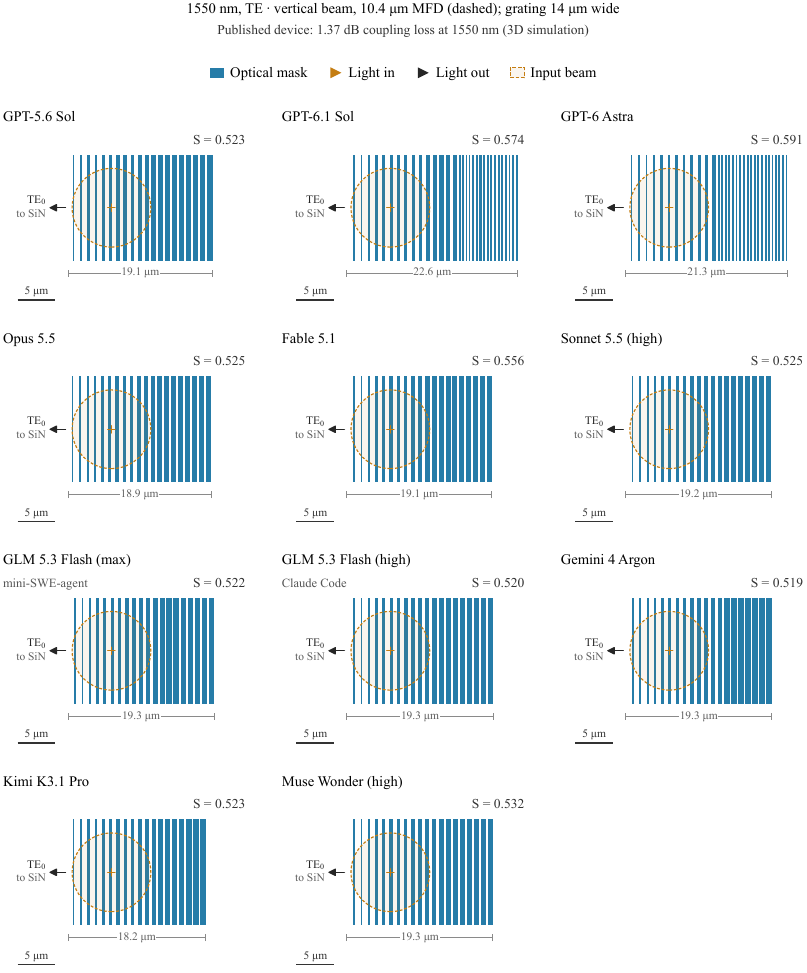}
    \caption{\textbf{SRSN grating coupler final layouts.} Submitted SRSN grating-coupler masks, highlighting differences in grating geometry.}
    \label{fig:final-layouts-srsn-grating-coupler-v0}
\end{figure}
\clearpage

\subsection{TFLN polarization-insensitive MMI}
\label{app:layouts-tfln-polarization-insensitive-mmi-v0}

\begin{figure}[htbp]
    \centering
    \includegraphics[width=\linewidth]{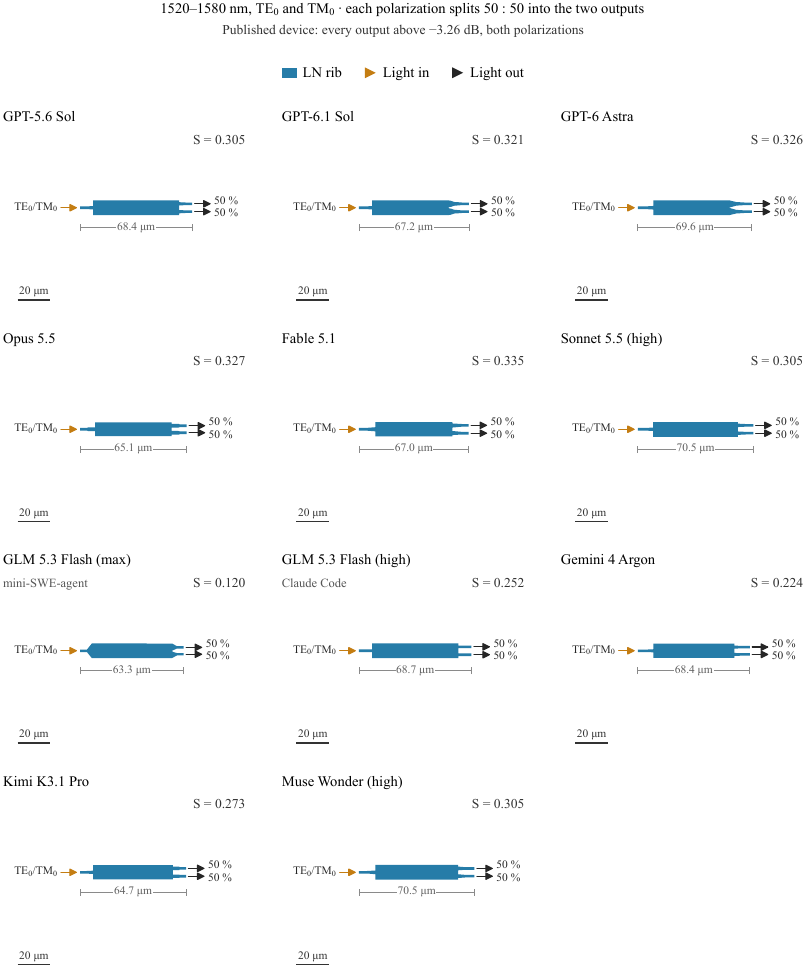}
    \caption{\textbf{TFLN polarization-insensitive MMI final layouts.} Submitted geometries for polarization-insensitive TFLN multimode-interference splitters.}
    \label{fig:final-layouts-tfln-polarization-insensitive-mmi-v0}
\end{figure}
\clearpage

\subsection{Trapezoidal SWG coupler}
\label{app:layouts-trapezoidal-swg-3db-coupler-v0}

\begin{figure}[htbp]
    \centering
    \includegraphics[width=\linewidth]{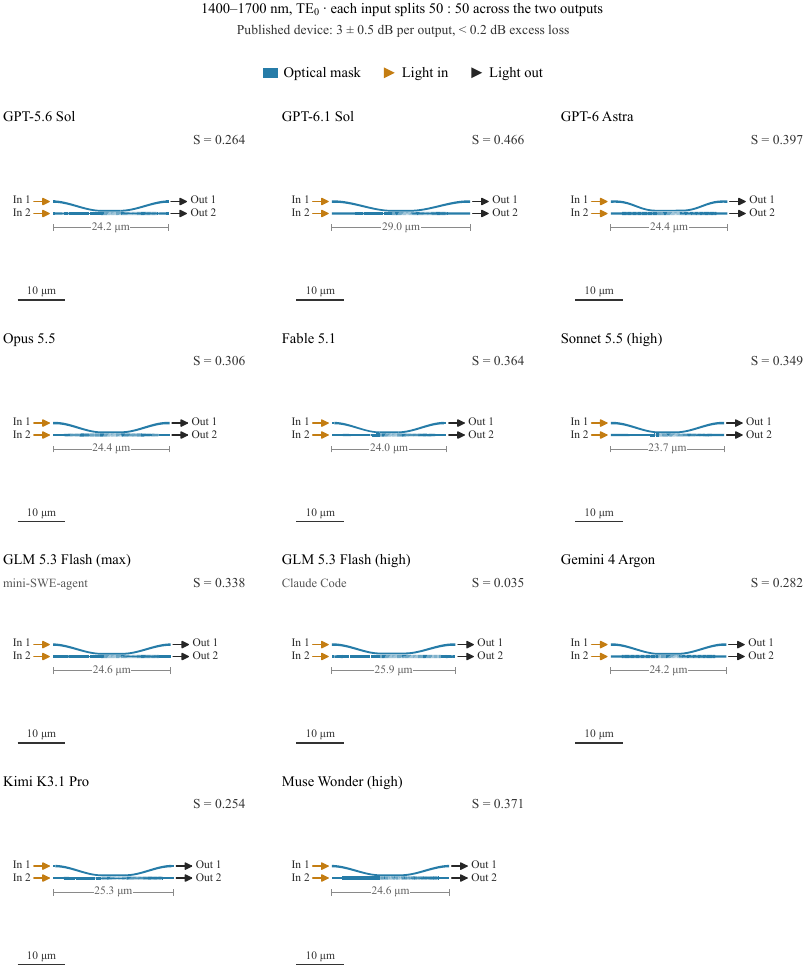}
    \caption{\textbf{Trapezoidal SWG coupler final layouts.} Submitted geometries for trapezoidal subwavelength-grating couplers.}
    \label{fig:final-layouts-trapezoidal-swg-3db-coupler-v0}
\end{figure}
\clearpage

\end{document}